\documentclass[10pt, journal]{IEEEtran}
\ifCLASSINFOpdf
\else
\fi
\usepackage{graphicx}
\usepackage{subfig}
\usepackage[font=scriptsize, skip=5pt]{caption}
\usepackage{amsmath}
\usepackage{bbm}
\usepackage{slashbox}
\usepackage{hhline}
\usepackage{multirow}
\usepackage{mathpazo}
\usepackage{enumitem}
\usepackage{epigraph}

\newcommand{\subparagraph}{}
\usepackage[pagestyles]{titlesec}

\usepackage{tikz}
\usepackage{lipsum}

\titlespacing*{\section}
{0pt}{2.3ex plus 1ex minus .2ex}{0.5ex plus .2ex}
\titlespacing*{\subsection}
{0pt}{1.2ex plus 1ex minus .2ex}{0.3ex plus .2ex}
\titlespacing*{\subsubsection}
{0pt}{1.2ex plus 0.5ex minus .1ex}{0.2ex plus .2ex minus .1ex}

\begin{document}

\setlength{\abovedisplayskip}{0pt}
\setlength{\belowdisplayskip}{2pt}
\setlength{\belowcaptionskip}{-4pt}
%
\title{Context Model for Pedestrian Intention Prediction using Factored Latent-Dynamic Conditional Random Fields}
%
%
%

\author{Satyajit~Neogi,~\IEEEmembership{Student Member,~IEEE,}
Michael~Hoy,~\IEEEmembership{Member,~IEEE,}
Kang~Dang,~\IEEEmembership{Member,~IEEE,}
Hang~Yu,~\IEEEmembership{Member,~IEEE,}
and~Justin Dauwels,~\IEEEmembership{Senior Member,~IEEE} }

%
%

\markboth{\copyright 2019 IEEE. Accepted by IEEE Transactions on Intelligent Transportation Systems}%
{Shell \MakeLowercase{\textit{et al.}}: Bare Demo of IEEEtran.cls for IEEE Journals}
%



\maketitle

\begin{abstract}
Smooth handling of pedestrian interactions is a key requirement for Autonomous Vehicles (AV) and Advanced Driver Assistance Systems (ADAS). Such systems call for early and accurate prediction of a pedestrian's crossing/not-crossing behaviour in front of the vehicle. Existing approaches to pedestrian behaviour prediction make use of pedestrian motion, his/her location in a scene and static context variables such as traffic lights, zebra crossings etc. We stress on the necessity of early prediction for smooth operation of such systems. We introduce the influence of vehicle interactions on pedestrian intention for this purpose. In this paper, we show a discernible advance in prediction time aided by the inclusion of such vehicle interaction context. We apply our methods to two different datasets, one in-house collected - NTU dataset and another public real-life benchmark - JAAD dataset. We also propose a generic graphical model Factored Latent-Dynamic Conditional Random Fields (FLDCRF) for single and multi-label sequence prediction as well as joint interaction modeling tasks. FLDCRF outperforms Long Short-Term Memory (LSTM) networks across the datasets ($\sim$100 sequences per dataset) over identical time-series features. While the existing best system predicts pedestrian stopping behaviour with 70\% accuracy 0.38 seconds before the actual events, our system achieves such accuracy at least 0.9 seconds on an average before the actual events across datasets. 
\end{abstract}



%
\IEEEpeerreviewmaketitle

\section{Introduction}   \label{sec:introduction}
%
%
%
%
\IEEEPARstart{A}{s}\footnote{S. Neogi, J. Dauwels, "PedAI: Smart Pedestrian Interactions for AV and ADAS", Singapore patent application number: 10202008970S. Filed 14th Sep, 2020.} we enter the era of autonomous driving with the first ever self-driving taxi launched in December 2018, smooth handling of pedestrian interactions still remains a challenge. The trade-off is between on-road pedestrian safety and smoothness of the ride. Recent user experiences and available online footage suggest conservative autonomous rides resulting from the emphasis on on-road pedestrian safety. To achieve rapid user adoption, the AVs must be able to simulate a smooth human driver-like experience without unnecessary interruptions, in addition to ensuring 100\% pedestrian safety. 

Automated braking systems in an ADAS tackle the emergency pedestrian interactions. These brakes get activated on detecting pedestrians' crossing behaviours within the vehicle safety range. Such a system may need to brake strongly on late prediction of pedestrian crossing behaviour, which can occur frequently in crowded areas. A future ADAS must be able of offer a smoother experience on such interactions.

The key to a safe and smooth autonomous pedestrian interaction lies in early and accurate prediction of a pedestrian's crossing/not-crossing behaviour in front of the vehicle. Accurate and timely prediction of pedestrian behaviour ensures on-road pedestrian safety, while early anticipation of the crossing/not-crossing behaviour offers more path planning time and consequently a smoother control over the vehicle dynamics. 

Recent works on on-road pedestrian behaviour prediction (\hspace{1sp}\cite{Daimler} - \cite{traj-est2}) rely on a pedestrian's motion, skeletal pose, his/her location in scene (on road, at curb etc.) and certain static context variables (e.g., presence of zebra crossings, traffic lights etc.). While pedestrian motion, skeletal pose and location are reliable indicators of the current pedestrian action, they hardly reflect on the long-term ($>$1 sec ahead) pedestrian behaviour. Static context variables (e.g., zebra crossings, traffic lights) certainly are important factors to predict future behaviour, but are seldom present in a scene. More often than not, an Autonomous Vehicle is likely to encounter pedestrian(s) without these variables. 

\begin{figure}
\includegraphics[scale=0.23]{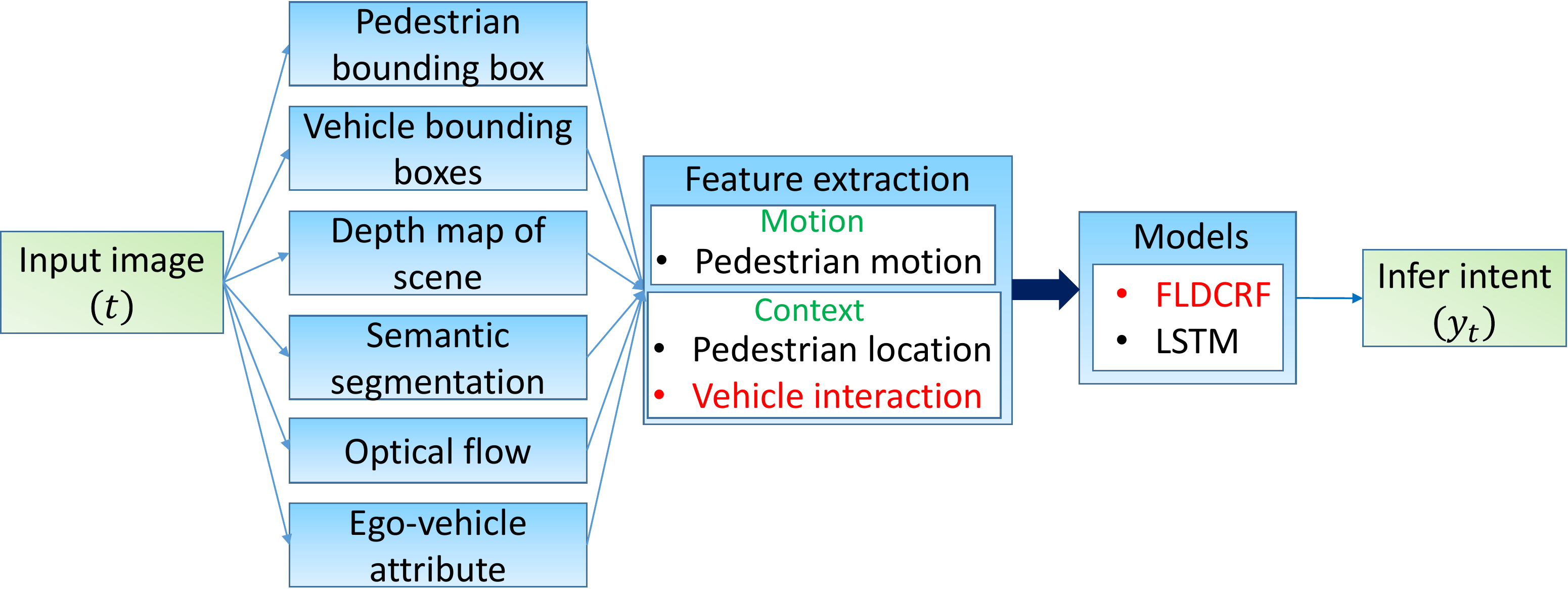}
\centering
\caption{Overview of our intention prediction system. Points in red highlight our main contributions.}
\label{fig:Overview}
\end{figure} 

As pedestrians, our crossing/not-crossing intentions are largely influenced by the distances and speeds of the approaching vehicles. We refer to this causal information as dynamic context or \textit{vehicle interaction context} (see Fig. \ref{fig:Overview}) and utilize it together with pedestrian motion and location information for early prediction of future behaviour. To the best of our knowledge, we are the first \cite{mypaper} to consider influence of such dynamic context on pedestrian intention, which proves to be the strongest factor for early prediction, as we demonstrate in Section \ref{sec:results}. Adding such context improves prediction time of the pedestrian stopping (at the curb) behaviour by $\ge$0.5 seconds on average across considered datasets (see Section \ref{sec:results}), and can potentially reduce unnecessary halts in future autonomous rides. This dynamic context also improves prediction time and accuracy of other behaviour types across the datasets.

Since each pedestrian interaction with the ego-vehicle is continuous in time for a certain duration (say, $T_p$) and we want to classify crossing/not-crossing intentions of a pedestrian at each instant within this duration, we formulate this problem as a time-series binary classification task, $P(y_t | x_{1:t})$, $t \in 1 \colon T_p$, where $y_t$ is the intention variable to be predicted and $x_t$ are the input features to the system at $t$. Conditional Random Fields (CRF) model such conditional task directly from the data. Each of the pedestrian crossing and not-crossing intentions has underlying latent intrinsic dynamics (see Section \ref{subsec:LDCRF} for examples), while there also exists an extrinsic dynamics between the intention labels. A Latent-Dynamic CRF (LDCRF, see Fig. \ref{fig:CDCRF}b) \cite{LDCRF} can capture the intrinsic dynamics within class labels $y_t$ by embedding latent states in a labeled CRF structure. In contrast to similar models (that capture the intrinsic dynamics) like Hidden Markov Models (HMM) \cite{HMM} and Hidden Conditional Random Fields (HCRF) \cite{HCRF}, a LDCRF does not require prior segmentation of the training sequences according to the class labels, thus preserving the extrinsic dynamics between labels as well in the model. LDCRF has been shown to outperform HMM and HCRF on similar tasks like gesture recognition \cite{LDCRF}. Considering its theoretical and experimental superiority over similar models, we apply LDCRF as baseline in our work.

The only way to improve learning performance with a LDCRF is by varying the number of hidden states (say, $N_h$) associated to each class label. Such a variation not only restricts the model capabilities, but also results in a rapidly growing state transition matrix with size $(N_l \cdot N_h) \times (N_l \cdot N_h)$, where $N_l$ is the number of class labels. Such increment results in greater model complexity and requires larger training data. Moreover, it is possible that there exists multiple interacting latent dynamics (e.g., contextual and motion-related, see Section \ref{subsec:LDCRF}) within the class labels. Considering these limitations of LDCRF and inspired by Factorial HMM \cite{Factorial-HMM}, we propose a generalization of LDCRF, called Factored LDCRF (FLDCRF) \cite{mypaper}, in order to 

\begin{itemize}[nosep]
\item{generate new models by varying the number of hidden layers,}
\item{generate factorized models \cite{Factorial-HMM} with fewer parameters to fit better small/medium datasets, and}
\item{capture relationship between multiple interacting latent dynamics (context and motion) within class labels. Each hidden layer models a different latent dynamics and the connections among different hidden layers model their interactions.}
\end{itemize}

We show such generalization of LDCRF to improve performance on our validation sets (see Section \ref{sec:results}). Such interacting hidden layers also allow to capture sequential multi-label/multi-agent interactions (see Fig. \ref{fig:FLDCRF-variants}). See Section \ref{subsec:FLDCRF} for more details on FLDCRF. 

LSTMs are the most popular kind of Recurrent Neural Networks (RNN), largely due to their ability to capture longer range dependencies among sequence values. They are frequently applied to similar sequence modeling tasks, e.g., path prediction \cite{LSTM}, action recognition \cite{action-lstm} etc. Deep learning systems like LSTMs have been known to involve tedious hyperparameter tuning and their performance are largely big-data-driven. We compare LSTM and FLDCRF (general LDCRF) on our datasets over identical input features and labels. FLDCRF not only outperforms LSTM across our datasets, but also provides easier model selection and requires significantly lesser training time (see Section \ref{sec:results}).

We highlight the main contributions of this paper below:
\begin{itemize}[nosep]
\item We propose a graphical model Factored Latent-Dynamic Conditional Random Fields (FLDCRF\footnote{https://github.com/satyajitneogiju/FLDCRF-for-sequence-labeling.}) for joint sequence prediction/tagging tasks. 
\item We introduce the influence of vehicle interactions (dynamic context) on pedestrian intention for early and accurate prediction. 
\item We have created a dataset\footnote{Dataset link to be provided.} specifically designed to capture vehicle influences on pedestrian intention, collected inside Nanyang Technological University campus (NTU dataset). We also conduct an evaluation set on a public real-life dataset, known as JAAD (Joint Attention for Autonomous Driving \cite{JAAD}), devised for early pedestrian intention prediction. 
\item We compare FLDCRF and LSTM over identical features and labels on both datasets.
\end{itemize}

In addition to early and accurate pedestrian intention prediction, we put stress on the necessity to simulate real-life prediction environment for our experiments. We apply state-of-the-art object detectors \cite{obj-det} and semantic segmentation systems \cite{semantic-seg} to detect pedestrians/vehicles and to determine pedestrian location respectively. Fig. \ref{fig:Overview} shows the different modules and their inter-connections in our intention prediction system.

The outline of this paper is as follows. In Section \ref{sec:lit-review}, we describe state-of-the-art systems on pedestrian behaviour prediction and CRFs. We briefly define a standard nomenclature on considered pedestrian behaviour types in Section \ref{sec:nomenclature}. We propose FLDCRF in Section \ref{sec:model} and introduce our context and motion feature extraction techniques in Section \ref{sec:features-labels}. Next, we discuss characteristics of the considered datasets in Section \ref{sec:dataset}. We describe our experiments in Section \ref{sec:exp-setup} and present our results in Section \ref{sec:results}. We offer concluding remarks and ideas for future research in Section \ref{sec:conclude}.

\section{Literature Review} \label{sec:lit-review}

From an Autonomous Vehicle perspective, there are two popular problems concerning pedestrian behaviour predictions, a) continuous path prediction \cite{Daimler}, \cite{Context-based} - \cite{traj-est1}, \cite{traj-est3}, \cite{traj-est5}, where future trajectory of a pedestrian is estimated and b) discrete intention prediction \cite{Daimler} - \cite{Michael}, where the goal is to predict discretized pedestrian behaviours. Given a detected pedestrian, his/her position w.r.t.~the AV and a marked AV safety range, solving any problem (path or intent prediction) can ensure a safe autonomous ride. However, a smooth autonomous ride of future (avoiding repeated halts and hard brakes) would require early predictions of pedestrian behaviour for optimal path and dynamics planning. Although pedestrian interactions and static traffic scene elements have been successfully utilized \cite{traj-est2} - \cite{traj-est1}, \cite{traj-est3}, \cite{traj-est5} to improve path prediction accuracy, long-term path prediction ($>$1 sec ahead) is still challenging and error prone, prediction error growing exponentially with prediction horizon \cite{traj-est2}. Additionally, path prediction given fewer ($\leq$10) observed frames, distant objects and moving cameras invoke additional uncertainties. Existing discrete intention prediction approaches (described shortly) primarily focus on accurate prediction rather than early prediction. In this paper, we analyze the feasibility of accurate long-term (up to 2 sec ahead) discrete intent prediction in a variety of scenarios.

We briefly discuss existing literature on discrete intent prediction and sequential CRFs below.

\subsection{Pedestrian Intention Prediction}

Existing intent prediction approaches can be broadly classified into two categories, a) Pedestrian-only, which only take features from the pedestrian of interest and b) Context based, which consider context variables (location, traffic elements etc.) together with pedestrian features.

\textbf{Pedestrian-only} approaches \cite{Daimler}, \cite{LDCRF-intent}, \cite{Intent-intersections}, \cite{HMM-intent},\cite{Openpose}, \cite{Michael}, in general, learn machine learning models over pedestrian features (motion and/or appearance) and n-ary intention labels. Pedestrian features include dense optical flow \cite{Daimler}, concatenated position, velocity and head pose \cite{LDCRF-intent}, Motion Contour image based Histogram of Oriented Gradients descriptor (MCHOG) \cite{Intent-intersections}, skeletal features \cite{HMM-intent}, \cite{Openpose} and deep learning appearance features \cite{Michael}. Applied machine learning have been either time-series models (GPDM, PHTM \cite{Daimler}, LDCRF \cite{LDCRF-intent}, HMM \cite{HMM-intent}, LSTM \cite{Michael}) or SVM \cite{Intent-intersections}, \cite{Openpose}. 

\textbf{Context based} intent prediction approaches \cite{Ped-crossing-intent}, \cite{Motion-classification}, \cite{Particle-intent}, \cite{JAAD-application} consider information like scene spatial layout, i.e. location of curb \cite{Ped-crossing-intent}, \cite{Motion-classification} and traffic elements \cite{Ped-crossing-intent}, \cite{Motion-classification}, \cite{Particle-intent}, \cite{JAAD-application}, such as cross walks, bus stops, traffic lights, zebra crossings etc. along with pedestrian features.

The majority of these approaches \cite{Daimler}, \cite{LDCRF-intent}, \cite{Openpose}, \cite{Michael} were evaluated on the Daimler dataset \cite{Daimler Dataset}, a public benchmark consisting of pedestrian crossing/not-crossing sequences. Stopping probability vs time \cite{Daimler}, \cite{LDCRF-intent}, \cite{Openpose}, \cite{Michael} and classification accuracy \cite{Daimler}, \cite{Intent-intersections}, \cite{HMM-intent}, \cite{JAAD-application} are the most common metrics. Since the Daimler data does not capture spontaneous vehicular interactions with pedestrians, we present our results on this data to Appendix \ref{Appen:Daim}. 

Recently a group of researchers have proposed a relatively large, labeled pedestrian dataset called JAAD \cite{JAAD}, captured from moving vehicles. This dataset provides real-life examples of pedestrian interactions with a vehicle in complex scenes. In \cite{JAAD-application}, the same group proposes a context based system to classify pedestrian crossing/not-crossing behaviour, where they combine an action recognition classifier and a learned attention model for relevant static context variables (traffic light, zebra crossing, stop sign etc.) to classify pedestrian intention. In \cite{JAAD-application2}, Gujjar et al. showed improvements in performance on the same dataset by considering a future generation model. However, we find their evaluation dataset to be weakly relevant to early prediction of intention (evaluation does not involve any temporal information) and hence select our own evaluation sequences (described in Section \ref{sec:exp-setup}) from the JAAD dataset.

All context models discussed above for pedestrian behaviour prediction rely on static scene context variables and do not consider dynamic vehicle interactions. To our best knowledge, we are the first to consider effects of such vehicle interactions on pedestrian behaviour \cite{mypaper}. While \cite{HMM-intent} presents a 70\% prediction accuracy of pedestrian stopping actions 0.06 seconds before the event on \cite{CMU-data}, we achieve such accuracy at least 0.9 seconds before the event across our datasets in presence of vehicle interactions. Current best approach \cite{Michael} on the Daimler dataset gives such an accuracy 0.38 seconds before event, although it is tested on a small test dataset (with 13 pedestrian sequences) with limited variations in pedestrian behaviour.

We apply proposed FLDCRF over input features (context + motion) and intention labels. We briefly describe related existing sequential CRF models below.

\subsection{Conditional Random Fields (CRFs)}

\begin{figure}%
\centering
\subfloat[\scriptsize{LCCRF}]{{\includegraphics[scale=0.45]{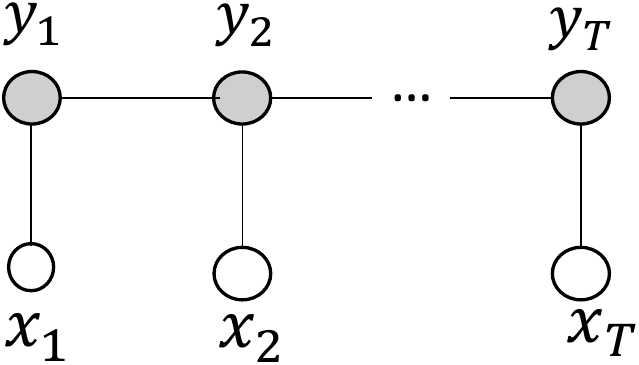} }}%
\qquad
\subfloat[\scriptsize{LDCRF}]{{\includegraphics[scale=0.45]{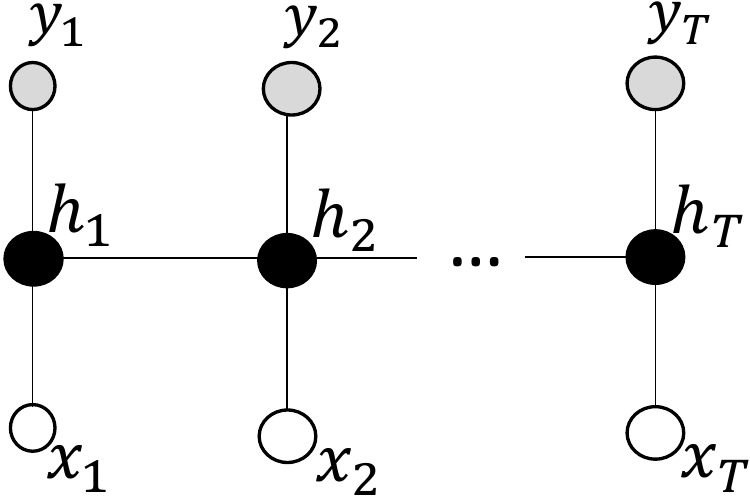} }}%
\caption{Sequential CRF variants. White nodes are (training + testing) observed and black nodes are hidden. Grey nodes are observed only during training.}%
\label{fig:CDCRF}%
\end{figure}

CRFs were first introduced in \cite{CRF-first} to accomplish the task of segmenting and labeling sequence data. CRFs are discriminative classifiers and aim at directly modeling the conditional distribution $P(\textbf{y}\mid\textbf{x})$ over input features \textbf{x} and classification labels \textbf{y}. A simple sequential type of CRFs is the Linear Chain CRF (LCCRF, see Fig. \ref{fig:CDCRF}a), which is frequently applied to sequence labeling tasks like word recognition, NP chunking, POS tagging etc. 

Over the years, multiple variants like semi-markov CRFs \cite{semi-CRF} and Dynamic CRFs \cite{DCRF} have been proposed. Semi-markov CRFs allow long range dependencies among labels $\textbf{y}$ = $\{y_1, y_2, ..., y_T\}$ at the expense of higher computational cost. Dynamic CRFs \cite{DCRF} allow multiple interacting labels ($y_{1,t}$, $y_{2,t}$ etc.) at each time step, making it suitable for multi-label classification tasks.

LDCRF was proposed in \cite{LDCRF} (see Fig. \ref{fig:CDCRF}b), where hidden states are embedded in a labeled CRF structure by means of an additional layer of hidden variables $\textbf{h}$ = $\{h_1, h_2, ..., h_T\}$. Each possible classification label is associated with a given set of hidden states. The hidden states model the intrinsic dynamics within each label.

FLDCRF utilizes primary LDCRF concept and constraints. We will briefly discuss the concept behind LDCRF and its applicability on the intent prediction problem in Section \ref{subsec:LDCRF}.

\section{Nomenclature} \label{sec:nomenclature}

In general, an on-road or near-road pedestrian's crossing/not-crossing behaviour is characterized by the real world lateral component of his/her motion w.r.t.~the ego-vehicle moving direction, i.e.,
\vspace{2mm}
\begin{align*}
laterally \quad moving \quad &\longleftrightarrow \quad crossing \\
laterally \quad static \quad &\longleftrightarrow \quad not-crossing. 
\end{align*}


At each instant $t$, we try to predict whether a detected pedestrian will be laterally moving or laterally static\footnote{Laterally static instances include pedestrians walking along the curb or sidewalk.} in the near future. Associated critical measures and necessary vehicle dynamics can be derived by further analyzing pedestrian's position w.r.t.~the vehicle.

\begin{figure}[h!]
\vspace{-2mm}
\centering
\subfloat[\scriptsize{Continuous crossing - continue moving laterally.}]{\label{fig:crossing-example}
\includegraphics[scale=0.27]{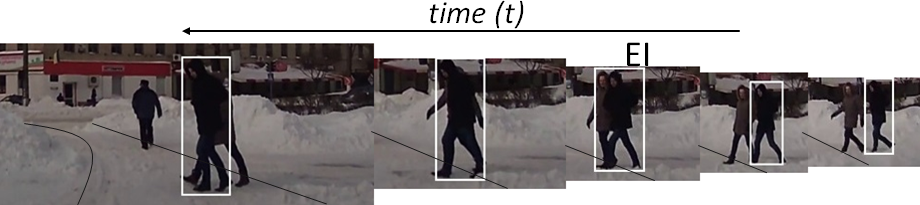}} \\[-0.3ex]
\vspace{-3mm}
\subfloat[\scriptsize{Stopping at the curb - laterally moving then static.}]{\label{fig:stopping-example}
\includegraphics[scale=0.46]{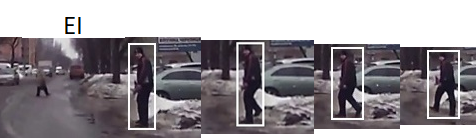}} \\[-0.3ex]
\vspace{-2mm}
\subfloat[\scriptsize{Standing at the curb - continue to be laterally static.}]{\label{fig:standing-example}
\includegraphics[scale=0.32]{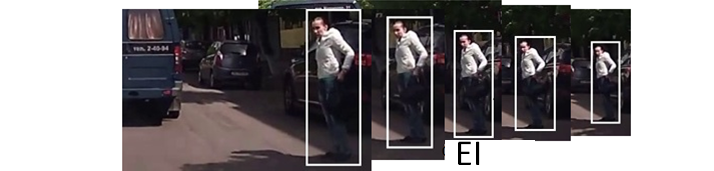}} \\[-0.3ex]
\vspace{-3mm}
\subfloat[\scriptsize{Starting to cross - laterally static then laterally moving.}]{\label{fig:starting-example}
\includegraphics[scale=0.46]{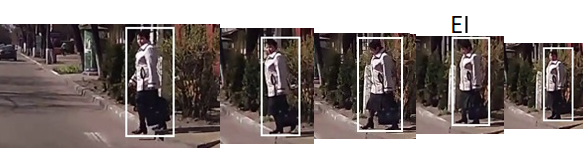}} \\[-0.3ex]

\caption{Examples of common pedestrian sequence types observed from a vehicle (from JAAD \cite{JAAD} dataset). EI stands for event (crossing, stopping etc.) instant.}
\label{fig:Example_sequences}
\end{figure}

We evaluate pedestrian sequences where early prediction is relevant from an AV perspective. We categorize each such encountered pedestrian sequence into one of the following (see Fig. \ref{fig:Example_sequences}): 

\begin{enumerate}
\item{\textbf{Continuous crossing}: The pedestrian initially has a lateral movement before the curb, continues to move laterally and crosses in front of the vehicle.}
\item{\textbf{Stopping at the curb}: The pedestrian initially has a lateral movement before the curb and stops at/near the curb. We call this behaviour `stopping' in the paper for simplicity.}
\item{\textbf{Standing at the curb}: The pedestrian continues to be laterally static at/near the curb. This behaviour is referred as `standing' in the paper.}
\item{\textbf{Starting to cross}: The pedestrian is initially static laterally at/near the curb and starts to cross in front of the vehicle. This behaviour is called `starting' in the paper.}
\end{enumerate}

Stopping and starting sequences carry a change in state of motion (laterally moving to laterally static and vice versa) of the pedestrian. As our goal is to predict pedestrian intention early, our primary aim is to detect these state changes early (preferably before any discernible change in motion/appearance) and accurately. In addition, we also want stable and accurate predictions of `crossing' and `not-crossing' intentions for sequences with no change in state of motion (continuous crossing and standing respectively). 

Each sequence type is associated with an \textit{event instant} (e.g., \textit{crossing}, \textit{stopping} or \textit{starting instant}), illustrated in Fig. \ref{fig:Example_sequences}, which provides the ground truth classification labels and also serves as reference during evaluation. Since a standing sequence cannot be characterized with such an instant, we define a \textit{critical point} based on pedestrian distance and ego-vehicle velocity. To aid the labeling process during training (for early prediction of the transitions in stopping and starting sequences), we define a \textit{pred\_ahead} parameter for each dataset, in number of frames before the respective event instants. We explain how we labeled the pedestrian sequences during training in Section \ref{sec:exp-setup}.

\section{Models} \label{sec:model}

In this section, we propose Factored Latent-Dynamic Conditional Random Fields (FLDCRF). We briefly discuss the idea behind applying LDCRF to our problem in Section \ref{subsec:LDCRF} and introduce FLDCRF in Section \ref{subsec:FLDCRF}. Since, FLDCRF subsumes LDCRF (structure and model constraints), we move the mathematical details about LDCRF to Appendix \ref{Appen:LDCRF}. 

\subsection{LDCRF} \label{subsec:LDCRF}

As mentioned before, we employ CRF type classifiers in our problem as they directly model the discriminative task $P(\textbf{y} \mid \textbf{x})$. Each of the pedestrian `crossing' and `not-crossing' intentions has underlying intrinsic dynamics. Such intrinsic dynamics can be contextual or motion-related. For example, stopping behaviour of a pedestrian (see Fig. \ref{fig:stopping-example}) can be characterized by a transition from pedestrian `moving' to `rest' state (motion) as well as an interacting vehicle going from `far' to `near' state (context). A standing behaviour may be associated with continuation of the `rest' and `near' states respectively. Since such motion-related or contextual states are hidden (latent), overall intrinsic dynamics of the `not-crossing' intention can be captured by associating hidden states to the intention label and allowing their transitions. These states can be learned from training data by connecting them to the observed context and motion features $x_t$. A LCCRF (see Fig. \ref{fig:CDCRF}a) only allows transition between class labels and hence fails to capture such intrinsic dynamics. LDCRF captures such dynamics by means of an additional layer of hidden variables $\{h_t\}_{t = 1:T}$ (see Fig. \ref{fig:CDCRF}b). $h_t$ can assume values from a predefined set of hidden states. For example, if the class labels `cross' or `not-cross' are associated with sets (of hidden states) $\mathcal{H}_{c}$ and $\mathcal{H}_{nc}$ respectively, and $y_t$ is `cross', then $h_t \in \mathcal{H}_{c}$. To keep training and inference tractable, LDCRF restricts sets $\mathcal{H}_{c}$ and $\mathcal{H}_{nc}$ to be disjoint.

\subsection{FLDCRF} \label{subsec:FLDCRF}

\begin{figure}[h]
\includegraphics[scale=0.45]{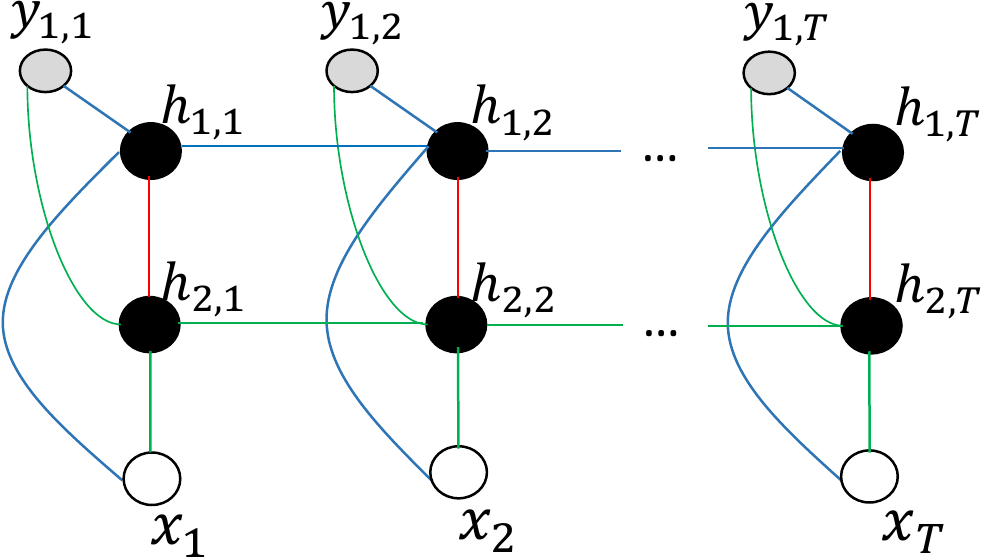}
\centering
\caption{FLDCRF graphical model for single-label sequence prediction. The graph shows only Markov connections for transitions between states. Longer range dependencies (semi-Markov for transitions and Markov/semi-Markov for \textit{red} inter-layer influences) are also possible but omitted for simplicity.}
\label{fig:FLDCRF single-label model}
\end{figure}

LDCRF captures the latent dynamics within class labels by means of a layer of hidden variables. We propose multiple interacting hidden layers in a LDCRF structure (see Fig. \ref{fig:FLDCRF single-label model}) to generate new models, reduce model parameters and to capture interaction among different latent dynamics within class labels (see Introduction).

\begin{figure}[h]
\includegraphics[scale=0.45]{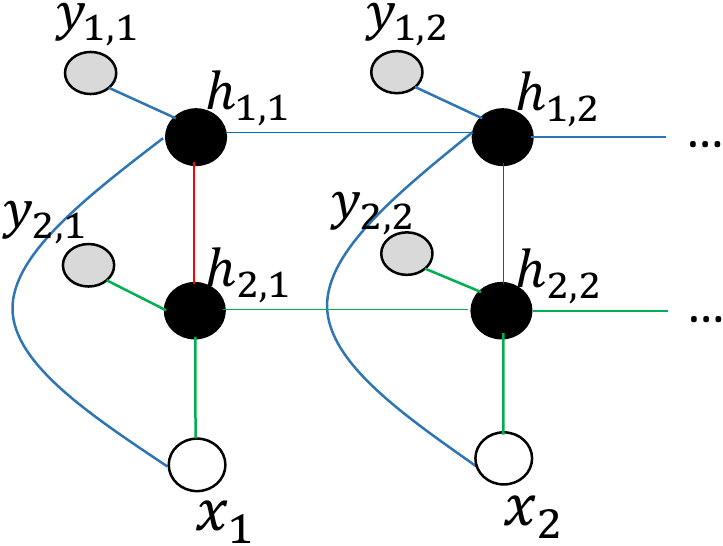}
\centering
\caption{FLDCRF graphical model for multi-label sequence prediction. Different label categories, $y_{1,t}$ and $y_{2,t}$, over input $x_t$ are connected through their respective hidden layers, $h_{1,t}$ and $h_{2,t}$, influencing each other.}
\label{fig:FLDCRF-variants}
\end{figure}

Such interacting hidden layers can also aid multi-label sequence prediction (see Fig. \ref{fig:FLDCRF-variants}) and social interactions in a multi-agent environment (see \cite{FLDCRF-arxiv}). We defer possible applications of such models to our future work. However, as the single-label variant of FLDCRF (see Fig. \ref{fig:FLDCRF single-label model}) is graphically a special case of the multi-label variant (see Fig. \ref{fig:FLDCRF-variants}), we describe the multi-label mathematical model for better generalization.

\subsubsection{Model} \label{FLDCRF-model}

Fig. \ref{fig:FLDCRF-variants} shows the graph structure for FLDCRF in multi-label classification problems. Although, we depict the model for only two hidden layers, we mathematically define the model for $L$ layers. 

Let, \textbf{x} = \{\(x_1, x_2, ... , x_T\)\} denote the sequence of observations. \({\textbf{y}_i}\) = \{\(y_{i,1}, y_{i,2}, ... , y_{i,T}\)\} are the observed labels along layer $i$, $i = 1:L$. In the case of a single-label prediction task, all layers take the same labels, i.e. $y_{i,t}$ is same $\forall i$. 

Let, \(y_{i,t} \) $\in$ $\Upsilon_i$, where \(\Upsilon_i \) is the alphabet for all possible label categories in layer $i$. \({\textbf{h}_i}\) = \{\(h_{i,1}, h_{i.2}, ... , h_{i,T}\)\} constitutes the $i$-th hidden layer. Each label \(\ell_i\) \(\in \) \(\Upsilon_i \) is associated with a set of hidden states \(\textit{$\mathcal{H}$}_{i,\ell_i} \). \({\mathcal{H}_i}\) is the set of all possible hidden states for layer \textit{i} written as \({\mathcal{H}_i}\) = \(\bigcup_{\ell_i} {\mathcal{H}_{i,\ell_i}} \). 

\vspace{1mm}
\noindent \textbf{Model constraints:} 

\begin{enumerate}
\item{\(\textit{$\mathcal{H}$}_{i,\ell_i} \) are disjoint $\forall \ell_i \in \Upsilon_i $, $\forall i = 1:L$. }

\item{$h_{i,t}$ can only assume values from the set of hidden states assigned to the label $y_{i,t}$, i.e., $h_{i,t}$ $\in$ $\textit{$\mathcal{H}$}_{i,y_{i,t}}$, $\forall i = 1:L$ and $\forall t = 1:T$.}
\end{enumerate}

The joint conditional model is defined as:
\vspace{2mm}
\begin{equation}
\begin{split}
\label{feqn1}
P\Big( \{{\textbf{y}_i}\}_{1:L} \mid \textbf{x}, \theta \Big) = \sum_{\{{\textbf{h}_i}\}_{1:L}} P\Big(\{{\textbf{y}_i}\}_{1:L} \mid \{{\textbf{h}_i}\}_{1:L}, \textbf{x}, \theta\Big) \\ \cdot \; P\Big(\{{\textbf{h}_i}\}_{1:L} \mid \textbf{x}, \theta\Big).
\end{split}
\end{equation}

Using the graph structure in Fig. \ref{fig:FLDCRF-variants}a, we obtain:
\vspace{2mm}
\begin{equation}
\begin{aligned}
\label{feqn1-new}
P\Big(& \{{\textbf{y}_i}\}_{1:L} \mid \textbf{x}, \theta \Big) = \sum_{\{{\textbf{h}_i}\}_{1:L}} \Bigg( \prod_{i = 1}^{L} P({\textbf{y}_i} \mid {\textbf{h}_i})\Bigg) P\Big(\{{\textbf{h}_i}\}_{1:L} \mid \textbf{x}, \theta\Big) \\ 
&= \sum_{\{{\textbf{h}_i}\}_{1:L} : \forall h_{i,t} \in \textit{$\mathcal{H}$}_{i,{y_{i,t}}}} \Bigg( \prod_{i = 1}^{L} P({\textbf{y}_i} \mid {\textbf{h}_i})\Bigg) P\Big(\{{\textbf{h}_i}\}_{1:L} \mid \textbf{x}, \theta\Big) \\ 
&+ \sum_{\{{\textbf{h}_i}\}_{1:L} : \exists h_{i,t} \not\in \textit{$\mathcal{H}$}_{i,{y_{i,t}}}} \Bigg( \prod_{i = 1}^{L} P({\textbf{y}_i} \mid {\textbf{h}_i})\Bigg) P\Big(\{{\textbf{h}_i}\}_{1:L} \mid \textbf{x}, \theta\Big).
\end{aligned}
\end{equation}

$P({\textbf{y}_i} \mid {\textbf{h}_i})$ can be further simplified by the graph as,
\vspace{1mm}
\begin{equation}
\label{feqn1-explanation}
P({\textbf{y}_i} \mid {\textbf{h}_i}) = \prod_{t = 1}^{T} P({y_{i,t}} \mid {h_{i,t}}).
\end{equation}

Applying model constraints, we can write,
\vspace{1mm}
\begin{equation}
P({y_{i,t}}=\ell_i \mid {h_{i,t}}) = 
\begin{cases}
1, & h_{i,t} \in \mathcal{H}_{i,{y_{i,t}=\ell_i}} \\
0, & h_{i,t} \not\in \mathcal{H}_{i,{y_{i,t}=\ell_i}}.
\end{cases}
\label{feqn1-constraint}
\end{equation}

Finally, the FLDCRF model in equation \eqref{feqn1-new} can be simplified by \eqref{feqn1-explanation} and \eqref{feqn1-constraint} as:
\vspace{2mm}
\begin{equation}
\label{feqn3}
P\Big( \{{\textbf{y}_i}\}_{1:L} \mid \textbf{x}, \theta \Big) = \sum_{\{{\textbf{h}_i}\}_{1:L} : \forall h_{i,t} \in \textit{$\mathcal{H}$}_{i,{y_{i,t}}}} P\Big(\{{\textbf{h}_i}\}_{1:L} \mid \textbf{x}, \theta\Big).
\end{equation}

$P\Big(\{{\textbf{h}_i}\}_{1:L} \mid \textbf{x}, \theta\Big)$ is defined from the CRF formulation,
\vspace{2mm}
\begin{equation}
\label{feqn4}
P\Big(\{{\textbf{h}_i}\}_{1:L} \mid \textbf{x}, \theta\Big) = \frac{1}{\textbf{\textit{Z}}(\textbf{x},\theta)} \exp\left(\sum_k \theta_k . F_k\Big(\{{\textbf{h}_i}\}_{1:L},\textbf{x}\Big)\right),
\end{equation}

\noindent where index $k$ ranges over all parameters $\theta = \{\theta_k\}$ and \( \textbf{\textit{Z}}(\textbf{x},\theta) \) is the partition function defined as:
\vspace{2mm}
\begin{equation}
\label{feqn5}
\textbf{\textit{Z}}(\textbf{x},\theta) = \sum_{\{{\textbf{h}_i}\}_{1:L}} {\exp\left(\sum_k \theta_k . F_k\Big(\{{\textbf{h}_i}\}_{1:L},\textbf{x}\Big)\right)}.
\end{equation}

In this paper, we only assume Markov connections (as depicted in Figures \ref{fig:FLDCRF single-label model} and \ref{fig:FLDCRF-variants}) along different hidden layers. Therefore, the feature functions \(F_k \)'s are defined as: 
\vspace{2mm}
\begin{equation}
\label{feqn6}
F_k\Big(\{{\textbf{h}_i}\}_{1:L},\textbf{x}\Big) = \sum_{t=1}^T f_k\Big(\{h_{i,t-1}\}_{1:L}, \{h_{i,t}\}_{1:L}, \textbf{x}, \textit{t}\Big).
\end{equation}

Also, we only allow different hidden layers to influence each other at the same time instant $t$. Thus, each component function $f_k\big(\{h_{i,t-1}\}_{1:L}, \{h_{i,t}\}_{1:L}, \textbf{x}, \textit{t}\big)$ can be a \textit{state} function $s_k(h_{i,t}, \textbf{x}, t) $, a \textit{transition} function $t_k(h_{i,t-1}, h_{i,t}, \textbf{x}, t) $ or an \textit{influence} function $i_k\big(h_{i,t}, h_{j,t}, \textbf{x}, \textit{t}\big)$, $i,j \in \{1:L\}$. We define \textit{state} and \textit{transition} functions by the following indicator functions,
\vspace{1mm}
\begin{equation}
\begin{split}
\label{eqn-state_tran}
s_k(h_{i, t}, \textbf{x}, \textit{t}) &= \mathbbm{1}_{\{(h_{i, t},x_t)= k\}}, \\
t_k(h_{i, t-1}, h_{i, t}, \textbf{x}, \textit{t}) &= \mathbbm{1}_{\{(h_{i, t},h_{i, t-1})= k\}},
\end{split}
\end{equation}

\noindent for discrete observations \textbf{x}. For continuous observations \textbf{x}, state functions are defined by:
\vspace{1mm}
\begin{equation}
\label{eqn-state}
s_k(h_{i, t}, \textbf{x}, \textit{t}) = \mathbbm{1}_{\{(h_{i, t})= k\}}. x_t.
\end{equation}

We define \textit{influence} functions $i_k\big(h_{i,t}, h_{j,t}, \textbf{x}, \textit{t}\big)$ over two or more (depending on the number of inter-related layers of hidden states) hidden variables between layers as:
\vspace{2mm}
\begin{equation}
\label{feqn-state_tran}
i_k\big(h_{i,t}, h_{j,t}, \textbf{x}, \textit{t}\big)= \mathbbm{1}_{\big\{\big(h_{i,t}, h_{j,t} \big)= k\big\}}, \quad i,j \in \{1:L\}.
\end{equation}

We avoid longer range influences (e.g., $h_{i,t-1}$ to $h_{j,t}$, $i,j \in \{1:L\} $) to reduce parameters. For the interaction model depicted in Fig. \ref{fig:FLDCRF-variants}b, the mathematical details remain same only with a minor change in the \textit{state} function, being $s_k(h_{i,t}, x_{i,t}$) instead of $s_k(h_{i,t}, x_t$).

A FLDCRF single-label variant with 1 hidden layer and $>$1 hidden states per label corresponds to a LDCRF, while 1 layer and 1 hidden state per label constitutes a LCCRF.

\subsubsection{Training model parameters} \label{FLDCRF-training}

We estimate the model parameters by maximizing the conditional log-likelihood of the training data given by:
\vspace{2mm}
\begin{equation}
\label{feqn7}
\textit{\textbf{L}}(\theta) = \sum_{n=1}^N {\log P\left(\{{\textbf{y}_i}\}_{1:L}^{\left(n\right)} \mid \textbf{x}^{\left(n\right)}, \theta\right)} - \frac{{\parallel\theta\parallel}^2}{2\sigma^2},
\end{equation}

\noindent where \textit{N} is the total number of available labeled sequences. The second term in equation \eqref{feqn7} is the log of a Gaussian prior with variance \(\sigma^2\).

\subsubsection{Inference} \label{FLDCRF-inference}

Multiple label sequences \({\textbf{y}_i}\), $i = 1:L$, can be inferred from the same graph structure by marginalizing over other labels:
\vspace{1mm}
\begin{equation}
\label{feqn8}
\hat{\textbf{y}}_i = \text{argmax}_{\textbf{y}_i} \quad \sum_{\big\{\{{\textbf{y}_i}\}_{1:L} - \; \textbf{y}_i\big\}} P\left(\{{\textbf{y}_i}\}_{1:L} \mid \textbf{x}, \hat{\theta}\right),
\end{equation}

\noindent where $P\left(\{{\textbf{y}_i}\}_{1:L} \mid \textbf{x}, \hat{\theta}\right)$ is obtained by \eqref{feqn3} and estimated parameters $\hat{\theta}$. At each instant $t$, the marginals $P(\{h_{i,t}\}_{1:L} \mid {x_{1:t}}, \hat{\theta})$ are computed and summed according to the disjoint sets of hidden states to obtain joint estimates of desired labels $\hat{y}_{i,t}$, $t = 1,2, ...$, $\forall i = 1:L $, as follows: 
\vspace{2mm}
\begin{equation}
\label{feqn9}
P(\{y_{i,t}\}_{1:L} \mid \textbf{x}) = \sum_{\{h_{i,t}\}_{1:L}:h_{i,t} \in \textit{$\mathcal{H}$}_{i,{y_{i,t}} }} P\left(\{h_{i,t}\}_{1:L} \mid {x_{1:t}}, \hat{\theta}\right).
\end{equation}

After marginalizing according to \eqref{feqn8}, the label $\hat{y}_{i,t}$ corresponding to the maximum probability is inferred. As the intention prediction problem must be solved online, we compute $P\left(\{h_{i,t}\}_{1:L} \mid {x_{1:t}}, \hat{\theta}\right)$ by the forward algorithm. Forward-backward algorithm \cite{HMM} and Viterbi algorithm \cite{Viterbi} can also be applied for problems where online inference is not required.

\section{Feature Extraction} \label{sec:features-labels}

We describe the process to prepare our feature vector $x_t$ in this section. $x_t$ comprises of two components: context features, $x_{c,t}$, and motion features, $x_{m,t}$.

\subsection{Context Features} \label{subsec:context}

Context features employed in our work are of two types:

\begin{figure}%
\centering
\subfloat[\scriptsize{}]{{\includegraphics[scale=0.3]{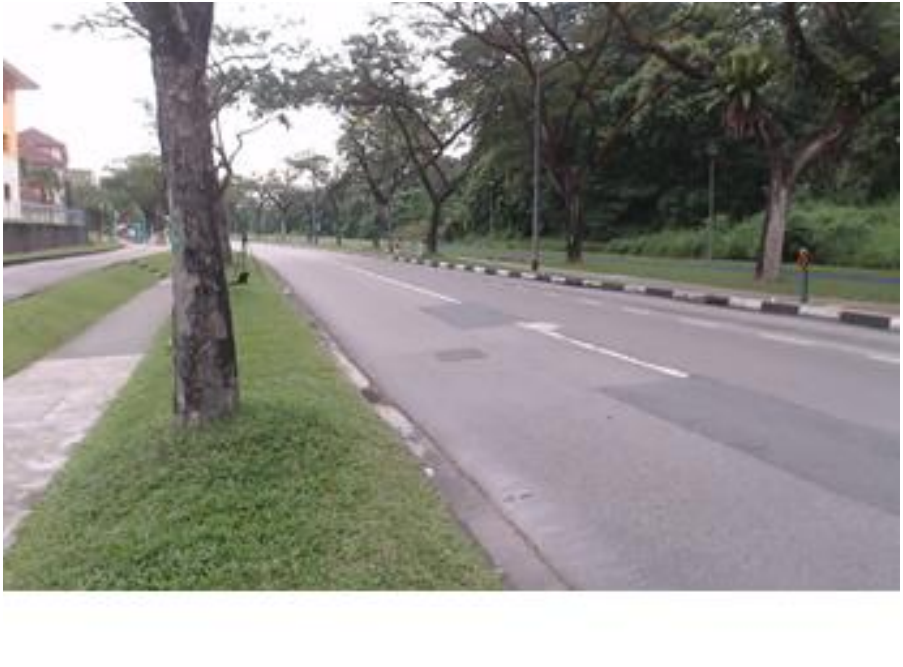} }}
\quad
\subfloat[\scriptsize{}]{{\includegraphics[scale=0.3]{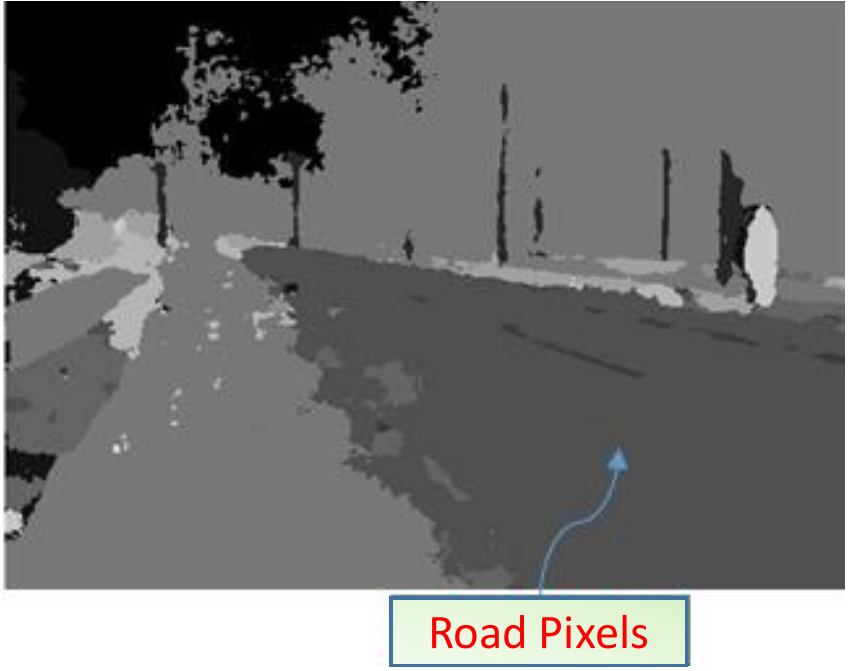} }}
\quad
\subfloat[\scriptsize{}]{{\includegraphics[scale=0.35]{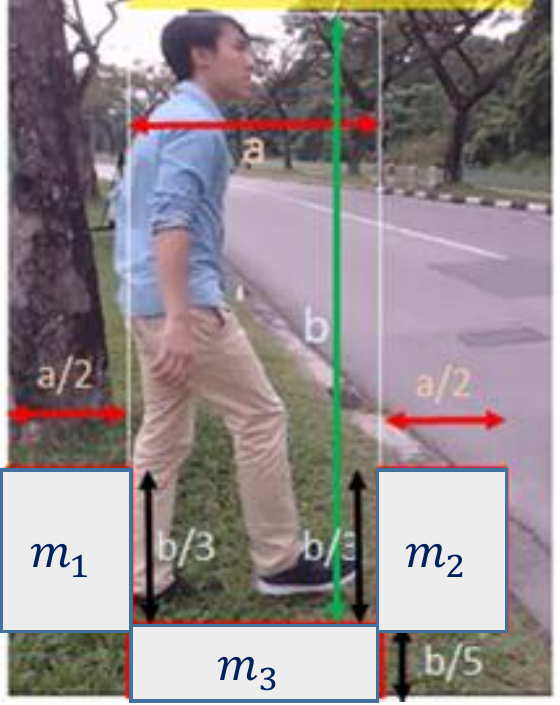} }}%
\quad
\subfloat[\scriptsize{}]{{\includegraphics[scale=0.4]{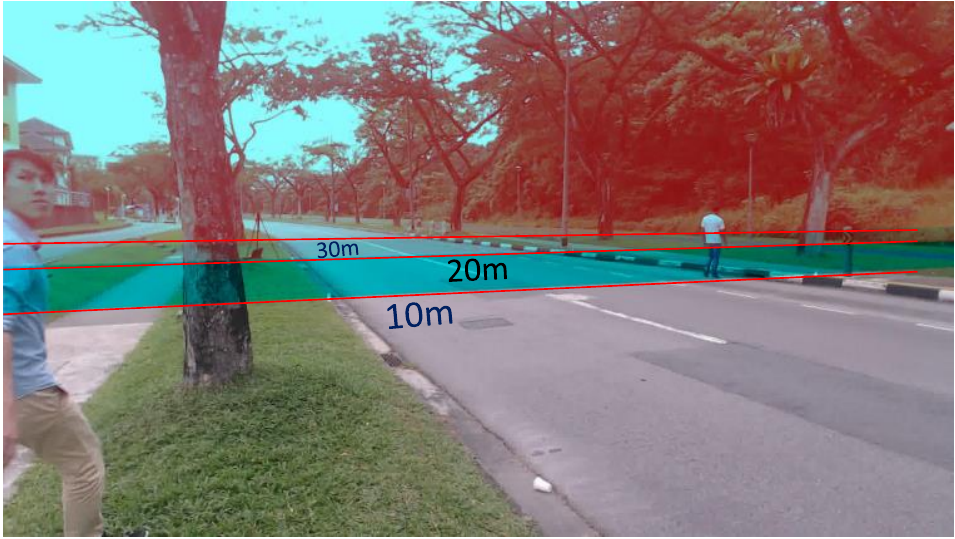} }}%
\quad
\subfloat[\scriptsize{}]{{\includegraphics[scale=0.27]{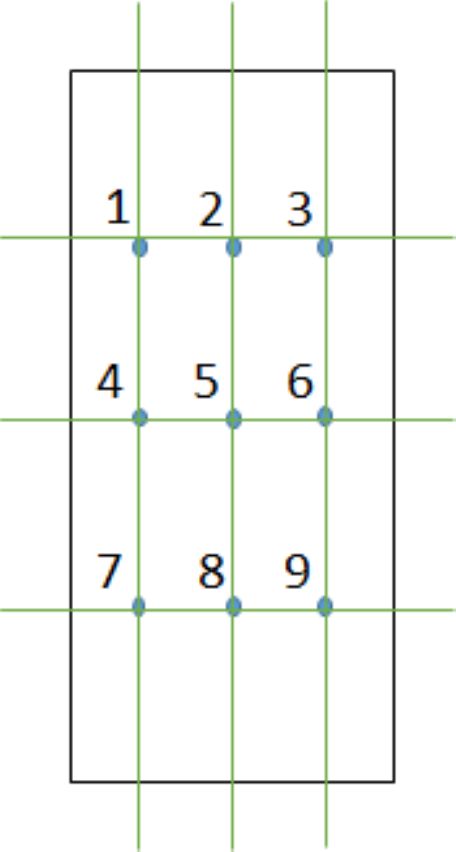} }}

\caption{Feature extraction: (a)-(c) depict pedestrian location context. Semantic segmentation of the scene in (a) is presented in (b). Road pixels are highlighted. (c) shows the different regions around pedestrian bounding box to capture necessary location context variables. (d) highlights pre-calibrated depth lines. Figures taken from the NTU dataset. (e) 9 points are selected within a pedestrian bounding box during motion feature extraction.}%
\label{fig:Context}%
\end{figure}

\begin{itemize}
\item{\textbf{Pedestrian location context}: Also called \textit{spatial context} in the paper, this context variable encodes the location (e.g. on road, at curb, away) of the pedestrian in a scene.}

\item{\textbf{Vehicle interaction context}: We call it \textit{vehicle context} in the paper. This context variable contains approximate longitudinal distance (depth) and velocity measures of the nearest vehicles to the pedestrian of interest.}
\end{itemize}

\subsubsection{Pedestrian location context}

Fig. \ref{fig:Context}(a-b) show a scene and its semantic segmentation. We utilize two segmentation categories, road pixels and non-road pixels. Road pixels are directly obtained from the pre-trained semantic segmentation software \cite{semantic-seg}, and all other pixels are marked as non-road pixels. We highlight 3 different regions around the pedestrian bounding box and their considered dimensions in Fig. \ref{fig:Context}c. We compute the mode of the refined semantic values (road or non-road) in each of the three regions, given by $\{m_1, m_2, m_3\}$, which constitute the location context feature vector $x_{sp,t}$.

\subsubsection{Vehicle interaction context}

Longitudinal distances (or depths) and velocities of approaching vehicles play a significant role in determining crossing intention of a pedestrian.

\vspace{1mm}

\noindent\textbf{Depth estimation:} As existing techniques for monocular depth estimation are not much reliable, we employ a pre-calibrated approach.

To estimate pedestrian and vehicle depths from camera, we pre-calibrate a scene by 10m, 20m and 30m depth lines (see Fig. \ref{fig:Context}d). We assume that once a camera is steadily mounted on a vehicle, these depth lines will approximately remain same during driving time, unless the road ahead has high slope\footnote{One of the reviewers kindly pointed out that the depth lines may vary with tyre pressure. We will look to replace with a better monocular depth estimation technique in future.}. The depth lines can be written as:
\vspace{1mm}
\begin{equation}
\label{eqn:st_line}
y = mx +c,
\end{equation}

\noindent where $(x,y)$ is a point co-ordinate on the image (in pixels). We assume that the slope $m$ and the intercept $c$ (in pixels) of the straight line have the following approximate exponential relations with depth $d$ (in meters) from camera:
\vspace{2mm}
\begin{equation}
\label{eqn:slope}
m(d) = k_m - p_m \cdot e^{{-} dl_m},
\end{equation}
\begin{equation}
\label{eqn:intercept}
c(d) = k_c - p_c \cdot e^{{-} dl_c},
\end{equation}

\noindent where $m(d)$ and $c(d)$ are known for $d$ $=$ 10, 20 and 30. $k_m, p_m, l_m, k_c, p_c$ and $l_c$ can be determined by solving equations \eqref{eqn:slope} and \eqref{eqn:intercept} using these 3 pairs of values of $\{m(d), c(d), d\}$. Once these parameters are obtained, depth of any given point $(x,y)$ in the image can be approximately determined from \eqref{eqn:st_line}, \eqref{eqn:slope} and \eqref{eqn:intercept}. We consider the bottom center pixels of the concerned pedestrian and vehicle bounding boxes to estimate their depths.

\vspace{1mm}

\noindent\textbf{Velocity estimation:} To obtain velocity measures of the nearest vehicles to a pedestrian in the NTU dataset, we compute mean optical flow \cite{Optical Flow} magnitude inside the corresponding vehicle bounding boxes. We utilize the provided ego-vehicle velocity labels in the JAAD dataset. At each instant, the ego-vehicle is assigned one of these velocity labels: `moving slow', `moving fast' or `standing'. Some instants only have the ego-acceleration labels (`slowing down' and `speeding up') and do not contain velocity labels. We associate velocity labels to these instants by imputation/observation. We ignore the ego-acceleration labels in this paper.

\vspace{1mm}

The complete (depth and velocity) vehicle context for the NTU dataset is given by, $x_{vNTU,t}$ = \{$ncl$, $ncr$, $nclf$, $ncrf$\}, where $ncl$ and $ncr$ are the relative depths (in meters) of the nearest vehicles to a pedestrian on left and right lane respectively and $nclf$, $ncrf$ are the mean optical flow of the non-pedestrian pixels within the bounding boxes of the respective vehicles. At instants with no influencing vehicle, we set $ncl$ and $ncr$ to -1 and $nclf$ and $ncrf$ to 0.

The complete vehicle context for the JAAD dataset is given by, $x_{vJAAD,t}$ = \{$ego\_dep$, $ego\_vel$\}, where $ego\_dep$ is the estimated pedestrian depth from the ego-vehicle and $ego\_vel$ is the ego-velocity attribute (`moving slow', `moving fast' etc.). As the other vehicles do not influence pedestrian behaviours in most cases within JAAD dataset, we keep things simpler by only considering the ego-vehicle context. We defer more complete interactions including other vehicles in the scene to our future work.

\subsection{Motion Features} \label{subsec:motion}

We propose a computationally inexpensive heuristic method to capture relevant pedestrian motion dynamics. The features obtained by this method perform comparably to one of the better benchmarks \cite{Daimler} on the Daimler dataset \cite{Daimler Dataset} (see Appendix \ref{Appen:Daim}). First, we apply a pre-trained object detector \cite{obj-det} to detect and track pedestrians in the videos\footnote{For considered pedestrians in the NTU data. We utilized provided bounding box annotations for pedestrians in the JAAD data.}. For extracting pedestrian motion features at $t$, we consider a sliding window of length $\tau$, i.e., frames $t-\tau+1$ to $t$. The feature extraction process involves three steps:

$\bullet$ \textbf{Pre-processing:} We run a Kalman filter on tracked pedestrian bounding boxes to remove noise/occlusions. Further occlusions (e.g., by tree) are linearly interpolated.

$\bullet$ \textbf{Real world lateral displacement estimate:} A pedestrian's crossing/not-crossing behaviour is primarily characterized by the real world lateral motion. Lateral motion captured in an image sequence contains components from both lateral and longitudinal pedestrian movements (w.r.t.~the ego-moving direction) in real world. In addition, the real world longitudinal movement is mixed with the ego-vehicle motion. Assuming negligible pedestrian movement along the vertical direction in real world, we obtain approximate estimates of the real world lateral motion in image frame using the camera matrix information (see Appendix \ref{Appen:long-motion}).

$\bullet$ \textbf{Fitting non-linear dynamics:} We extract real world lateral displacement estimates for 9 symmetrical points inside the pedestrian bounding box, given by $\{v_p\}$, $p$ = $1:9$ (see Fig. \ref{fig:Context}e). The choice of the points is arbitrary, as long as they are well distributed within the bounding box. Each of these points is tracked and processed over the sliding window to generate $\tau - 1$ lateral displacement values $v_{p,t_1}$, $t-\tau + 2 \leq t_1 \leq t$. Finally, we fit a degree 2 curve to the values $\{{v}_{p,t-\tau+2:t}\}$, $p$ = $1:9$, by least-squares fitting:
\vspace{1mm}
\begin{equation}
\label{eqn-non-linear}
v_{p,t_1} = {a_0}_{p,t} + {a_1}_{p,t}t_1 + {a_2}_{p,t}t_1^2 , 
\end{equation}

\noindent resulting in a 27 dimensional feature set, $x_{m,t} = \{\{{a_0}_{p,t}\}_{p = 1:9}, \{{a_1}_{p,t}\}_{p = 1:9}, \{{a_2}_{p,t}\}_{p = 1:9}\}$. $\tau$ must be large enough ($\ge$4) for a good fitting but not too large to include unnecessary past frames in the current dynamics computation. We set $\tau = 10$, but for test sequences with less than 10 frames, we adjust $\tau$ accordingly to generate the motion features. We also tested optical flow features and visual representations as in another work from our group \cite{Michael}. However, including such features resulted in poorer performance on our datasets. Thus we do not include such features in this paper. We plan to test such visual representations using end-to-end models in future.

\section{Datasets} \label{sec:dataset}

We evaluate on an in-house collected NTU dataset and the public real-life JAAD dataset \cite{JAAD}.

\begin{figure}[h]
\includegraphics[scale=0.45]{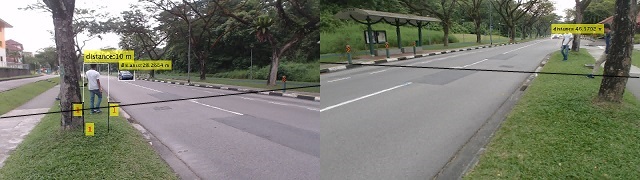}
\centering
\caption{NTU dataset sample. Camera1 (left) and Camera2 (right) together capture the whole scene of interest.}
\label{fig:NTU_dataset_sample}
\end{figure}

The NTU dataset is captured inside Nanyang Technological University Campus by a pair of static cameras at 30 fps framerate and $1920 \times 1080$ resolution. Since our aim is to capture natural vehicle influences on pedestrian intention, we place two synchronized cameras to capture the whole scene of interest, so that approaching vehicles on both sides of the pedestrian can be captured, as depicted in Fig. \ref{fig:NTU_dataset_sample}. In addition, Camera1 is placed in a way to simulate a camera mounted on a stationery ego-vehicle.

In the videos, we had actors and actual pedestrians strolling in a natural road scene with vehicles. Actors were not given any specific instructions and were asked to move around the scene naturally. We extract 35 continuous crossing and 35 stopping sequences from the videos for our evaluation.

The JAAD dataset \cite{JAAD} is a recent pedestrian benchmark which captures real-world pedestrian behaviours in complex scenes from cameras mounted on moving vehicles. It contains 346 video clips at 1920 $\times$ 1080 resolution (few at 1280 $\times$ 720) and 15 fps framerate, each containing multiple pedestrian sequences. All pedestrians of interest are annotated by bounding boxes and are assigned certain movement labels (moving, crossing, stopping, standing, etc.) and personal details (age, sex, movement direction, etc.). Each frame in the videos is also associated with certain traffic attributes, such as zebra crossing, parking lot, etc.

To our best knowledge, there is no JAAD data subset proposed for early pedestrian intention prediction. We extract a total of 120 pedestrian sequences distributed across continuous crossing, stopping, starting, and standing scenarios (see Appendix \ref{Appen:JAAD} for sequence details).

\section{Experimental Setup} \label{sec:exp-setup}

We describe here our experimental data, extracted features and labels and the model specifications.

\vspace{2mm}
\noindent \textbf{Data for evaluation:} The distribution of test sequences in the two datasets is presented in Table \ref{table:sequence_dist}.

We downsampled the NTU dataset sequences to 15 fps. We detect and track \cite{obj-det} pedestrians in Camera1 images. We also define pedestrian \textit{spatial context} only in Camera1 images. Vehicles are detected and tracked in both Camera1 and Camera2 images. We eliminate ambiguity within the common regions of Camera1 and Camera2 by a pre-defined separator (marked by black lines, see Fig. \ref{fig:NTU_dataset_sample}). The object detector could efficiently detect pedestrians and vehicles up to 80 meters from the camera. For pedestrian sequences in the JAAD data, we utilize provided bounding box annotations. 
\vspace{2mm}
\begin{table}[h]
\caption{Table showing distribution of test sequences in different datasets.}
\label{table:sequence_dist}
\begin{center}
\renewcommand{\arraystretch}{1.2}
\setlength\tabcolsep{1.7pt}
\begin{tabular}{|c |c | c| c| c| c|} 
\hline
\backslashbox{Dataset}{Seq type} & \shortstack{Continuous \\ crossing} & Stopping & Starting & Standing & \shortstack{Total \#\\ instances} \\[0.4ex] 

\hline\hline
NTU & 35 & 35 & - & - & 9276 \\
\hline
JAAD & 45 & 20 & 40 & 15 & 19650 \\
\hline

\end{tabular}
\end{center}
\vspace{-4mm}
\end{table}

\begin{figure}[h]
\includegraphics[scale=0.28]{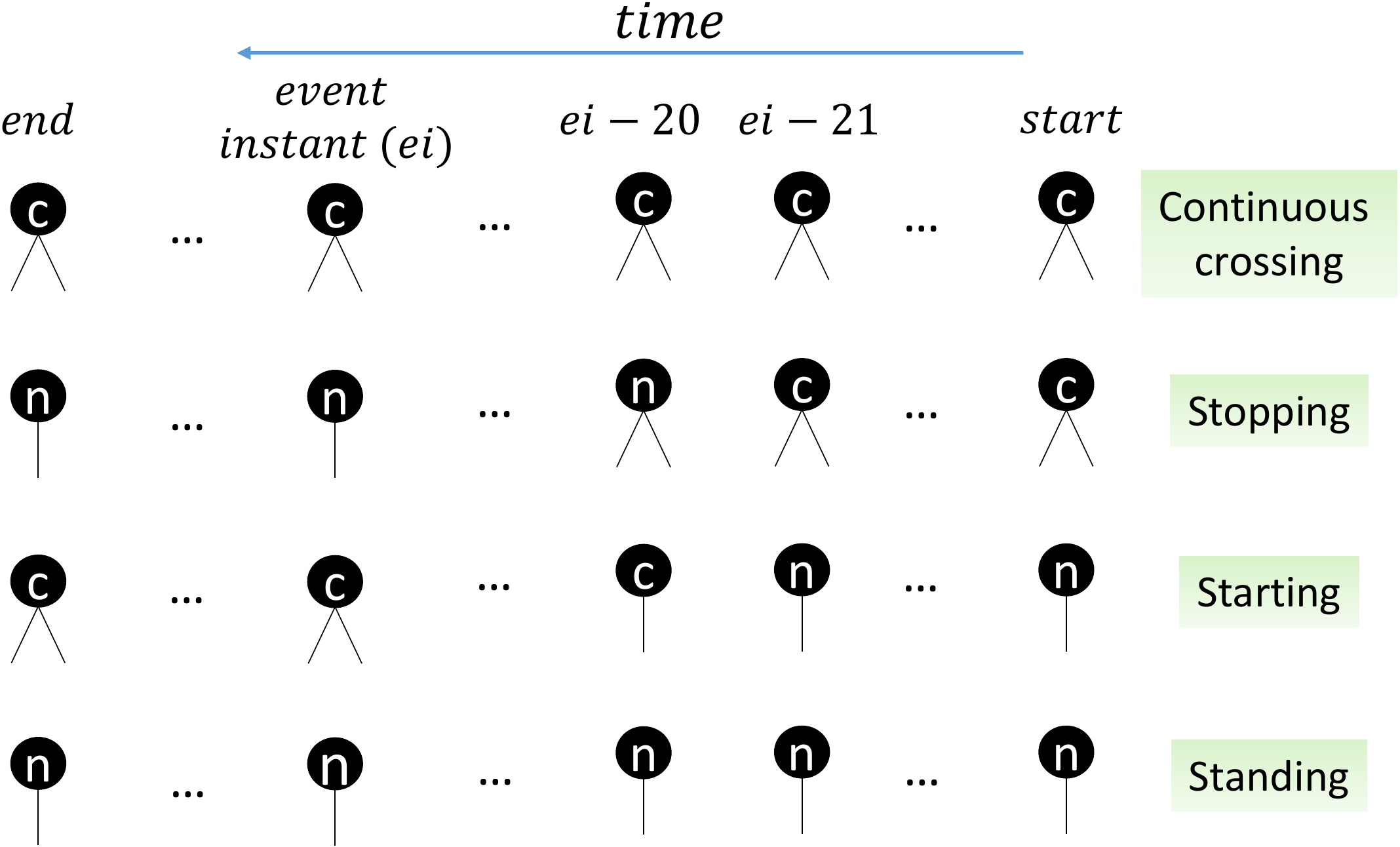}
\centering
\caption{Labeling the training sequences in the JAAD. `c' denotes `crossing' and `n' denotes `not-crossing' intention. For the JAAD data, \textit{pred\_ahead} = 20. The NTU data follows similar labeling with \textit{pred\_ahead} = 30.}
\label{label_jaad}
\end{figure}

\vspace{1mm}
\noindent \textbf{Features and labels:} In the NTU data, we try to predict the stopping event 30 frames (2 sec) before the \textit{stopping instant}, i.e., \textit{pred\_ahead} = 30. Therefore, we label data from 30 frames before the \textit{stopping instant} in stopping sequences as `not-crossing' during training. Remaining frames in stopping sequences and all frames in continuous crossing sequences are labeled `crossing'. The annotations are depicted in Fig. \ref{label_jaad}. We do not have standing and starting sequences for the NTU data. The complete feature set (see Section \ref{sec:features-labels}) on the NTU data is given by, $x_{t,NTU} = \{x_{m,t}, x_{sp,t}, x_{vNTU,t}\}$. 

For JAAD dataset, we set \textit{pred\_ahead} = 20 (1.33 sec). The labeling is illustrated in Fig. \ref{label_jaad}. Considered feature set on the JAAD data is given by, $x_{t,JAAD} = \{x_{m,t}, x_{sp,t}, x_{vJAAD,t}\}$.

\vspace{1mm}
\noindent \textbf{Models compared:} We utilized FLDCRF single-label variant and LSTM as our learning models. We show results for three different systems for both the NTU and JAAD data:

1) Only pedestrian motion based, referred as $mtn$ (or $m$) in the Results section. This system only utilizes pedestrian motion as input features, i.e., $x_t$ = \{$x_{m,t}$\}.

2) Pedestrian motion and location based, referred as $mtn+spa$ (or $ms$). This system takes pedestrian's semantic location (see Section \ref{subsec:context}) along with pedestrian motion as inputs, i.e., $x_t$ = \{$x_{m,t}, x_{sp,t}$\}.

3) Pedestrian motion, location and vehicle context based, referred as $mtn+spa+veh$ (or $msv$). This system considers vehicle context features (see Section \ref{subsec:context}) together with pedestrian motion and location, i.e., $x_t$ = \{$x_{m,t}, x_{sp,t}, x_{v,t}$\}.

We employ a nested cross-validation (with 5 outer folds and 4 inner folds) for selecting models and generating results. For selecting LSTM models, we tested the following number of hidden units ($HU$): 2, 5, 10, 20, 50, 150, 300, 500. We trained each model for 1000 epochs, saving model performances at every 100 epochs during validation. We fed all training sequences at each epoch at the rate of 1 sequence per batch.

An FLDCRF single-label has two hyper-parameters: number of layers ($num\_layer$) and number of hidden states per label ($\{num\_state_i\}$) along layers $i= 1:L$. For simplicity, we keep same $num\_state$ along all layers, i.e. $num\_state_i = num\_state$, $\forall i = 1:L$. We denote such model by FLDCRF-$<$$num\_layer$$>$/$<$$num\_state$$>$. We selected our models from the following $<$$num\_layer$$>$/$<$$num\_state$$>$ combinations: 1/1, 1/2, 1/3, 1/4, 1/5, 1/6, 2/1, 2/2, 2/3. 

We trained FLDCRF model parameters with Stan's \cite{Stan} in-built BFGS optimizer. We implemented the LSTM models defined in Keras, a deep learning library for Python running Tensorflow in the backend. LSTM parameters were trained by a default Adam optimizer in Keras. We plotted the results in MATLAB 2015b \cite{MATLAB}. LSTM models are trained with a Nvidia Tesla K80 GPU, while the FLDCRF models are trained with a Intel Xeon E5-1630 v3 CPU.

\section{Results} \label{sec:results}

We compare different systems on the NTU and JAAD datasets using existing state-of-the-art (LSTM) and proposed (FLDCRF) sequential models. We evaluate the models by two metrics: stopping probability vs time and classification accuracy vs time. Stopping probability corresponds to the probability of the `not-crossing' (laterally static) label.

We ideally want high classification accuracy of appropriate class labels (e.g., `not-crossing' in stopping sequences) over the prediction interval. At the same time, it is desired to have high and stable average probability values of appropriate class labels over the prediction interval. It is also preferred that the sequences of same type have minor standard deviation in predicted probability values at each instant. So, our model hyper-parameters are chosen based on the following metric ($mt$), 
\vspace{1mm}
\begin{equation}
\label{eqn:metric}
mt = \frac{1}{T_p} \sum_{t=T_l.f}^{T_u.f} (acc_t + \frac{1}{N_v} \sum_{i=1}^{N_v} prob_{i,t} - \frac{1}{N_s} \sum_{j=1}^{N_s} std_{j,t}),
\end{equation}

\noindent where $acc_t$ denotes overall classification accuracy, $prob_{i,t}$ denotes predicted probability of the appropriate class (e.g., `not-crossing' in stopping sequences) for sequence $i$ and $std_{j,t}$ is standard deviation in probability values among sequences of type $j$ at $t$. $f$ is the dataset framerate (15 fps for both datasets). The prediction interval is given by $[T_l \: T_u]$s and $T_p$ = ($\mid$$T_u.f - T_l.f$$\mid$ + 1). $N_v$ is the number of considered validation sequences and $N_s$ is the number of considered sequence types (2 for NTU and 4 for JAAD data).

\subsection{NTU dataset}

\begin{figure}[h!]
\centering
\subfloat[\scriptsize{Stopping probability vs time.}]{\label{fig:Stopping-a}
\includegraphics[scale=0.45, height=3.5cm]{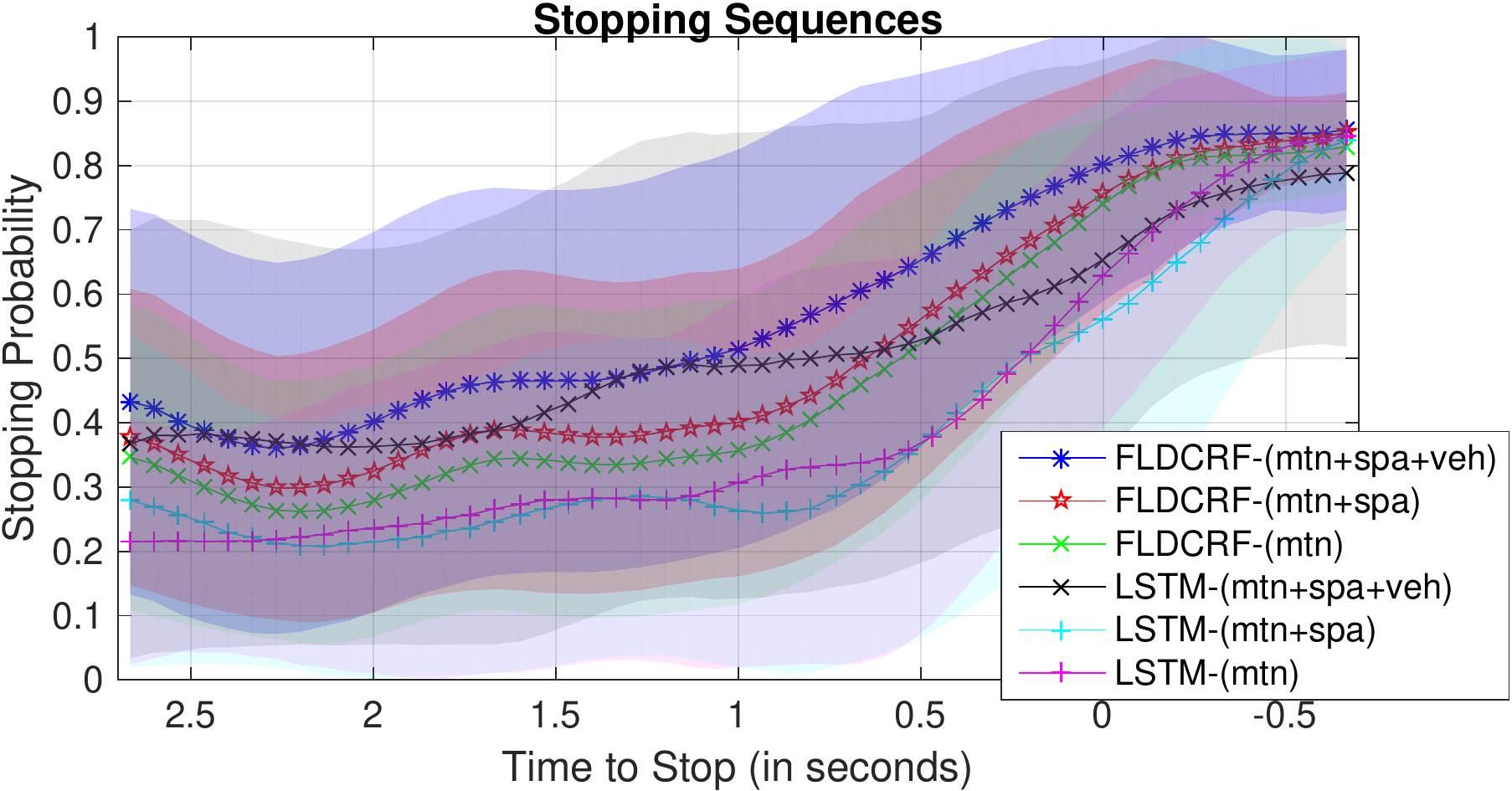}} \\[-0.2ex]
\vspace{-2.2mm}
\subfloat[\scriptsize{Classification accuracy vs time.}]{\label{fig:Stopping-b}
\includegraphics[scale=0.45, height=3.5cm]{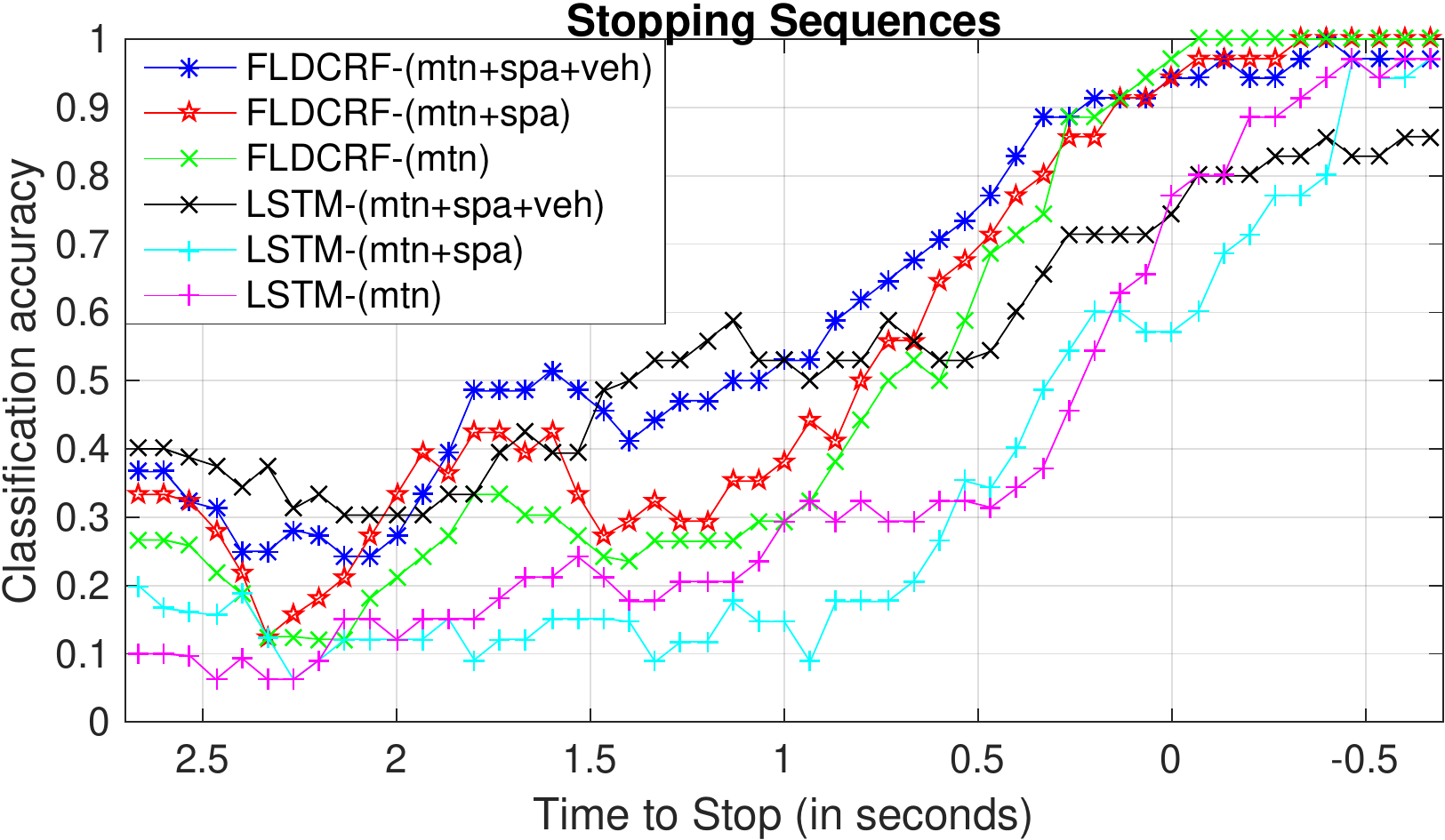}} \\[-0.2ex]

\caption{Performance of different systems on stopping sequences of NTU dataset.}
\label{fig:NTU stopping}
\end{figure}

\begin{figure}[h!]
\centering
\subfloat[\scriptsize{Stopping probability vs time.}]{\label{fig:Stopping-con-a}
\includegraphics[scale=0.45, height=3.5cm]{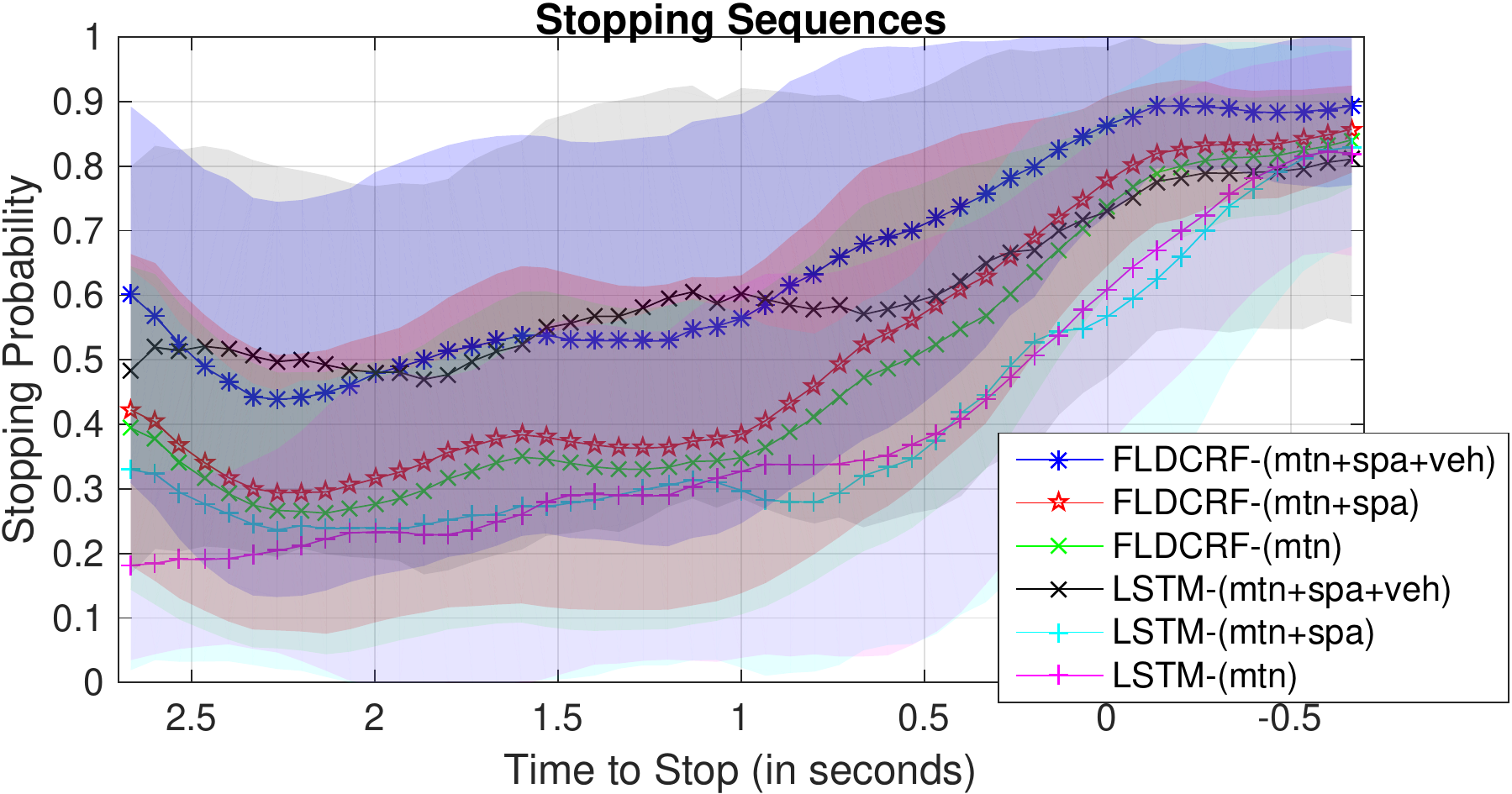}} \\[-0.2ex]
\vspace{-2.2mm}
\subfloat[\scriptsize{Classification accuracy vs time.}]{\label{fig:Stopping-con-b}
\includegraphics[scale=0.45, height=3.5cm]{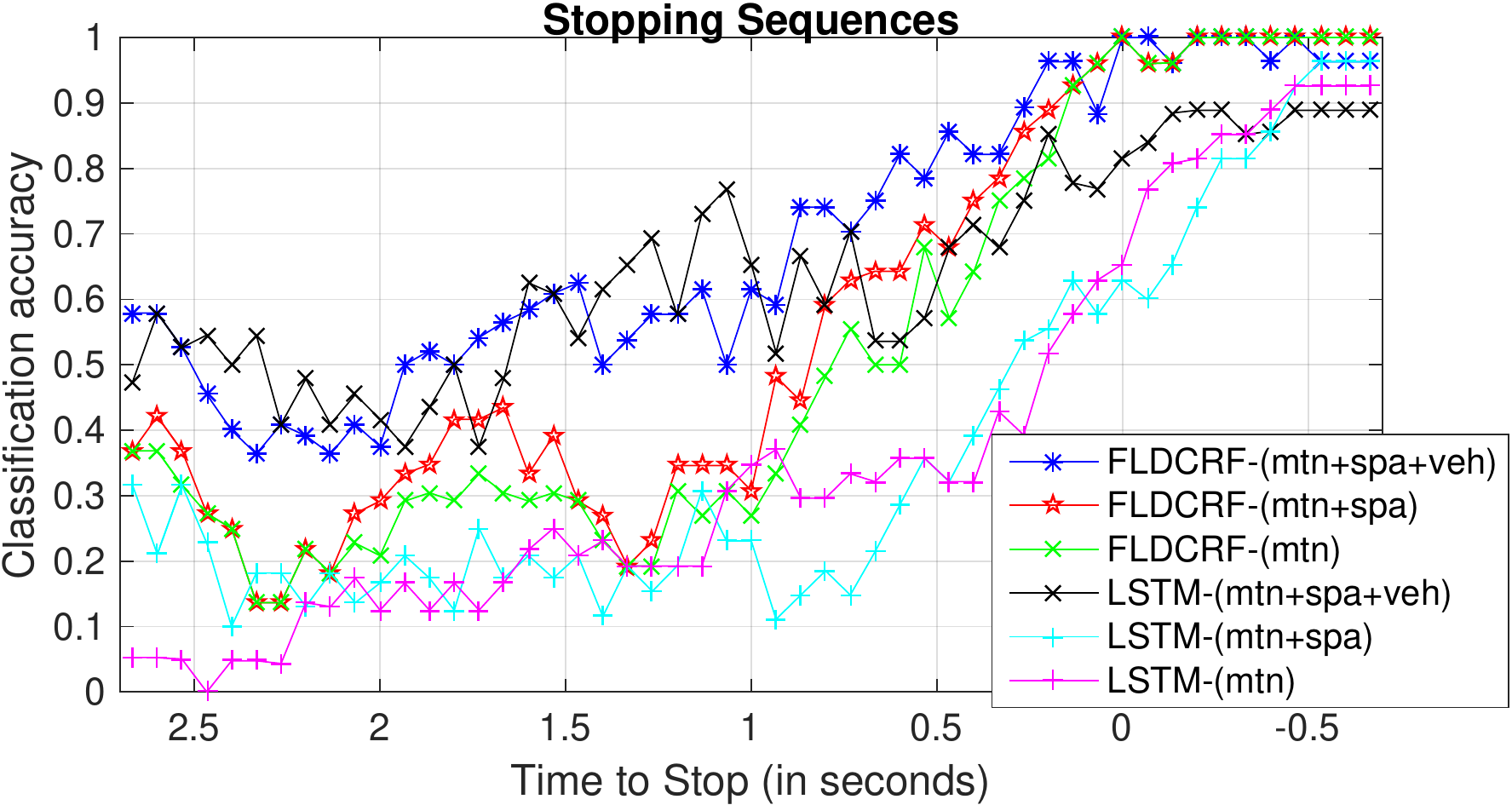}} \\[-0.2ex]

\caption{Performance for stopping sequences with positive vehicle context in the NTU dataset.}
\label{fig:NTU stopping context}
\end{figure}

\begin{figure}[h!]
\centering
\subfloat[\scriptsize{Stopping probability vs time.}]{\label{fig:Crossing-a}
\includegraphics[scale=0.45, height=3cm]{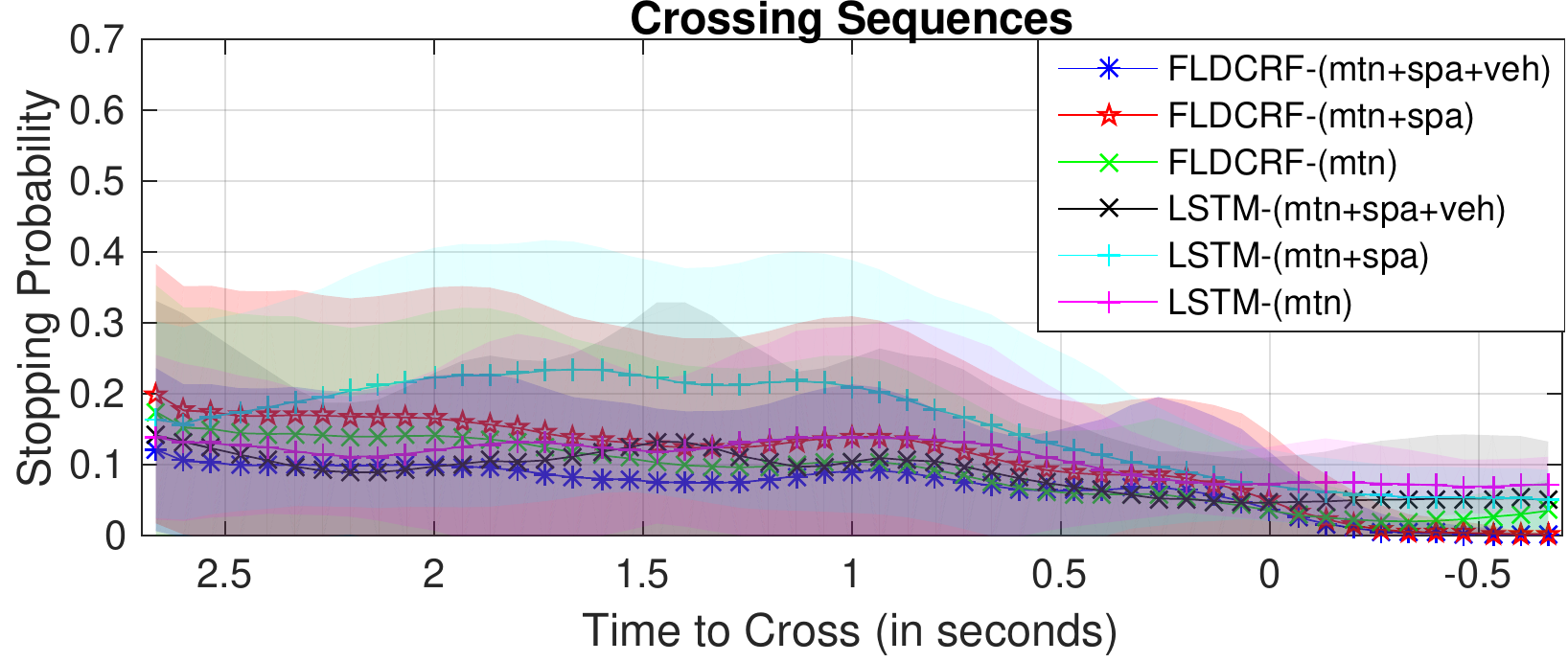}} \\[-0.2ex]
\vspace{-2.2mm}
\subfloat[\scriptsize{Classification accuracy vs time.}]{\label{fig:Crossing-b}
\includegraphics[scale=0.45, height=2.2cm]{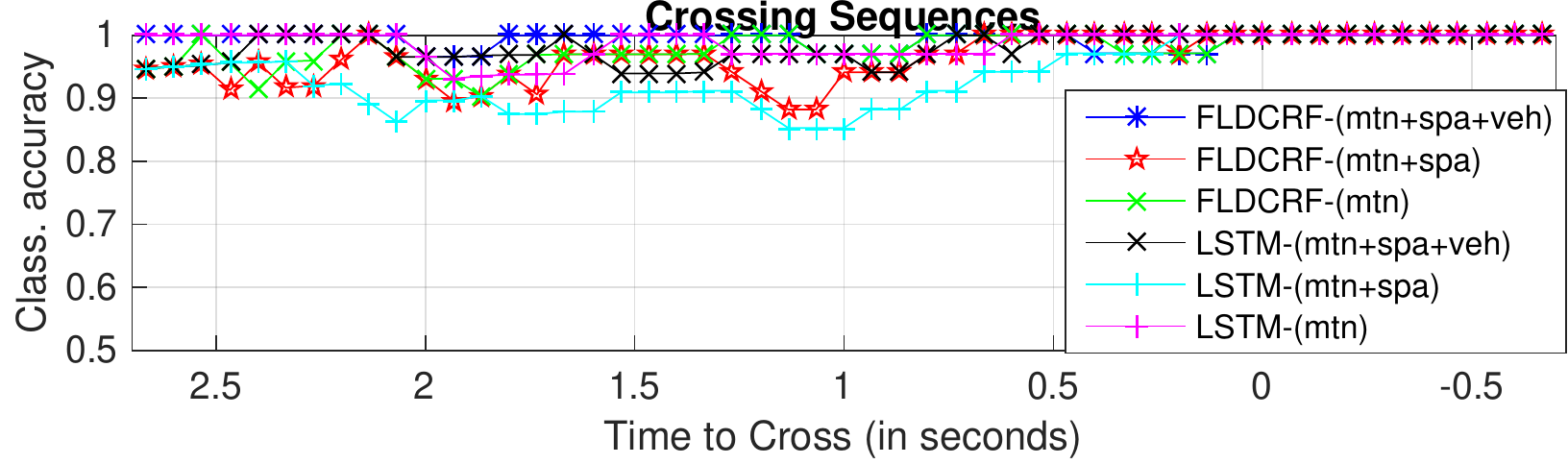}} \\[-0.2ex]

\caption{Results on continuous crossing sequences of the NTU dataset.}
\label{fig:NTU crossing}
\end{figure}

\begin{figure}[h]
\includegraphics[scale=0.45, height=3cm]{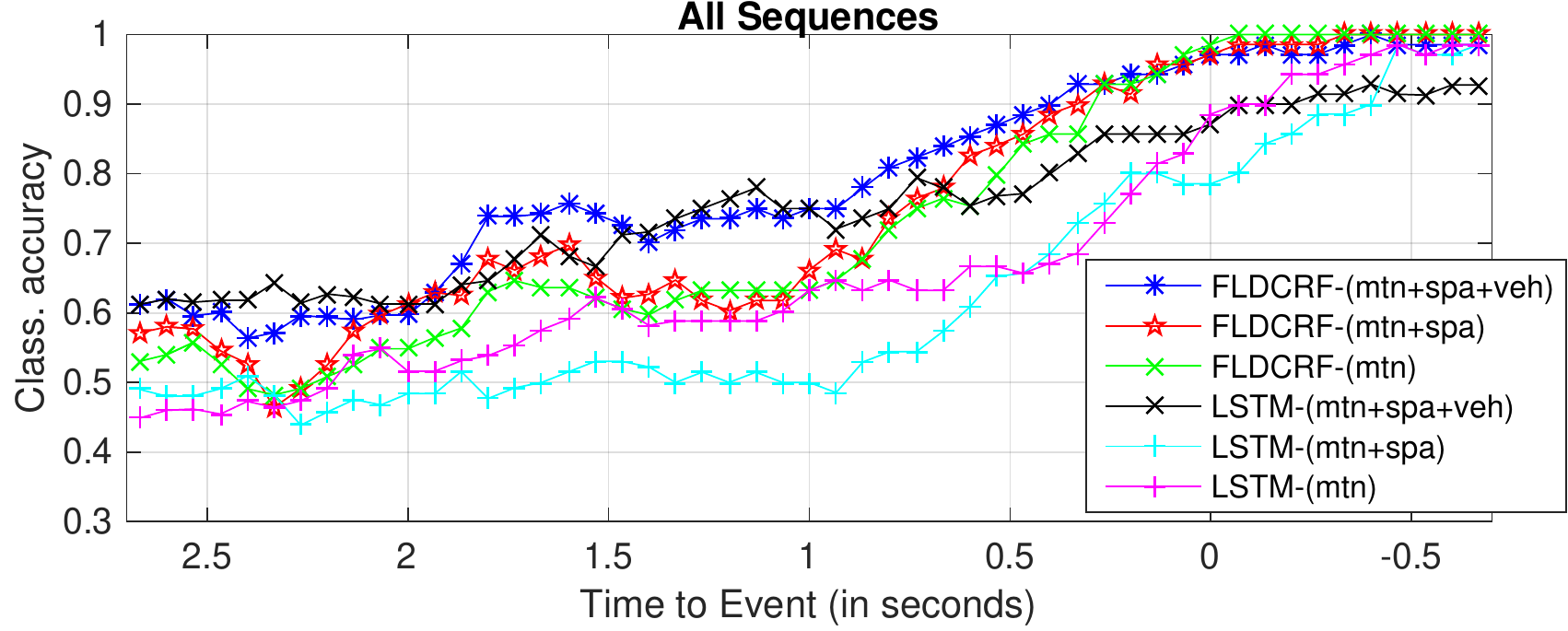}
\centering
\caption{Overall classification accuracy of different systems on the NTU dataset.}
\label{fig:all-NTU}
\end{figure}

\begin{figure}%
\centering
\subfloat[\scriptsize{Continuous crossing sequences.}]{{\includegraphics[width=4.15cm]{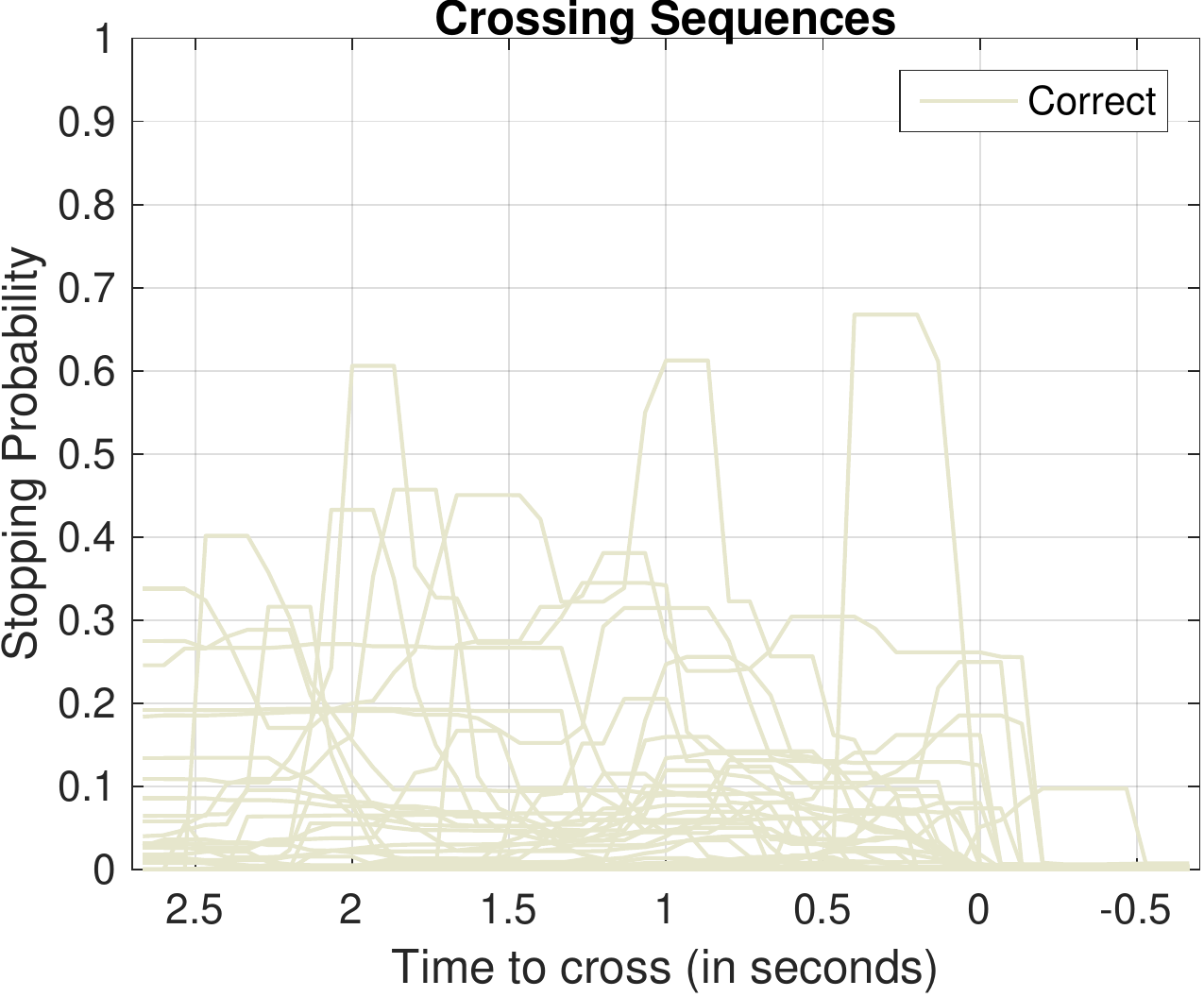} }}%
\subfloat[\scriptsize{Stopping sequences.}]{{\includegraphics[width=4.15cm]{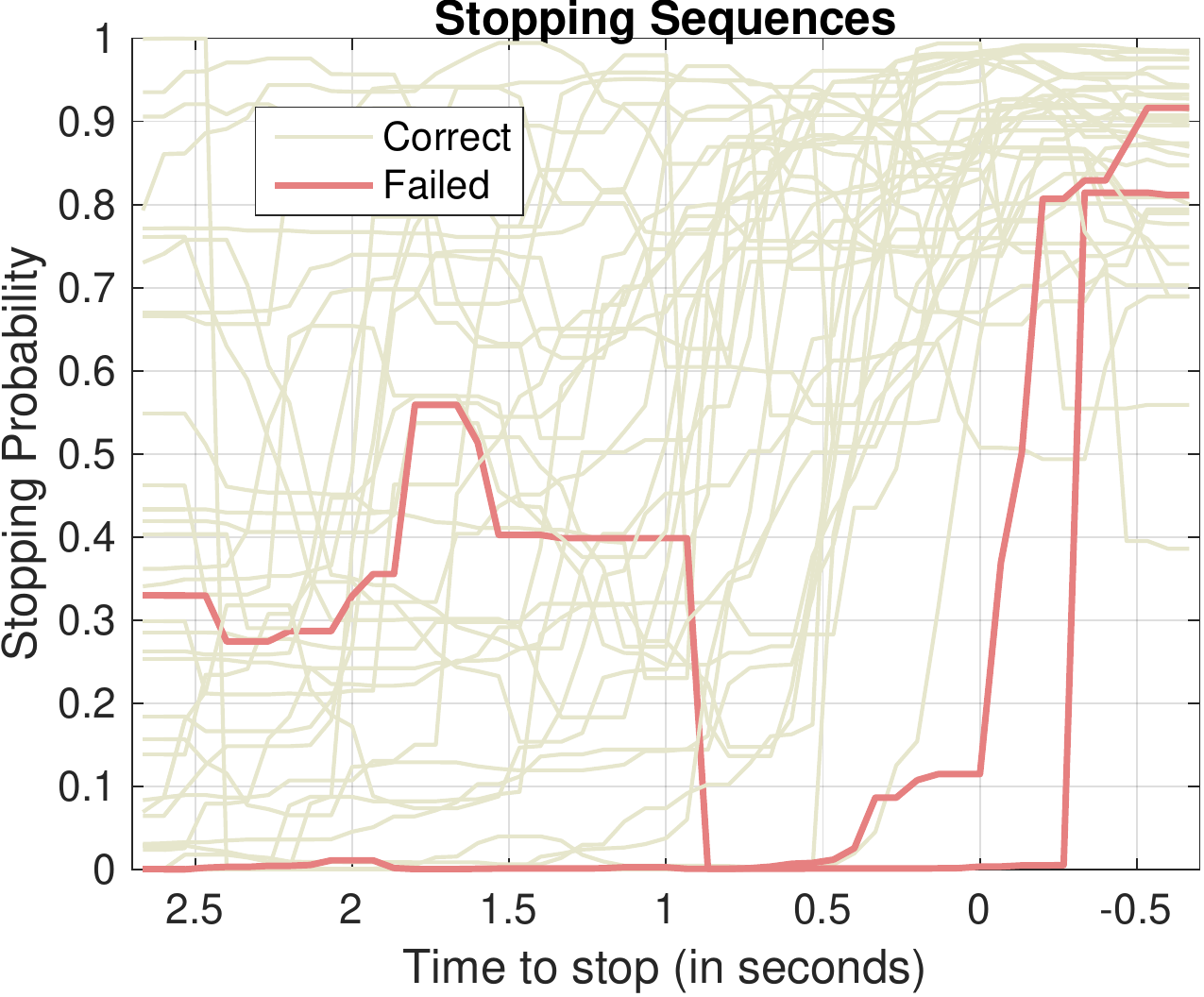} }}%
\caption{Individual sequence outputs by $FLDCRF-msv$ on the NTU dataset. Failed cases are shown in bold red.}%
\label{fig:NTU-failed}%
\end{figure}

We compare three different systems - $mtn$ (\textit{m}), $mtn+spa$ (\textit{ms}) and $mtn+spa+veh$ (\textit{msv}) with FLDCRF and LSTM on the NTU dataset. `0' along x-axis (see Fig. \ref{fig:NTU stopping} - \ref{fig:all} and Fig. \ref{fig:JAAD-failed}) represents the event (\textit{crossing}, \textit{stopping} etc.) instants. Positive and negative values along x-axis indicate gain and loss in prediction time (in seconds) respectively. Since we try to predict 2 sec (30 frames) ahead of the event, we will limit our analysis within [2 -0.5] s prediction interval. 

Model hyper-parameters were chosen by considering the value of the metric in \eqref{eqn:metric} within [2 -0.5] s interval.  Table \ref{table:system-select} presents a comparison between LDCRF (FLDCRF with 1 hidden layer), FLDCRF and LSTM over nested validation sets on $mtn+spa+veh$ (\textit{msv}) systems. Two layered FLDCRF settings (2/1, 2/2, 2/3) performed better than the 1 layered settings (1/1, 1/2, 1/3, 1/4, 1/5, 1/6) within most of the inner validation sets. The new models (2/1, 2/2, 2/3) also help to enhance the overall FLDCRF performance. LSTM models perform worse than the FLDCRF models on the validation sets in general. We select best performing FLDCRF and LSTM settings for each of the three systems on the inner validation sets and average their performance on the outer folds to obtain our results.

\begin {table}[h!]
\vspace{3mm}
\caption{Comparison between FLDCRF, LDCRF and LSTM over nested validation sets on $mtn+spa+veh$ system (on the NTU data).}
\label{table:system-select}
\begin{center}
\renewcommand{\arraystretch}{1.2}
\setlength\tabcolsep{1.7pt}
\begin{tabular}{|c |c | c| c| c| c| c|} 
\hline
\backslashbox{Model}{Valid. set} & 1 & 2 & 3 & 4 & 5 & \shortstack{Average \\ performance} \\ [0.4ex] 

\hline\hline
FLDCRF-1layer & 1.4324 & 1.4125 & \textbf{1.4447} & \textbf{1.4458} & 1.4436 & 1.4358\\
\hline
FLDCRF-2layers & \textbf{1.4330} & \textbf{1.4199} & \textbf{1.4447} & 1.4419 & \textbf{1.4439} & 1.4366\\
\hline
FLDCRF-overall & \textbf{1.4330} & \textbf{1.4199} & \textbf{1.4447} & \textbf{1.4458} & \textbf{1.4439} & \textbf{1.4374}\\
\hline
LSTM & 1.3315 & 1.3355 & 1.3077 & 1.4085 & 1.3706 & 1.3508\\
\hline

\end{tabular}
\end{center}
\vspace{-3mm}
\end{table}

Fig. \ref{fig:NTU stopping} displays the performances of different systems on 35 stopping sequences of the NTU dataset. Motion-only FLDCRF model ($FLDCRF-m$) has consistently low average stopping probability (see Fig. \ref{fig:NTU stopping}a) and classification accuracy (see Fig. \ref{fig:NTU stopping}b) values than models with context in the earlier regions of the curves (before `0'). Adding location context to FLDCRF ($FLDCRF-ms$) slightly improves time, probability and accuracy of predicting the stopping intention in earlier regions. The results improve significantly with higher average stopping probability and accuracy on adding vehicle context features to the FLDCRF model $\left(FLDCRF-msv\right)$. On average, all LSTM models perform worse than the corresponding FLDCRF models in terms of prediction time, accuracy and probability values. 

Fig. \ref{fig:NTU stopping context} compares different models on stopping scenarios aided by positive vehicle context ($ncl\ge0$ and/or $ncr\ge0$). At each instant of the plots, only frames with such vehicle context values are considered. While we find models without vehicle context ($mtn$ and $mtn+spa$) to perform similarly to the earlier results, models with vehicle context ($mtn+spa+veh$) display an enhanced prediction performance with better average probability and accuracy in earlier regions. An average stopping probability of 0.6 and a classification accuracy of 0.7 was achieved and consistently improved from $\sim$0.9 seconds before the actual stopping events by $FLDCRF-msv$ system. The vehicle context aids a gain of $\sim$0.5 seconds in prediction time over $FLDCRF-ms$ system. $LSTM-msv$ performance also improves in these scenarios, but the accuracy values still fail to approach 1 (like FLDCRF models) in the vicinity of the stopping event.

Different systems are compared in Fig. \ref{fig:NTU crossing} on 35 continuous crossing sequences of the NTU dataset. Accuracy-wise, all $mtn$ and $mtn+spa+veh$ type models display quite consistent and accurate (close to 1) performance across time. Although, joint models ($mtn+spa+veh$) output smaller average probability values indicating better reliability of the system. LSTM and FLDCRF joint ($mtn+spa+veh$) models perform comparably across time, with FLDCRF type model exhibiting marginally better probability and accuracy values at certain time points.

Overall accuracy of considered systems at different prediction horizons on NTU dataset is displayed in Fig. \ref{fig:all-NTU}. `0' in the figure represents event (\textit{stopping} and \textit{crossing}) instants of respective sequences. $FLDCRF-msv$ system proves to be the most accurate at majority of early prediction instants, while performing comparable to certain systems ($FLDCRF-ms$ and $FLDCRF-m$) after the `0' mark, bolstering itself to be selected as the primary choice for an early and accurate intention prediction system on the NTU data. 

\vspace{2.5mm}
\noindent \textbf{Failed case analysis:}

Fig. \ref{fig:NTU-failed} shows individual sequence outputs by the $FLDCRF-msv$ system. We highlight sequences (in bold red) where the system fails to make early prediction (i.e., before `0') of the events. 

All crossing events were predicted correctly before respective \textit{crossing instants} (see Fig. \ref{fig:NTU-failed}a). An unwanted spike can be observed on sequence $crossing\_17$ near the `0' mark, caused by an approaching vehicle. However, the system was able to correct the prediction before the \textit{event instant} to avoid any critical failure. 

Individual probability outputs of 35 stopping sequences by $FLDCRF-msv$ system are displayed in Fig. \ref{fig:NTU-failed}b. The system fails to make early prediction (before `0') of the stopping event in the two highlighted sequences (by bold red), $stopping\_9$ and $stopping\_32$. 

$FLDCRF-msv$ system performs the earliest and most accurate predictions of the stopping behaviour among considered systems and provides better reliability on crossing sequences at the same time, indicating the power of including vehicle context and better performance of FLDCRF over LSTM on the NTU dataset. Next, we investigate our approach on real-life pedestrian behaviour prediction.

\subsection{JAAD dataset}

On the JAAD dataset, we evaluate on all four types of sequences: 45 continuous crossing, 20 stopping, 40 standing and 15 starting (see Section \ref{sec:nomenclature}). 


We compare results from FLDCRF and LSTM on three different systems: $mtn$ (\textit{m}), $mtn+spa$ (\textit{ms}) and $mtn+veh$ (\textit{mv}). Since the JAAD data contains variety of scenarios (some during night-time), semantic segmentation \cite{semantic-seg} performance is not consistent. This results in $mtn+spa$ models performing slightly worse than only $mtn$ models. Therefore, we omit pedestrian location information from our final model, which takes pedestrian motion ($x_{m,t}$) and vehicle context ($x_{vJAAD,t}$) features.

Similar to the NTU data, we obtain our test results from a 5-fold nested CV. We make use of the same metric in \eqref{eqn:metric}, within [1.33 -1] s interval, to select hyper-parameter settings on inner validation sets. FLDCRF (1.3640) improves over the LDCRF (1.3627) average performance on the validation sets.

Due to space limitation, we only present classification accuracy vs time by different systems on JAAD sequences. We refer to Appendix \ref{Appen:JAAD} for stopping probability vs time curves. 

\begin{figure}[h]%
\centering
\subfloat[\scriptsize{Continuous crossing sequences.}]{{\includegraphics[width=4.1cm]{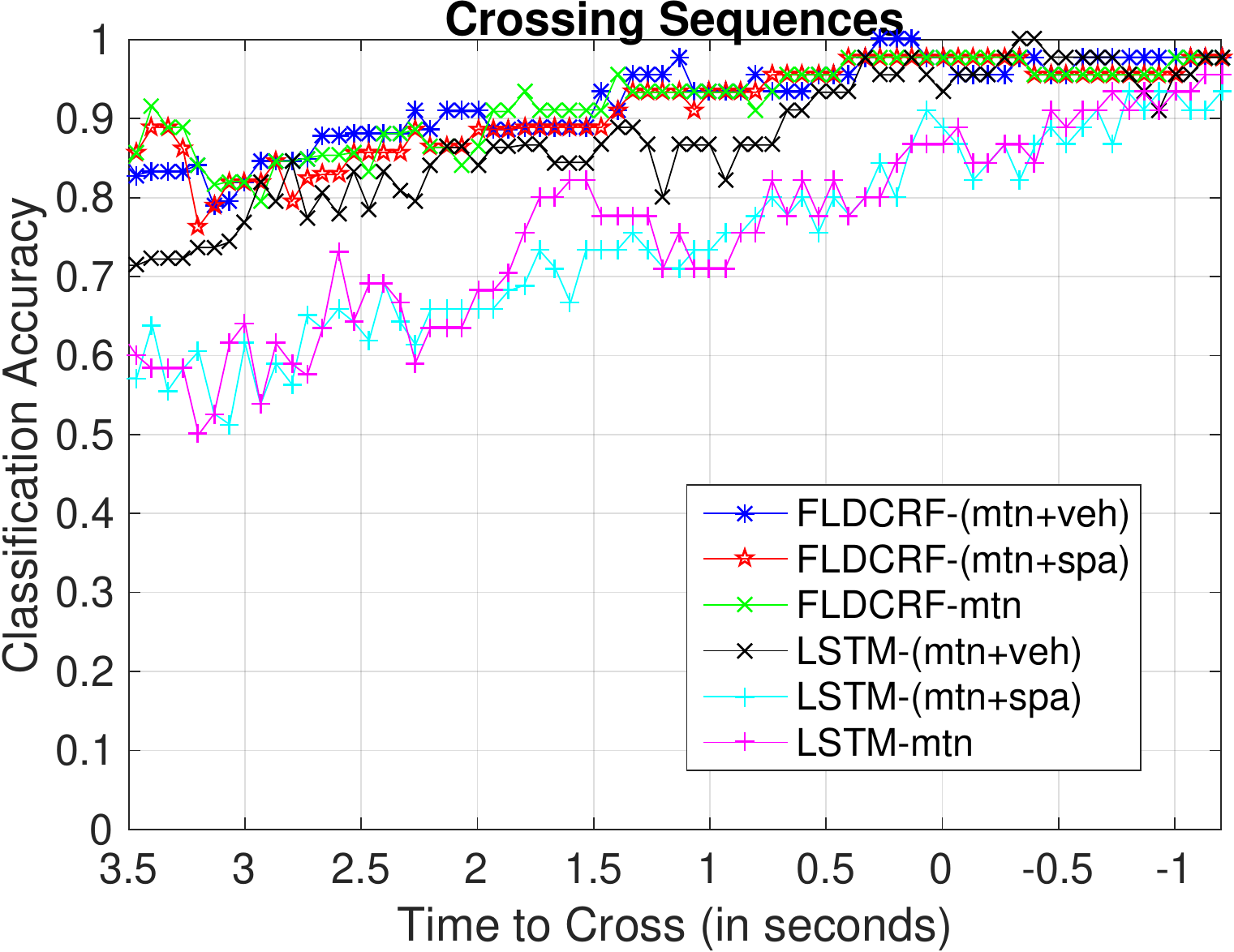} }}%
\quad
\subfloat[\scriptsize{Stopping sequences.}]{{\includegraphics[width=4.1cm]{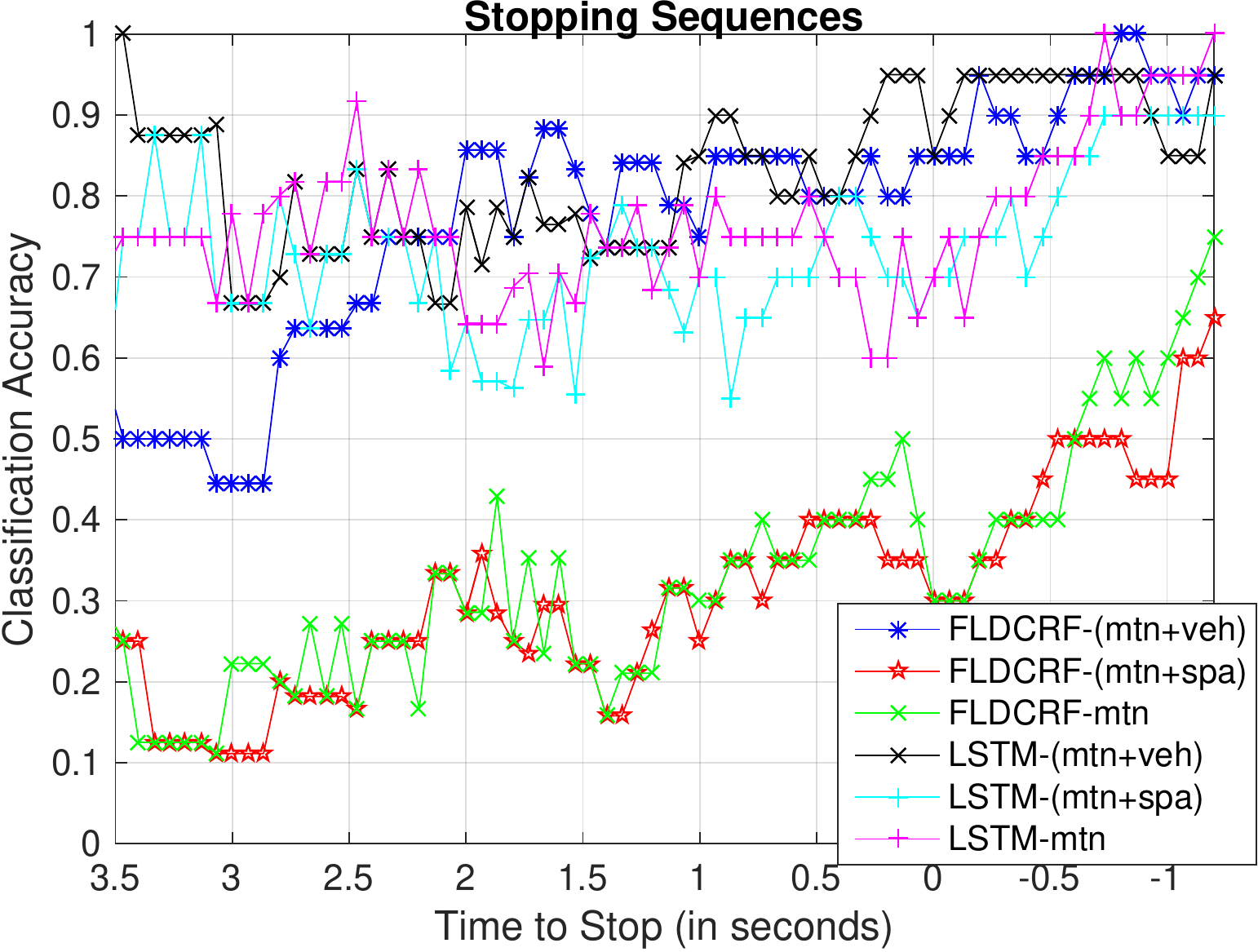} }}%
\quad
\subfloat[\scriptsize{Starting sequences.}]{{\includegraphics[width=4.1cm]{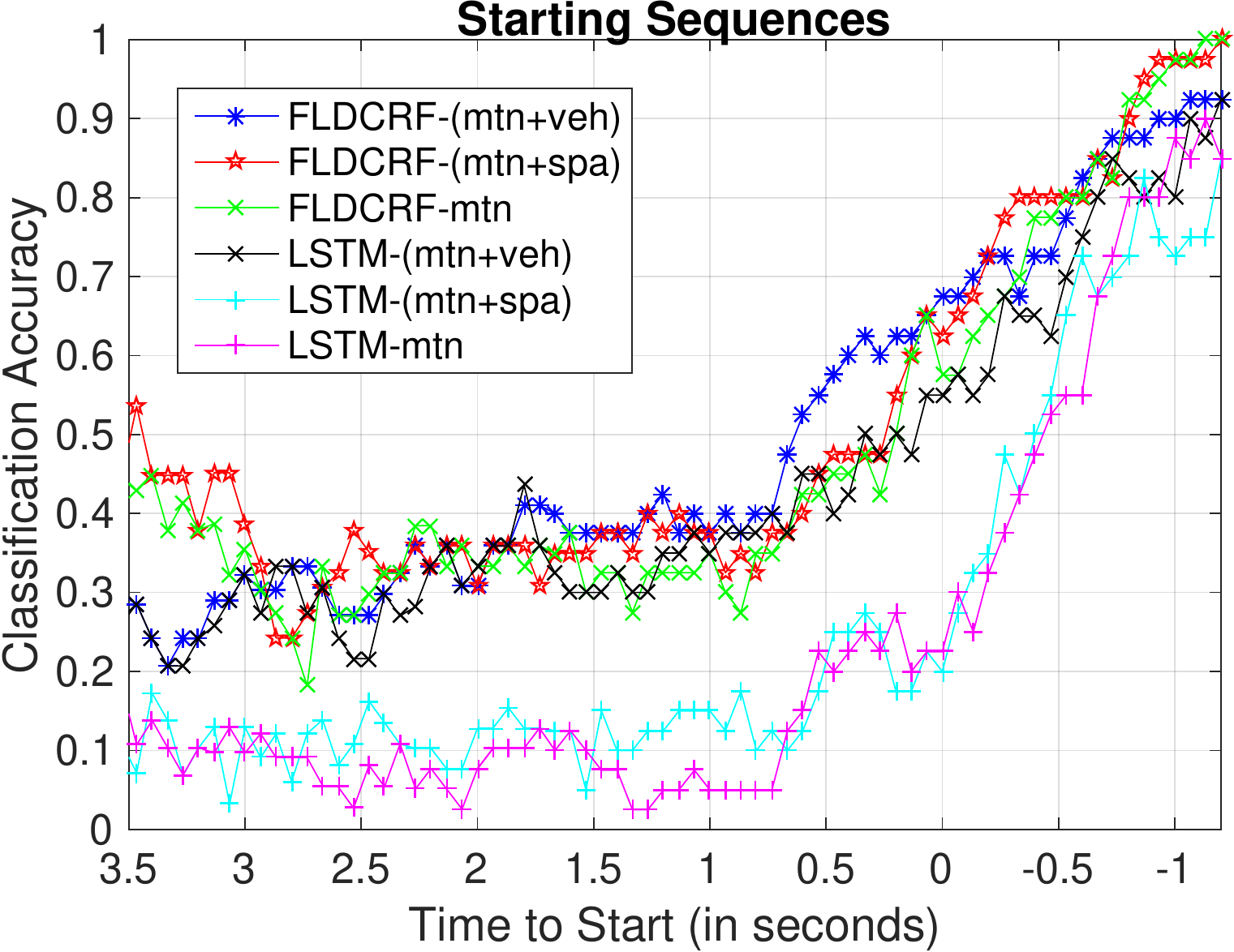} }}%
\quad
\subfloat[\scriptsize{Standing sequences.}]{{\includegraphics[width=4.1cm]{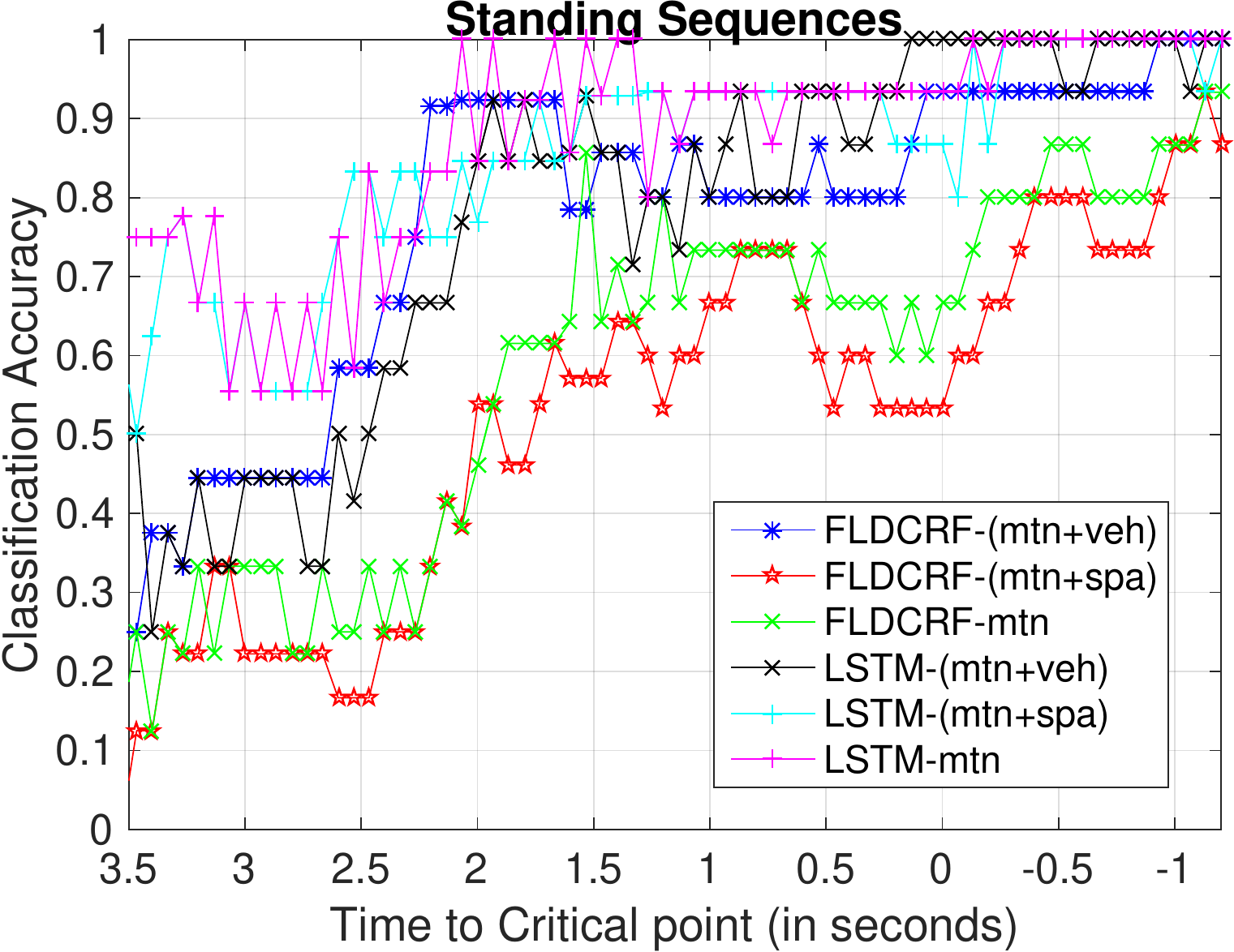} }}%
\caption{Accuracy performances of different systems vs time on the JAAD data.}%
\label{fig:JAAD-accuracy}%
\end{figure}

Fig. \ref{fig:JAAD-accuracy}a compares different systems on JAAD continuous crossing sequences. FLDCRF systems produce comparable and stable performance with more than 95\% average accuracy in the important region of [1 -0.5] s. $LSTM-mv$ consistently performs worse than $FLDCRF-mv$ in early prediction regions and lacks stability in accuracy across time. LSTM systems without vehicle context exhibit significantly poorer performance compared to other systems.

On JAAD stopping scenarios, systems with vehicle context ($FLDCRF-mv$ and $LSTM-mv$) output more stable and early prediction of the stopping behaviour (see Fig. \ref{fig:JAAD-accuracy}b) than $mtn$ and $mtn+spa$ models. These two systems also produce comparable accuracy performances across time, $LSTM-mv$ being marginally better around the stopping instant (`0' mark). $LSTM-m$ and $LSTM-ms$ perform considerably better than $FLDCRF-m$ and $FLDCRF-ms$ respectively, but lacks stability in accuracy across time compared to the systems with vehicle context.

All considered systems produce similar prediction patterns on the JAAD starting scenarios (see Fig. \ref{fig:JAAD-accuracy}c), with accuracy performances improving over time, indicating successful predictions of the change of state in pedestrian motion. FLDCRF systems perform considerably better than their LSTM counterparts. $FLDCRF-mv$ performs best among all systems in the early prediction region [1 0] s. However, $FLDCRF-m$ and $FLDCRF-ms$ outperform $FLDCRF-mv$ after the `0' mark.

We compare different systems on 15 standing sequences in Fig. \ref{fig:JAAD-accuracy}d. $LSTM-m$ and $LSTM-mv$ perform comparably and better than other systems, reaching a stable 100\% prediction accuracy around/before the critical point. $FLDCRF-mv$ produces a stable 93\% prediction accuracy around and after the critical point, failing to predict the not-crossing behaviour in time in one of the sequences. We will analyze such failed cases shortly. 

The systems are compared on the task of predicting early transitions (standing-starting and crossing-stopping) in Tables \ref{table:stand-start} and \ref{table:cross-stop}. We consider accuracy of the systems over stipulated prediction windows. As expected, systems with vehicle context $FLDCRF-mv$ and $LSTM-mv$ perform significantly better than systems without vehicle context. $FLDCRF-mv$ outperforms $LSTM-mv$ in all early prediction windows. However, $LSTM-mv$ performs marginally better than $FLDCRF-mv$ after the `0' mark on crossing and stopping scenarios considered together (see Table \ref{table:cross-stop}).

\begin{table}
\vspace{2mm}
\caption{Classification accuracy of all considered JAAD standing \& starting sequences by different systems within different prediction windows.}
\centering
\renewcommand{\arraystretch}{1.2}
\setlength\tabcolsep{1.5pt}
\begin{tabular}{|p{3.35cm}|c|c|c|c|c|c|}
\hline
\multirow{2}{5cm}{\textbf{System}} & \multicolumn{6}{c|}{\textbf{Classification Accuracy(\%)}}\\
\cline{2-7}
& 2-0 s & 1.5-0 s & 1-0 s & 0.5-0 s & 0-(-0.5) s & 0-(-1) s\\
\hline \hline
FLDCRF- (mtn+veh) & \textbf{56.02} & \textbf{57.98} & \textbf{61.02} & \textbf{68.18} & \textbf{77.17} & \textbf{82.50} \\ \hline
FLDCRF- (mtn+spa) & 45.84 & 48.17 & 50.11 & 54.32 & 72.53 & 78.52 \\ \hline
FLDCRF- (mtn) & 46.15 & 47.84 & 50.45 & 55.23 & 70.91 & 78.18 \\ \hline
LSTM- (mtn+veh) & 51.82 & 53.15 & 56.48 & 60.91 & 71.92 & 77.95 \\ \hline
LSTM- (mtn+spa) & 35.24 & 36.84 & 38.07 & 40.91 & 56.16 & 66.82 \\ \hline
LSTM- (mtn) & 33.59 & 34.46 & 37.05 & 42.05 & 54.55 & 66.36 \\ \hline
\end{tabular}
\label{table:stand-start}
\vspace{-3mm}
\end{table}

\begin{table}
\vspace{3mm}
\caption{Classification accuracy of all considered JAAD continuous crossing \& stopping sequences by different systems within different prediction windows.}
\centering
\renewcommand{\arraystretch}{1.2}
\setlength\tabcolsep{1.5pt}
\begin{tabular}{|p{3.35cm}|c|c|c|c|c|c|}
\hline
\multirow{2}{5cm}{\textbf{System}} & \multicolumn{6}{c|}{\textbf{Classification Accuracy(\%)}}\\
\cline{2-7}
& 2-0 s & 1.5-0 s & 1-0 s & 0.5-0 s & 0-(-0.5) s & 0-(-1) s\\
\hline \hline
FLDCRF- (mtn+veh) & \textbf{90.47} & \textbf{91.39} & \textbf{91.83} & \textbf{93.08} & 93.68 & 95.29 \\ \hline
FLDCRF- (mtn+spa) & 74.85 & 75.57 & 77.21 & 78.85 & 78.63 & 79.62 \\ \hline
FLDCRF- (mtn) & 75.93 & 76.31 & 77.88 & 80.19 & 78.29 & 80.67 \\ \hline
LSTM- (mtn+veh) & 86.73 & 87.94 & 90.10 & \textbf{93.08} & \textbf{95.90} & \textbf{95.38} \\ \hline
LSTM- (mtn+spa) & 73.53 & 75.77 & 77.12 & 80.38 & 82.39 & 85.87 \\ \hline
LSTM- (mtn) & 75.87 & 76.79 & 77.4 & 78.46 & 83.93 & 87.5 \\ \hline
\end{tabular}
\label{table:cross-stop}
\vspace{-3mm}
\end{table}

\begin{figure}[h]
\includegraphics[scale=0.45, height=3cm]{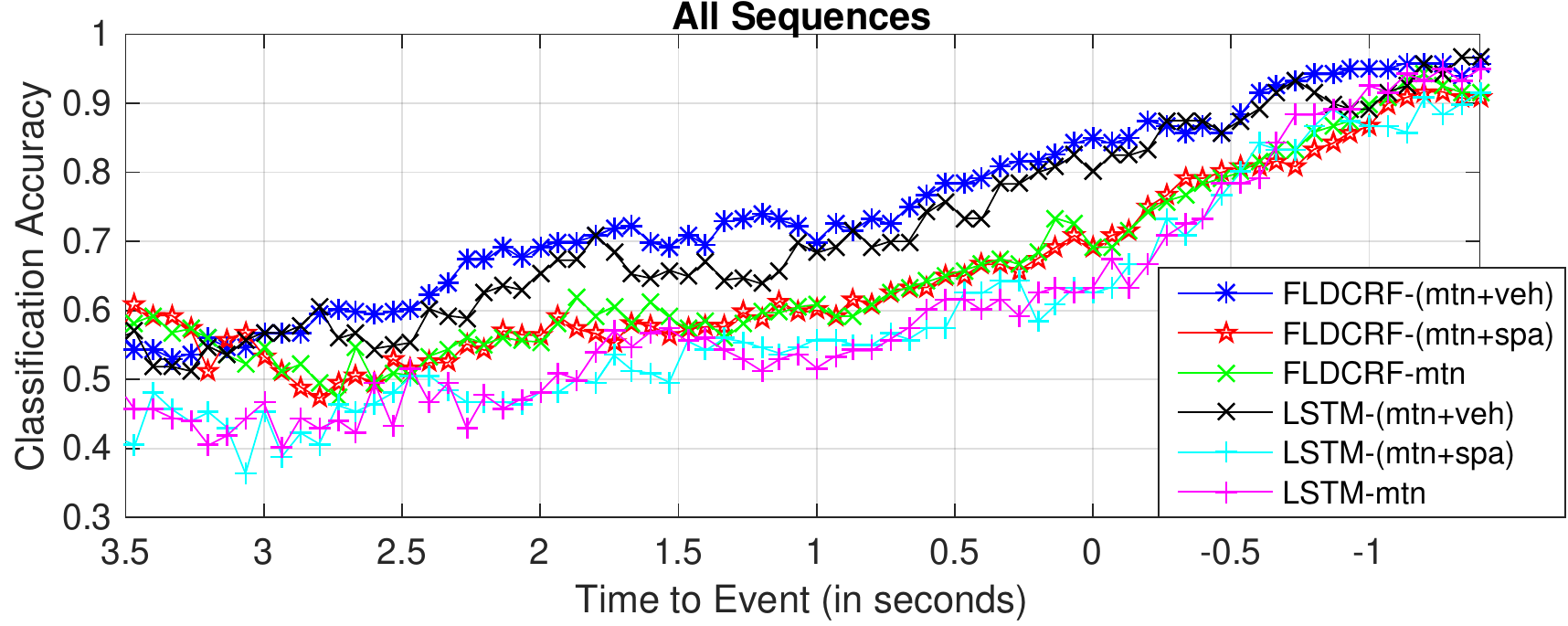}
\centering
\caption{Overall classification accuracy of different systems on JAAD sequences.}
\label{fig:all}
\end{figure}

$FLDCRF-mv$ and $LSTM-mv$ consistently performed well across all sequence types. Other systems have been better at times but lacked consistency. We can observe dominant performance of systems with vehicle context ($FLDCRF-mv$ and $LSTM-mv$) in Fig. \ref{fig:all}, producing superior overall accuracy compared to systems without vehicle context at all prediction instants. Moreover, $FLDCRF-mv$ performs similar/better than $LSTM-mv$ at all considered prediction instants, proving to be the best performing model on the JAAD dataset. $FLDCRF-mv$ produces an average accuracy of $\sim$78\% within the 1-0 s window and $\sim$82\% within the 0.5-0 s window, while $LSTM-mv$ outputs $\sim$75\% and $\sim$78\% in respective windows.

\begin{figure}%
\centering
\subfloat[\scriptsize{Average training time per outer fold.}]{{\includegraphics[width=4.1cm]{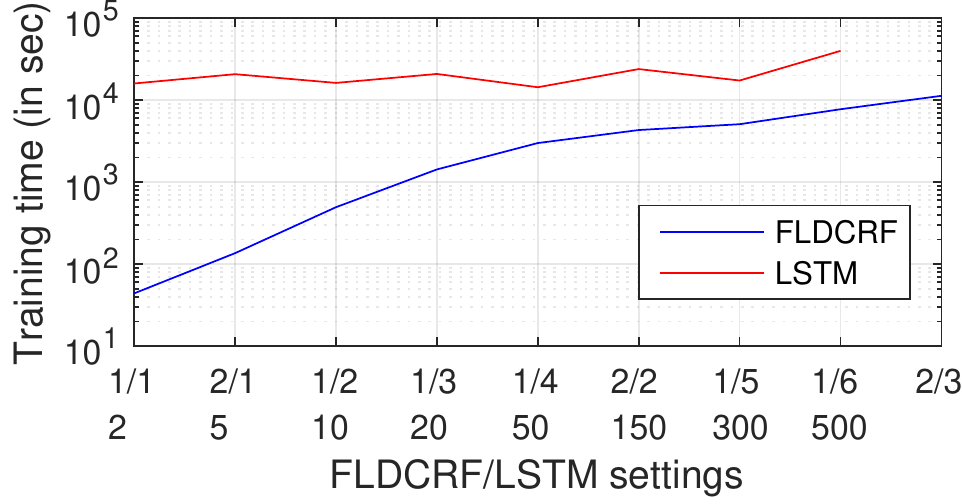} }}%
\quad
\subfloat[\scriptsize{Average inference time per frame.}]{{\includegraphics[width=4.1cm]{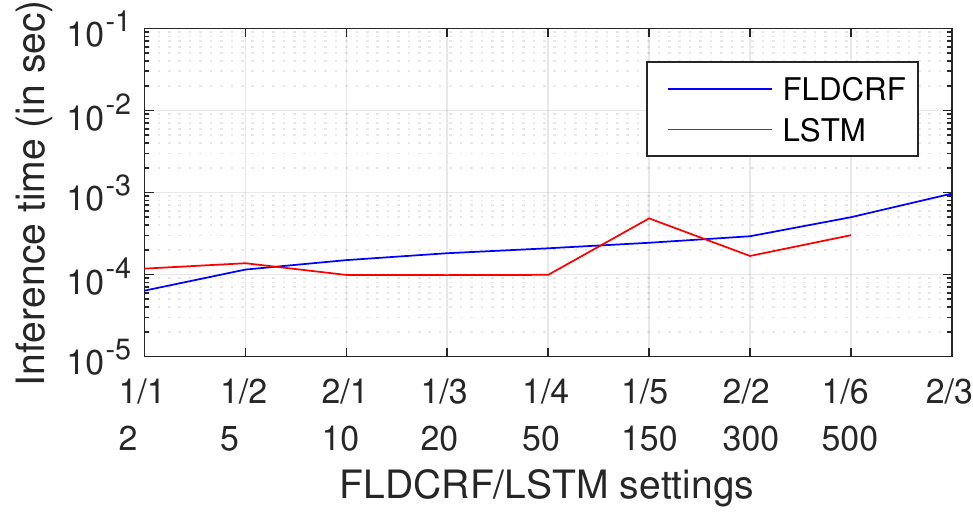} }}%
\caption{Training and inference times (excluding feature extraction) required by different FLDCRF and LSTM settings on $mtn+veh$ system.}%
\label{fig:JAAD-runtime}%
\end{figure}

Figure \ref{fig:JAAD-runtime} compares the required training and inference time by the considered FLDCRF and LSTM settings on $mtn+veh$ system. While both models needed similar time for inference, FLDCRF models required significantly lesser training time compared to the LSTM models. FLDCRF training and inference times can be further reduced by GPU implementation.

We analyze individual sequence outputs and failed cases by the $FLDCRF-mv$ system below.

\vspace{2mm}
\noindent \textbf{Failed case analysis:}

\begin{figure}%
\centering
\subfloat[\scriptsize{Continuous crossing sequences.}]{{\includegraphics[width=4.1cm]{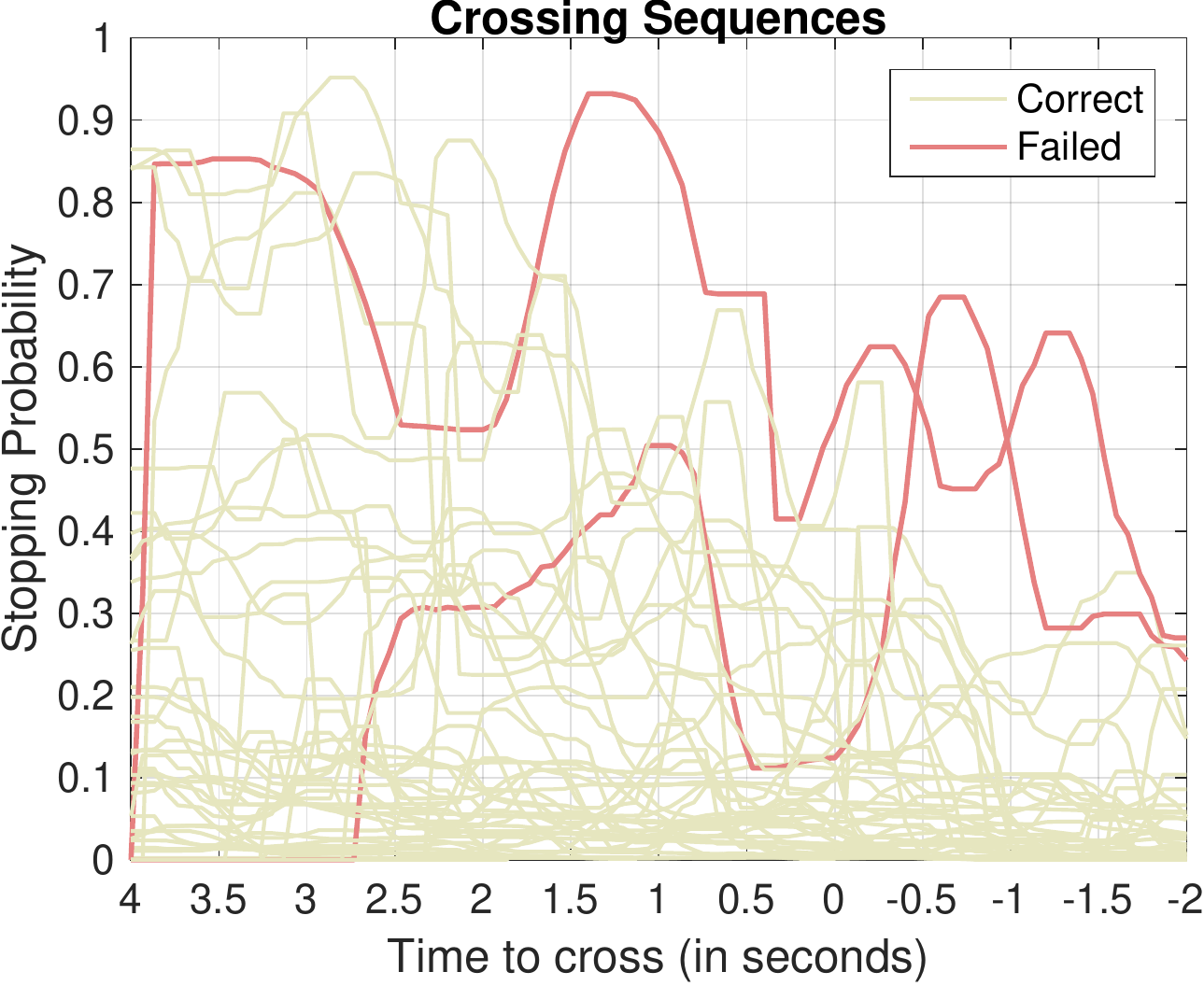} }}%
\quad
\subfloat[\scriptsize{Stopping sequences.}]{{\includegraphics[width=4.1cm]{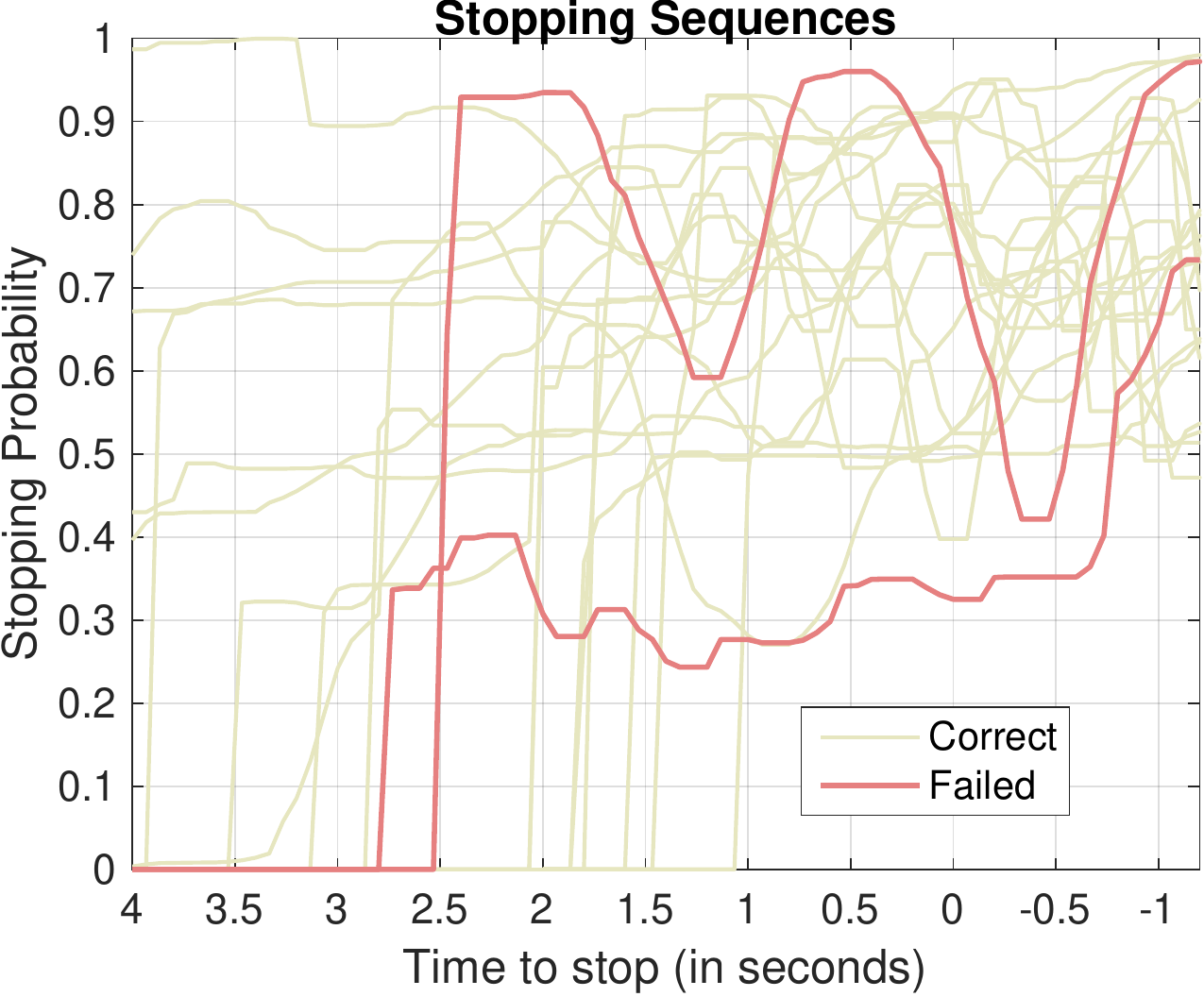} }}%
\quad
\subfloat[\scriptsize{Starting sequences.}]{{\includegraphics[width=4.1cm]{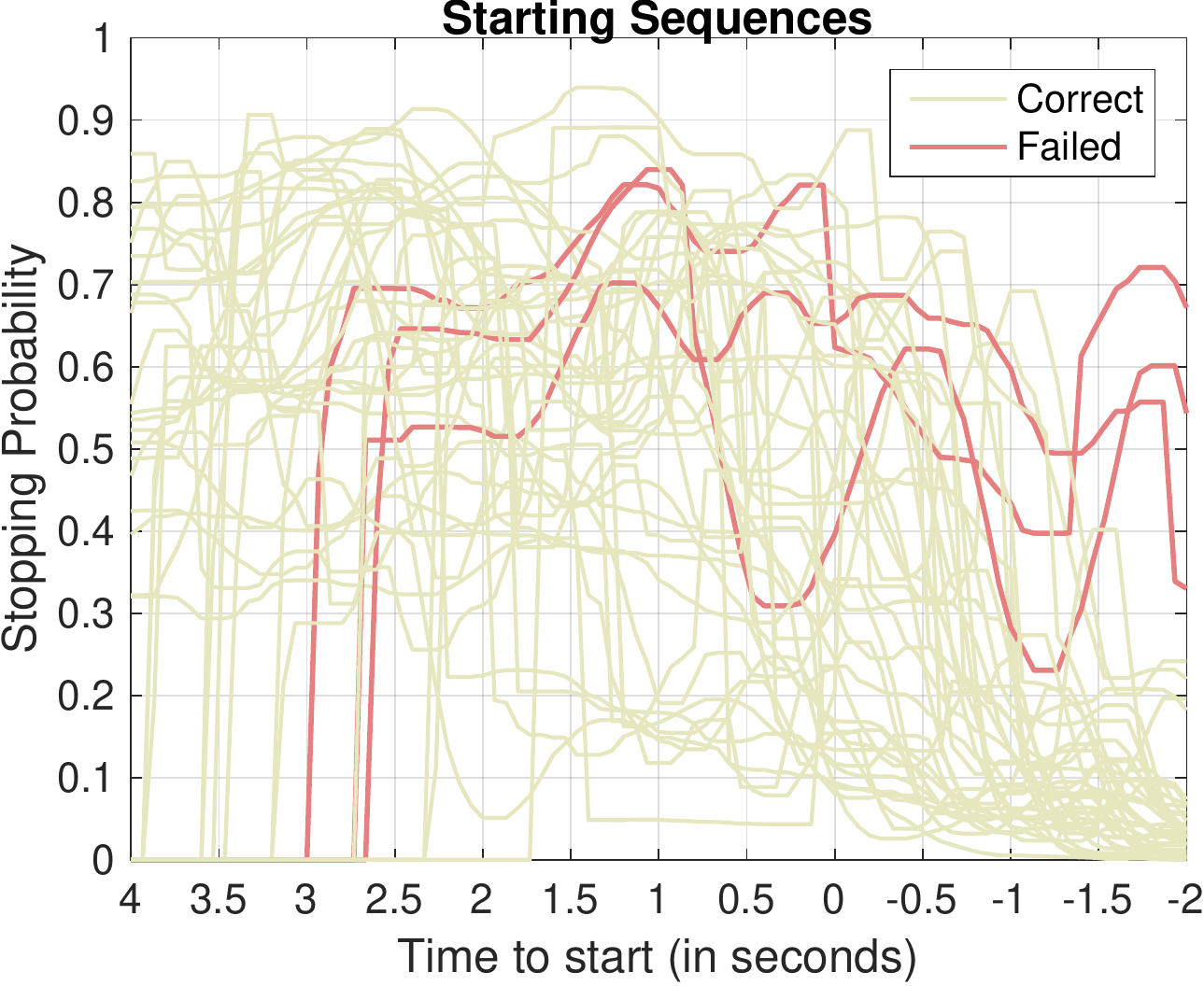} }}%
\quad
\subfloat[\scriptsize{Standing sequences.}]{{\includegraphics[width=4.1cm]{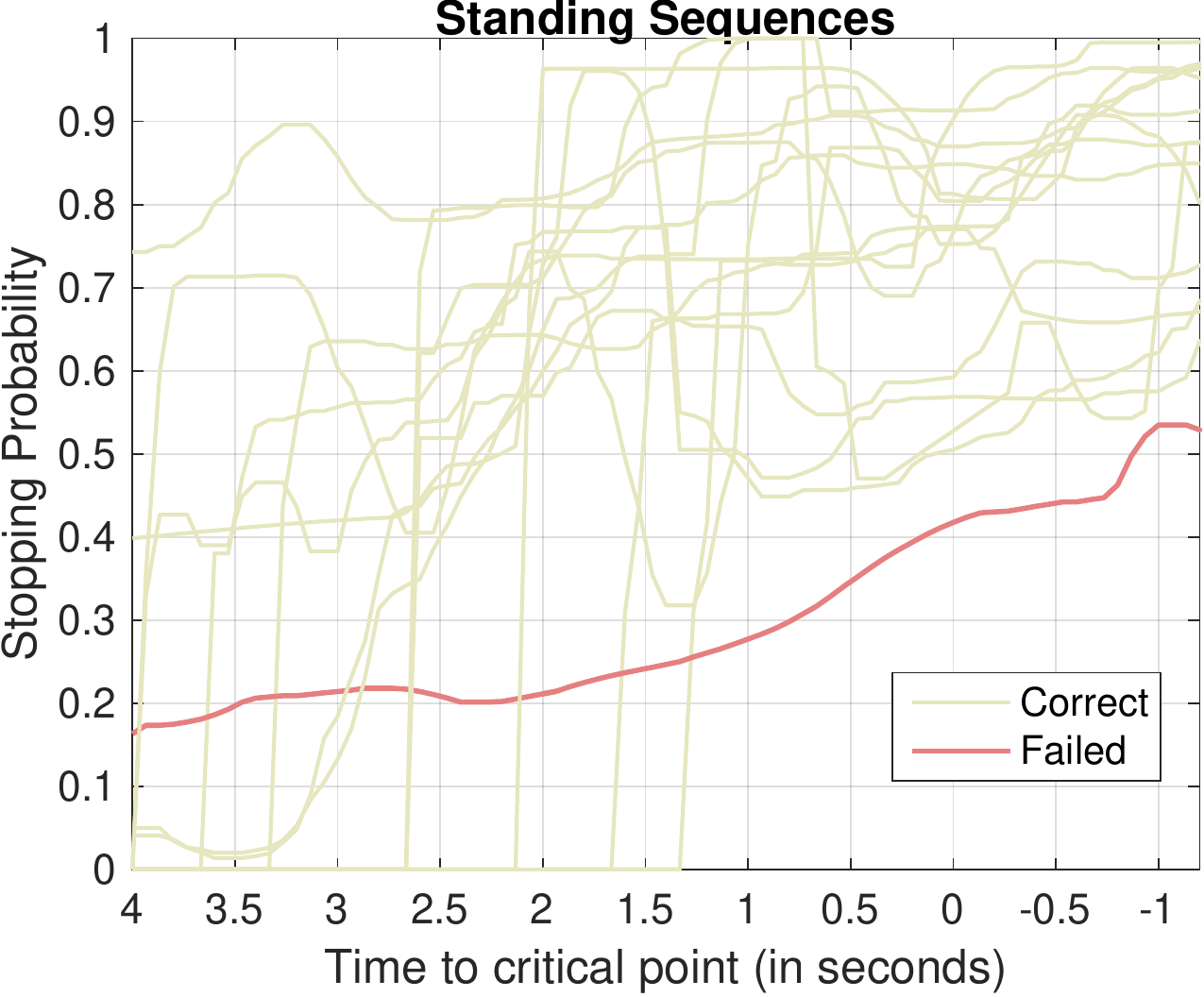} }}%
\caption{Individual sequence outputs by $FLDCRF-(mtn+veh)$ on the JAAD dataset. Failed cases are shown in bold red.}%
\label{fig:JAAD-failed}%
\end{figure}

Fig. \ref{fig:JAAD-failed} depicts failures in bold red from $FLDCRF-mv$ system. We denote a pedestrian sequence by its $video\_id$ and $pedestrian\_id$ in JAAD dataset, i.e. $<$$video\_id$$>$\_$<$$pedestrian\_id$$>$.

Crossing sequences $0161\_1$ and $0177\_2$, where the system fails to establish stable and accurate (stopping probability $<$ 0.5) outputs after -0.25 s, are highlighted in Fig. \ref{fig:JAAD-failed}a. In sequence  0161\_1, we find a momentary prediction glitch within [-0.5 -1] s window caused by a temporary hesitance from the pedestrian to continue crossing. The vehicle happened to be quite far from the pedestrian during the glitch avoiding a critical failure by the system. The failure in sequence 0177\_2 is caused by inappropriate vehicle context as the ego-vehicle is moving fast quite near ($<$15 m) the pedestrian when the crossing event commenced in a different lane. However, the vehicle decelerated within a short period and stable, accurate output was achieved shortly after the -1.5 s mark. Such errors can be corrected by adding the lane information as context.

All stopping scenarios were predicted correctly by the $FLDCRF-mv$ system, latest by 0.75 s after the stopping instant. The system fails to make early prediction (before `0') of the event on the highlighted sequences (0334\_2 and 0336\_1). A few sequences fall below probability 0.5 after the -1 s mark, but all such cases correspond to a starting event followed by the stopping event.

We have found early prediction of the `starting' event quite challenging. Fig. \ref{fig:JAAD-failed}c highlights three starting scenarios that fail to be stable and accurate by -1.5 s on the curves. However, all of them become stable and accurate shortly after the -2 s mark. 

\begin{figure}[h]
\includegraphics[scale=0.27]{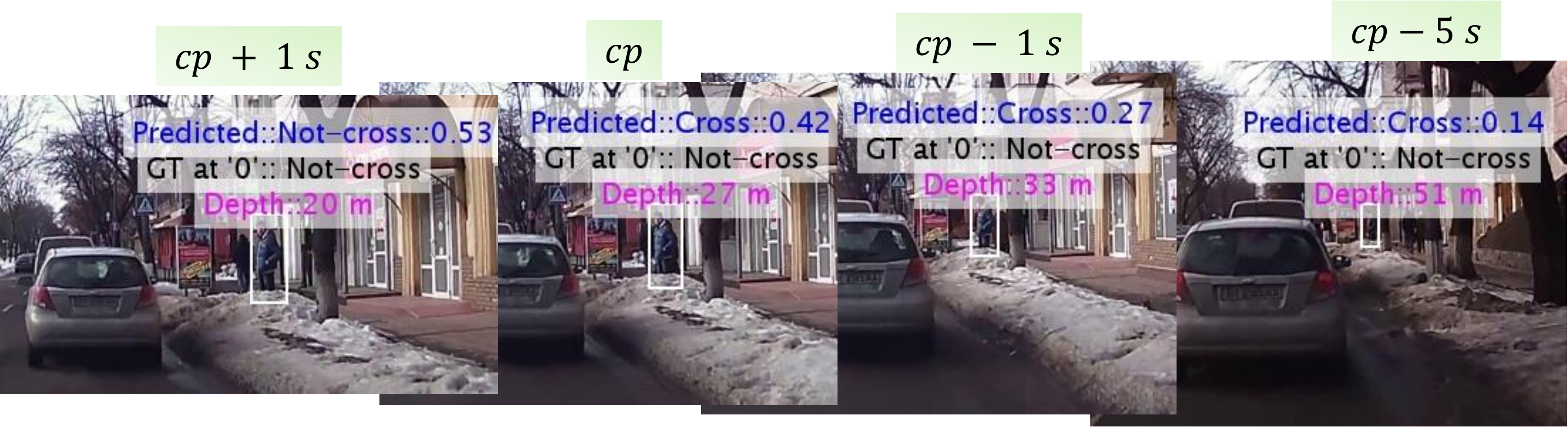}
\centering
\caption{Failed prediction by $FLDCRF-mv$ on standing sequence $0208\_02$. $cp$ deontes the critical point (`0' mark). GT is the ground-truth intention at `0'.}
\label{fig:stand_fail}
\end{figure}

Standing sequence $0208\_2$ is wrongly predicted as `crossing' by $FLDCRF-mv$ before and well after ($\approx$1 s) the defined critical point. A possible reason behind this is incomplete and inaccurate vehicle context. In this case, the other vehicles before the ego-vehicle are primarily responsible for the pedestrian to remain stationery (see Fig. \ref{fig:stand_fail}). The ego-vehicle is moving slowly during the event in a busy scene and is quite far from the pedestrian, causing the inaccurate context. We defined the critical point randomly due to limited number of images in the sequence. In such cases, we need to consider more complete pedestrian-vehicle interactions, which include other vehicles in the scene. 

JAAD dataset is not equipped with precise camera matrix information for the video sequences. Consequently, we utilized approximate values for these parameters, and observe our motion features to perform weaker than NTU dataset. However, combined with context features we obtain relatively early and accurate intention prediction for all four types of sequences from the $FLDCRF-mv$ and $LSTM-mv$ systems.

\section{Conclusion} \label{sec:conclude}

We presented a context model for pedestrian intention prediction for Autonomous Vehicles. We introduced vehicle interaction context in the problem for earlier and more accurate prediction. We also proposed Factored Latent-Dynamic Conditional Random Fields (FLDCRF) for single and multi-label sequence prediction/interaction tasks. FLDCRF led to more accurate and stable performance than LSTM over identical input features across our datasets. We plan to compare FLDCRF and LSTM on standard single and multi-label sequential machine learning datasets. We are also specifically interested in utilizing the interaction model variant of FLDCRF to capture complex pedestrian-vehicle interactions and apply the model on joint scene modeling tasks. In the system proposed in the paper, we will look to replace our pre-calibrated depth estimation with fully automated techniques, train with more data and build a real-time system on our approach. Moreover, in future work we will look to augment our intention prediction system with static scene context variables (e.g., zebra crossing, bus stop, lanes etc.) by attention models and propose more complete end-to-end models involving general static and dynamic interactions.

\section*{Acknowledgment}

We are very thankful to all the reviewers for their valuable feedback.

\ifCLASSOPTIONcaptionsoff
  \newpage
\fi




\begin{thebibliography}{1}

\bibitem{Daimler}
C. G. Keller, D.M. Gavrila, ``Will the Pedestrian Cross? A Study on Pedestrian Path Prediction", IEEE Transactions On Intelligent Transportation Systems, VOL. 15, NO. 2, APRIL 2014.

\bibitem{Ped-crossing-intent}

F. Schneemann, P. Heinemann, ``Context-based Detection of Pedestrian Crossing Intention for Autonomous Driving in Urban Environments", International Conference on Intelligent Robots and Systems (IROS), 2016.

\bibitem{LDCRF-intent}
A. Schulz and R. Stiefelhagen, ``Pedestrian Intention Recognition using Latent-dynamic Conditional Random Fields", 2015 IEEE Intelligent Vehicles Symposium (IV).

\bibitem{Intent-intersections}
S. Köhler, M. Goldhammer, S. Bauer, S. Zecha, K. Doll, U. Brunsmann, K. Dietmayer, ``Stationary Detection of the Pedestrian’s Intention at Intersections", IEEE Intelligent Transportation Systems Magazine (Volume: 5, Issue: 4, winter 2013).


\bibitem{Motion-classification}
J. Hariyono and K. H. Jo, ``Pedestrian Action Recognition using Motion Type Classification", 2015 IEEE 2nd International Conference on Cybernetics (CYBCONF).

\bibitem{Particle-intent}

Y. Hashimoto, Y.Gu, L. Hsu, M. Iryo-Asano, S. Kamijo, ``A Probabilistic Model of Pedestrian Crossing Behavior at Signalized Intersections for Connected Vehicles'', Transportation Research Part C: Emerging Technologies, Volume 71, October 2016, Pages 164-181.

\bibitem{HMM-intent}

R. Quintero, I. Parra, J. Lorenzo, D. Fernandez-Llorca and M. A. Sotelo, ``Pedestrian Intention Recognition by Means of a Hidden Markov Model and Body Language", 2017 IEEE 20th International Conference on Intelligent Transportation Systems (ITSC).

\bibitem{JAAD-application}

A. Rasouli, I. Kotseruba, J. K. Tsotsos. ``Are They Going to Cross? A Benchmark Dataset and Baseline for Pedestrian Crosswalk Behavior", ICCVW (2017). 

\bibitem{JAAD-application2}

P. Gujjar, R. Vaughn, ``Classifying Pedestrian Actions In Advance Using Predicted Video of Urban Driving Scenes", ICRA 2019.

\bibitem{Openpose}

Z. Fang, D. Vázquez, A. M. López, ``On-Board Detection of Pedestrian Intentions", Sensors, 2017, doi: 10.3390/s17102193.

\bibitem{Michael}

M. Hoy, Z. Tu, D. Kang, and J. Dauwels (2018), ``Learning to Predict Pedestrian Intention via Variational Tracking Networks", Proceedings of the {IEEE} 21st International Conference on Intelligent Transportation Systems, Maui, USA [accepted].

\bibitem{Context-based}
J. F. P. Kooij, N. Schneider, F. Flohr, D. M. Gavrila, ``Context-Based Pedestrian Path Prediction", ECCV 2014.

\bibitem{LDCRF-path}

A. T. Schulz, R. B. GmbH, ``A Controlled Interactive Multiple Model Filter for combined Pedestrian Intention Recognition and Path Prediction", 2015 IEEE 18th International Conference on Intelligent Transportation Systems.

\bibitem{Bayesian-filters}
N. Schneider, D. M. Gavrila, ``Pedestrian Path Prediction with Recursive Bayesian Filters”, GCPR 2013.

\bibitem{traj-est2}

V. Karasev, A. Ayvaci, B. Heisele, S. Soatto, ``Intent-Aware Long-Term Prediction of Pedestrian Motion", ICRA 2016.

\bibitem{LSTM}

A. Alahi, K. Goel, V. Ramanathan, A. Robicquet, L. Fei-Fei, S. Savarese, ``Social LSTM: Human Trajectory Prediction in Crowded Spaces", CVPR 2016.

\bibitem{traj-est1}

N. Lee, W. Choi, P. Vernaza, C. B. Choy, P. H. S. Torr, M. Chandraker, ``DESIRE: Distant Future Prediction in Dynamic Scenes with Interacting Agents", CVPR 2017.

\bibitem{mypaper}
S. Neogi, M. Hoy, W. Chaoqun, J. Dauwels, ``Context Based Pedestrian Intention Prediction Using Factored Latent Dynamic Conditional Random Fields", IEEE SSCI-2017.

\bibitem{JAAD}

I. Kotseruba, A. Rasouli, J. K. Tsotsos, ``Joint Attention in Autonomous Driving (JAAD)." arXiv preprint arXiv:1609.04741 (2016). 

\bibitem{CRF-first}

J. Lafferty, A. McCallum, F. C. N. Pereira, ``Conditional Random Fields: Probabilistic Models for Segmenting and Labeling Sequence Data", ACM, June 2001.

\bibitem{DCRF}
C. Sutton, K. Rohanimanesh, A. McCallum, ``Dynamic Conditional Random Fields: Factorized Probabilistic Models for Labeling and Segmenting Sequence Data", The Journal of Machine Learning Research, Volume 8, 5/1/2007, Pages 693-723.

\bibitem{semi-CRF}

S. Sarawagi, W. W. Cohen, . ``Semi-Markov Conditional Random Fields for Information Extraction". In Lawrence K. Saul, Yair Weiss, Léon Bottou (eds.). (2005). Advances in Neural Information Processing Systems 17. Cambridge, MA: MIT Press. pp. 1185–1192.

\bibitem{LDCRF}
L. P. Morency, A. Quattoni, T. Darrell, ``Latent-Dynamic Discriminative Models for Continuous Gesture Recognition", CVPR, IEEE Computer Society, (2007).

\bibitem{HMM}
L. R. Rabiner. ``A Tutorial on Hidden Markov Models and Selected Applications in Speech Recognition". Proceedings of the IEEE, 77(2):257–286, 1989.

\bibitem{Factorial-HMM}

Z. Ghahramani, M. I. Jordan, ``Factorial Hidden Markov Models", Machine Learning (1997) 29: 245. https://doi.org/10.1023/A:1007425814087.

\bibitem{traj-est3}

D. Varshneya, G. Srinivasaraghavan, ``Human Trajectory Prediction using Spatially aware Deep Attention Models", NIPS 2017.

\bibitem{traj-est5}

W. C. Ma, D. Huang, N. Lee, K. M. Kitani, ``Forecasting Interactive Dynamics of Pedestrians with Fictitious Play", CVPR 2017.

\bibitem{action-lstm}

A. Ullah, J. Ahmad, K. Muhammad, M. Sajjad, S. W. Baik, ``Action Recognition in Video Sequences using Deep Bi-directional LSTM with CNN features", IEEE Access, 2017.

\bibitem{Viterbi}
G. D. Forney Jr. ``The Viterbi algorithm". Proceedings of the IEEE. 61 (3): 268–278. doi:10.1109/PROC.1973.9030, (March 1973).

\bibitem{HCRF}

S. B. Wang, A. Quattoni, L. P. Morency, D. Demirdjian, T. Darrell, ``Hidden Conditional Random Fields for Gesture Recognition", IEEE Transactions on Pattern Analysis and Machine Intelligence (Volume: 29, Issue: 10, Oct. 2007).

\bibitem{Stan}
B. Carpenter, D.Lee, M. A. Brubaker, A. Riddell, A. Gelman, B. Goodrich, J. Guo, M. Hoffman, M. Betancourt, P. Li, ``Stan: A Probabilistic Programming Language", Journal of Statistical Software, Volume VV, Issue II, Nov. 2017.

\bibitem{Optical Flow}
C. Liu. ``Beyond Pixels: Exploring New Representations and Applications for Motion Analysis". Doctoral Thesis. Massachusetts Institute of Technology. May 2009.

\bibitem{obj-det}

S. Ren, K. He, R. Girshick, J. Sun, ``Faster R-CNN: Towards Real-Time Object Detection with Region Proposal Networks", NIPS 2015.

\bibitem{semantic-seg}

A. Kendall, V. Badrinarayanan, R. Cipolla, ``Bayesian SegNet: Model Uncertainty in Deep Convolutional Encoder-Decoder Architectures for Scene Understanding", BMVC 2017.

\bibitem{MATLAB}
MATLAB 8.0 and Statistics Toolbox 8.1, The MathWorks, Inc., Natick, Massachusetts, United States.

\bibitem{Daimler Dataset}
N. Schneider and D. M. Gavrila, ``Pedestrian Path Prediction with Recursive Bayesian Filters: A Comparative Study", In Lecture Notes in Computer Science: Proc. of the German Conference on Pattern Recognition (GCPR), vol. 8142, Springer, 2013.

\bibitem{CMU-data}

CMU, ``Cmu graphics lab motion capture database", http://mocap.cs.cmu.edu/subjects.php.

\bibitem{FLDCRF-arxiv} S. Neogi, J. Dauwels, ``Factored Latent-Dynamic Conditional Random Fields for Single and Multi-label Sequence Modeling", URL: https://arxiv.org/abs/1911.03667, 2019.


\end{thebibliography}
%

%

\begin{IEEEbiography}[{\includegraphics[width=1in,height=1.25in,clip,keepaspectratio]{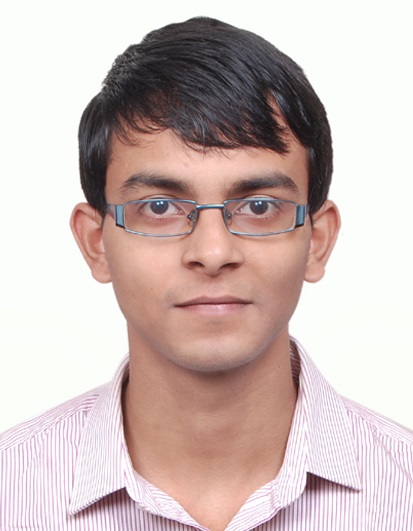}}]{Satyajit Neogi}
received his Bachelor of Engineering (B.E.) degree in Electronics and Telecommunications Engineering from Jadavpur University, Kolkata, India. He is currently pursuing his Ph.D. in Electrical Engineering at Nanyang Technological University, Singapore. His research interests include Probabilistic Graphical Models, Image and Video Processing and Scene Understanding.
\end{IEEEbiography}
\vskip 0pt plus -1fil
\begin{IEEEbiography}[{\includegraphics[width=1in,height=1.25in,clip,keepaspectratio]{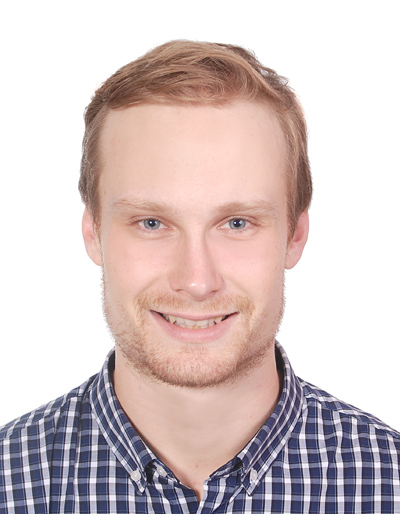}}]{Michael Hoy}
received a Ph.D. degree (Electrical Engineering) from the University of New South Wales, Australia in 2013. He is currently a poctdoctoral researcher at Nanyang Technological University, Singapore. His current research interests include Computer Vision, Statistical Signal Processing and perception for Autonomous systems (in particular, methods for applying deep learning to problems from these areas).
\end{IEEEbiography}
\vskip 0pt plus -1fil
\begin{IEEEbiography}[{\includegraphics[width=1in,height=1.25in,clip,keepaspectratio]{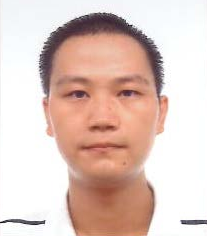}}]{Kang Dang}
received a Ph.D. degree from Nanyang Technological University in 2018. He is currently a research intern at Tencent Youtu Lab in Shanghai, China. His research interests include Computer Vision and Medical Image Analysis.
\end{IEEEbiography}
\vskip 0pt plus -1fil
\begin{IEEEbiography}[{\includegraphics[width=1in,height=1.25in,clip,keepaspectratio]{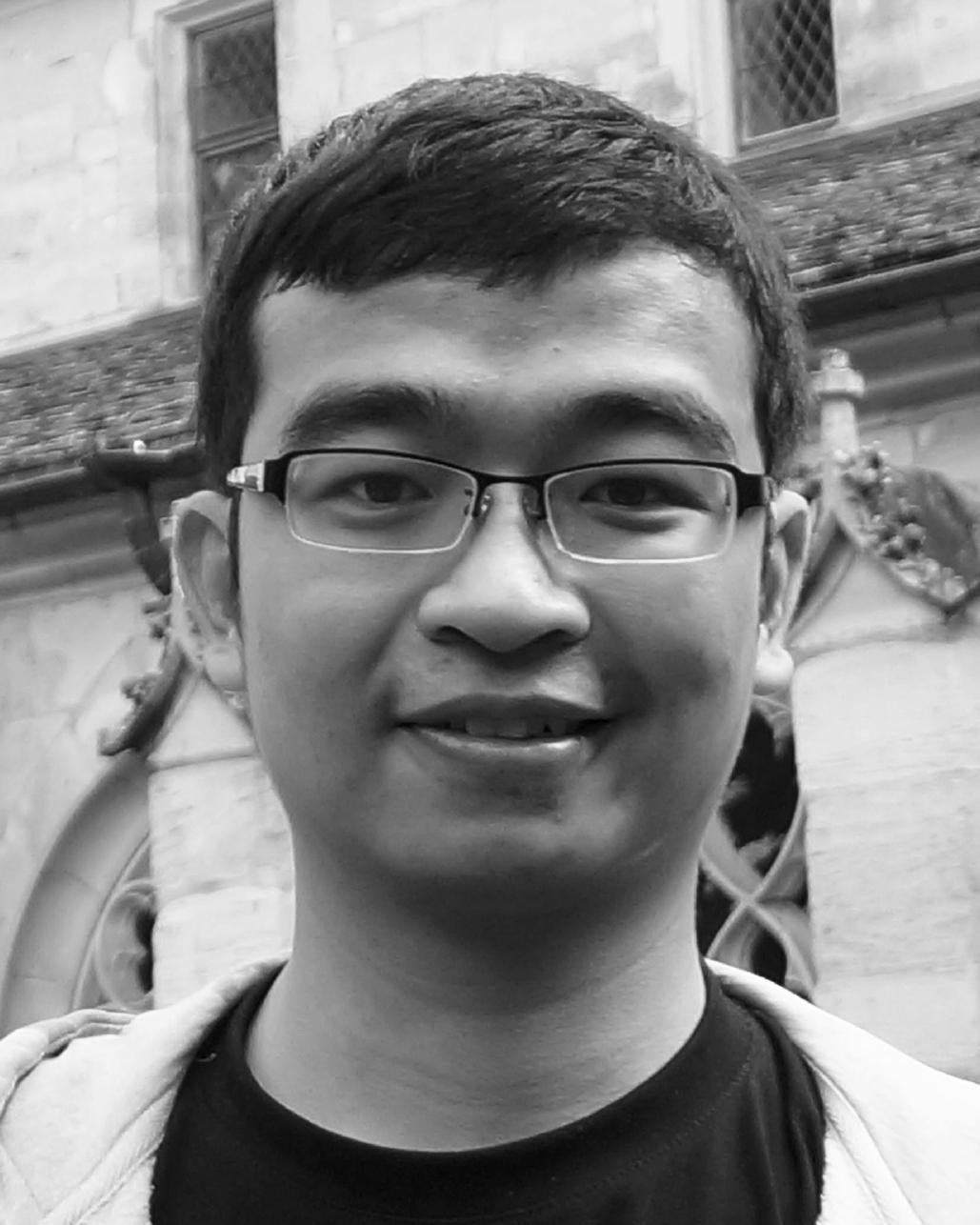}}]{Hang Yu}
Hang Yu (S'12-M'15) received the B.E. degree in Electronic and Information Engineering from University of Science and Technology Beijing (USTB), China in 2010, and the Ph.D. degree in Electrical and Electronics Engineering from Nanyang Technological University (NTU), Singapore in 2015. He is now a Postdoctoral Research Fellow in Centre for System Intelligence and Efficiency (EXQUISITUS), NTU, under the guidance of Prof. Justin Dauwels. His research interests include Statistical Signal Processing, Machine Learning, Graphical Models, Copulas, and Extreme-events modeling.
\end{IEEEbiography}
\vskip 0pt plus -1fil
\begin{IEEEbiography}[{\includegraphics[width=1in,height=1.25in,clip,keepaspectratio]{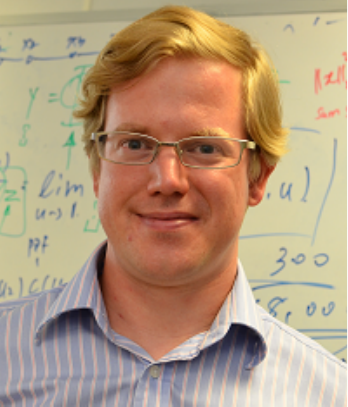}}]{Justin Dauwels}
is an Associate Professor with the School of Electrical and Electronic Engineering at Nanyang Technological University (NTU), Singapore. He is also the Deputy Director of the ST Engineering-NTU corporate lab. His research interests lie in Probabilistic Graphical Models, Intelligent Transportation and Human behaviour analysis. He received a PhD degree in Electrical Engineering from the Swiss Polytechnical Institute of Technology (ETH), Zurich in 2005. He was a postdoctoral fellow at the RIKEN Brain Science Institute (2006-2007) and a research scientist at the Massachusetts Institute of Technology (2008-2010). He has been a JSPS postdoctoral fellow (2007), a BAEF fellow (2008), a Henri-Benedictus fellow of the King Baudouin Foundation (2008), and a JSPS invited fellow (2010, 2011). His research on intelligent transportation systems has been featured by the BBC, Straits Times and Channel 5. His research on diagnosis of Alzheimers disease is featured at a 5-year exposition at the Science Center in Singapore. His research team has won several best paper awards at international conferences. He has filed 10+ US patents related to data analytics.
\end{IEEEbiography}




\newpage

\appendices

\section{Latent Dynamic Conditional Random Fields (LDCRF)} \label{Appen:LDCRF}

In this section, we will describe the LDCRF \cite{LDCRF} mathematical model.

\subsection{Model}

The task is to learn a probabilistic mapping between a time-series sequence of observed input features \textbf{x} = \{$x_1, x_2, ... , x_T$\} and a sequence of observed classifcation labels \textbf{y} = \{\(y_1, y_2, ... , y_T\)\}. \(y_t\) \(\in \) \(\Upsilon \), \(\forall j = 1,2, ... , T\), where \(\Upsilon \) is the set of classification labels. \textbf{h} = \{\(h_1, h_2, ... , h_T\)\} denotes the hidden layer for capturing intrinsic dynamics within each label. Each label $\ell$ \(\in \) \(\Upsilon \) is associated with a set of hidden states $\textit{$\mathcal{H}$}_{\ell}$. \textit{$\mathcal{H}$} is the set of all possible hidden states written as \textit{$\mathcal{H}$} = \(\bigcup_{\ell} \textit{$\mathcal{H}$}_{\ell} \). \(\textit{$\mathcal{H}$}_{\ell} \) are disjoint $\forall$$\ell$ $\in$ \(\Upsilon \). Each $h_t$ is restricted to belong to the set ${\mathcal{H}}_{y_t}$, i.e., \(h_t \) \(\in \) \(\textit{$\mathcal{H}$}_{y_t}, \forall t = 1,2, ..., T\) . The conditional model is defined as:

\begin{equation}
\begin{split}
\label{eqn1}
P\left(\textbf{y} \mid \textbf{x}, \theta\right) = \sum_\textbf{h} P\left(\textbf{y} \mid \textbf{x}, \textbf{h}, \theta\right) P\left(\textbf{h} \mid \textbf{x}, \theta\right).
\end{split}
\end{equation}

Equation \eqref{eqn1} can be re-written using the graph structure in Fig. \ref{fig:CDCRF}c as:

\begin{equation}
\begin{split}
\label{eqn1-new}
P\left(\textbf{y} \mid \textbf{x}, \theta\right) = \sum_{\textbf{h}:\forall h_t \in \textit{$\mathcal{H}$}_{y_t} } P\left(\textbf{y} \mid \textbf{h}, \theta\right) P\left(\textbf{h} \mid \textbf{x}, \theta\right) \\ +
\sum_{\textbf{h}:\exists h_t \not\in \textit{$\mathcal{H}$}_{y_t} } P\left(\textbf{y} \mid \textbf{h}, \theta\right) P\left(\textbf{h} \mid \textbf{x}, \theta\right).
\end{split}
\end{equation}

Applying model constraints, we can write the following:
\vspace{2mm}
\begin{equation}
P({y_{t}}=\ell \mid {h_t}) = 
\begin{cases}
1, & h_t \in \mathcal{H}_{y_{t}=\ell} \\
0, & h_t \not\in \mathcal{H}_{y_{t}=\ell}.
\end{cases}
\label{eqn1-constraint}
\end{equation}

The model in equation \eqref{eqn1-new} can be simplified using equation \eqref{eqn1-constraint} as:

\begin{equation}
\label{eqn3}
P\left(\textbf{y} \mid \textbf{x}, \theta\right) = \sum_{\textbf{h}:\forall h_t \in \textit{$\mathcal{H}$}_{y_t} } P\left(\textbf{h} \mid \textbf{x}, \theta\right).
\end{equation}

$P\left(\textbf{h}\mid\textbf{x}, \theta\right)$ is described using Conditional Random Field formulation given by:

\begin{equation}
\label{eqn4}
P\left(\textbf{h} \mid \textbf{x}, \theta\right) = \frac{1}{\textbf{\textit{Z}}\left(\textbf{x},\theta\right)} \exp \left(\sum_k \theta_k . F_k\left(\textbf{h},\textbf{x}\right)\right),
\end{equation}

\noindent where index $k$ ranges over all parameters $\theta = \{\theta_k\}$ and \( \textbf{\textit{Z}}(\textbf{x},\theta) \) is the partition function defined as:

\begin{equation}
\label{eqn5}
\textbf{\textit{Z}}(\textbf{x},\theta) = \sum_\textbf{h} {\exp\left(\sum_k \theta_k . F_k(\textbf{h},\textbf{x})\right)}.
\end{equation}

\noindent The feature functions \(F_k \)'s are defined as: \[ F_k(\textbf{h},\textbf{x}) = \sum_{t=1}^T f_k(h_{t-1}, h_t, \textbf{x}, \textit{t}),\] 

Feature functions \(f_k(h_{t-1}, h_t, \textbf{x}, \textit{t}) \) can be either an \textit{observation} (also called \textit{state}) function \(s_k(h_t, \textbf{x}, \textit{t}) \) or a \textit{transition} function \(t_k(h_{t-1}, h_t, \textbf{x}, \textit{t}) \).

\subsection{Training Model Parameters} \label{Training}

Parameters of the model can be estimated by maximizing the conditional log-likelihood of the training data given by equation \eqref{eqn6}:

\begin{equation}
\label{eqn6}
\textit{\textbf{L}}(\theta) = \sum_{n=1}^N {\log P(\textbf{y}^{(n)} \mid \textbf{x}^{(n)}, \theta)} - \frac{{\parallel\theta\parallel}^2}{2\sigma^2},
\end{equation}

\noindent where \textit{N} is the total number of available labeled sequences. The second term in equation \eqref{eqn6} is the log of a Gaussian prior with variance \(\sigma^2\).

\subsection{Inference}

Given a new test sequence \textbf{x}, the inference task is given by:

\begin{equation}
\label{eqn7}
\hat{\textbf{y}} = \text{argmax}_{\textbf{y}} \quad P(\textbf{y} \mid \textbf{x}, \hat{\theta}),
\end{equation}

Using model constraints, equation \eqref{eqn7} can be re-written as:

\begin{equation}
\label{eqn8}
\hat{\textbf{y}} = \text{argmax}_{\textbf{y}} \quad \sum_{\textbf{h}:\forall h_t \in \textit{$\mathcal{H}$}_{y_t} } P(\textbf{h} \mid \textbf{x}, \hat{\theta}),
\end{equation}

We apply forward recursions of belief propagation for our inference as the problem of intention prediction must be solved online. In other words, at each time instant $t$, we compute the marginals $P(h_t \mid {x_{1:t}}, \theta)$ and sum them according to the disjoint sets of hidden states to obtain $P(y_t \mid {x_{1:t}}, \theta)$ = $\sum_{h_t \in \textit{$\mathcal{H}$}_{y_t} } P(h_t \mid {x_{1:t}}, \theta), t = 1,2, ...$. Then, we infer the label $y_t$ corresponding to the maximum probability. Forward-backward algorithm \cite{HMM} and Viterbi algorithm \cite{Viterbi} can also be applied for problems where online inference is not required. \\

\section{Real world lateral motion estimation from monocular image sequence} \label{Appen:long-motion}

As we find a pedestrian's intent of crossing/not-crossing is characterized by the real world lateral component of his/her motion (w.r.t.~the ego-vehicle moving direction), we attempt to extract an approximate equivalent of such motion component in the image plane of a monocular image sequence using the camera matrix information. 

\begin{figure}[h]
\includegraphics[scale=0.6]{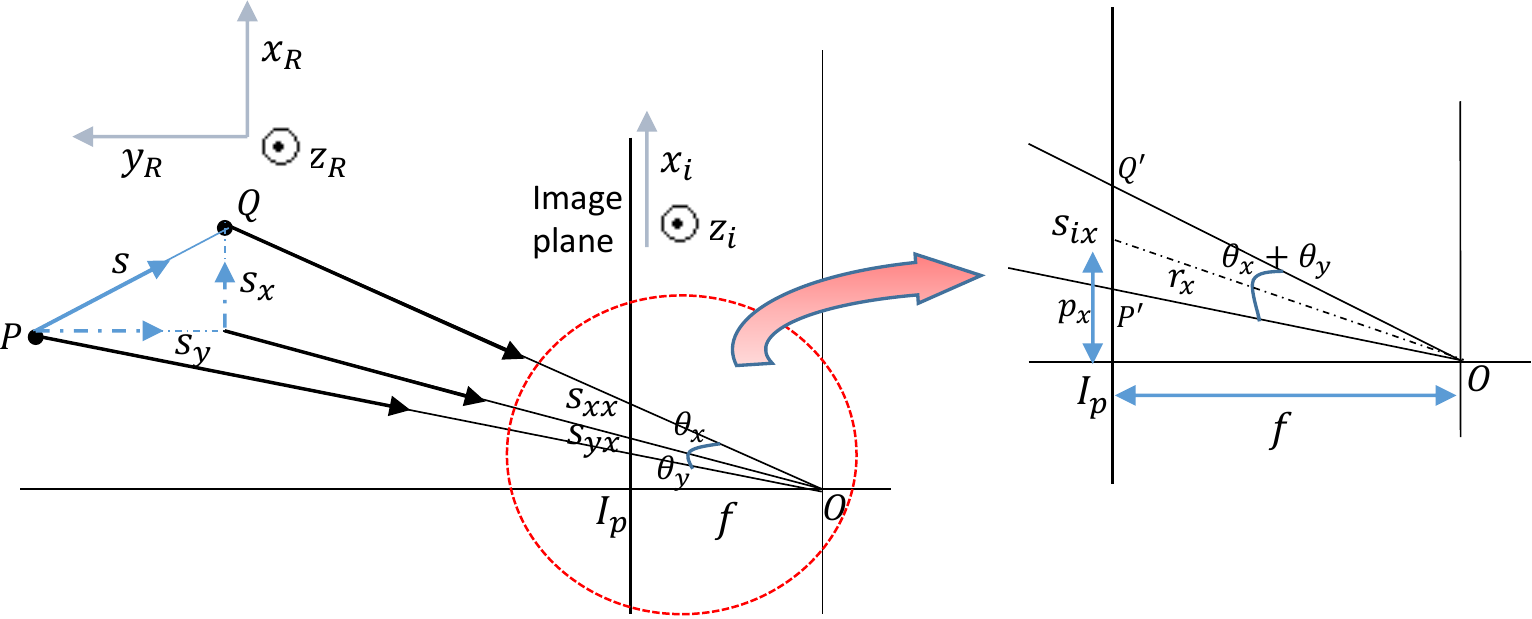}
\centering
\caption{Displacement in real world and its image equivalent.}
\label{fig:real-world}
\end{figure}

Fig. \ref{fig:real-world} depicts the scenario. $O$ is the camera center and $I_p$ is the principle point on image plane. $x_R$ and $y_R$ are the lateral and longitudinal (movement direction of the ego-vehicle) directions in real world respectively. $\vec{PQ} = \vec{S}$ represents relative displacement of a pedestrian point w.r.t. the ego-vehicle in real world between two consecutive time instants. We assume negligible pedestrian displacement along the vertical real world direction $z_R$. As we can see, corresponding displacement along $x_i$ image axis is $S_{ix}$, which contains components from both real world lateral ($S_x$) and longitudinal displacements ($S_y$), projected as $S_{xx}$ and $S_{yx}$ respectively onto the image plane, i.e., $S_{ix} = S_{xx} + S_{yx}$. Similarly, considering displacement along $z_i$ image axis to be $S_{iz}$, we can write the following two equations:

\begin{align*}
\label{eqn:real-world}
S_{ix} &= \pm r_x \cdot \theta_x \pm r_x \cdot \theta_y, \\
S_{iz} &= \pm r_z \cdot \theta_y
\end{align*}

\begin{figure}[h]
\includegraphics[scale=0.6]{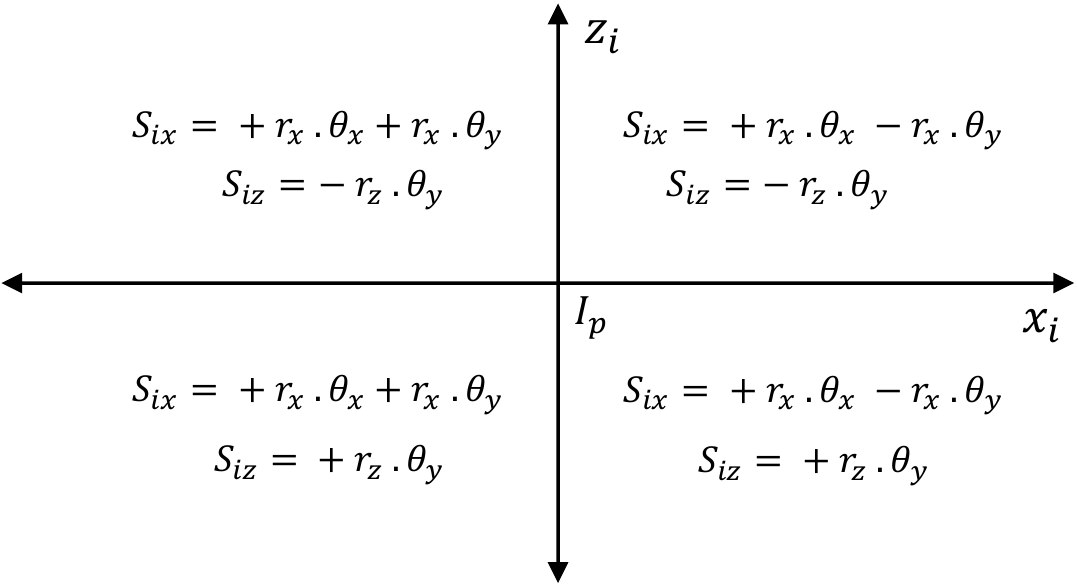}
\centering
\caption{Sign conventions on the image plane. $\theta_y$ is considered to be along +ve $y$ axis is real world.}
\label{fig:sign-convention}
\end{figure}

Since, we assume negligible pedestrian motion along vertical real world direction, $S_{iz}$ only carries traces of pedestrian's real world longitudinal motion. $\pm$ indicates the components can be additive or subtractive depending on the image quadrant of movement. Considering $+ \theta_x$ and $+ \theta_y$ to reflect real world +ve $x$ and +ve $y$ direction respectively, different possible scenarios are illustrated on the image plane $x_iz_i$ in Fig. \ref{fig:sign-convention}. $r_x$ and $r_z$ are the known respective distances from camera center O to the center of the displacements (situated at ($p_x$, $p_z$) on image plane) along axes $x_i$ and $z_i$, given by,

\begin{equation}
\label{eqn:r-x}
r_x = \sqrt{f^2 + p_x^2}, \quad
r_z = \sqrt{f^2 + p_z^2}
\end{equation}

The center of displacement is considered to minimize error in approximating $s = r \cdot \theta$ (instead of $\vec{s} = r \cdot \theta \cdot \hat{\theta}+ dr \cdot \hat{r}$), to reduce number of unknown variables. The goal is to solve for $\theta_x$ and $\theta_y$ using the equations illustrated in Fig. \ref{fig:sign-convention}.

$S_{xx}$ gives approximate pedestrian's lateral motion equivalent on image plane and is obtained using $S_{xx} = \pm r_x \cdot \theta_x$.

\section{Daimler Dataset} \label{Appen:Daim}

In this section, we present results of our motion-only approach on the Daimler dataset \cite{Daimler Dataset}.

Daimler Dataset is a public pedestrian benchmark dataset introduced in \cite{Daimler Dataset}. It contains several single pedestrian movement sequences captured at 16 fps using a stereo camera setup mounted on a moving vehicle. The movements primarily include three distinct scenarios - a) A laterally moving pedestrian `crossing' or `stopping' before a vehicle and b) A pedestrian moving along the sidewalk `bending in' to the road and c) A pedestrian `starting' to cross the road in front of a vehicle from a static position. Ground truth labels are provided by means of bounding boxes. For all scenarios, important events like `stopping' , `crossing', `bending' and `starting' instants are marked as `0'-th instant. The whole dataset contains 36 training and 32 testing scenarios. Each video sequence has a resolution $1176 \times 640$.

We evaluate on the same train and test splits as in \cite{Daimler}. We also utilize same training labels as in \cite{Daimler}. For training crossing scenarios, intent label `crossing' has been marked throughout. For training stopping sequences, data from 20 frames before the \textit{stopping instant} (specified in the dataset for each stopping sequence) has been labeled as `stopping' intent (or `not-crossing', as we call it). For Daimler dataset, our feature vector $x_{t,Daim}$ only contains pedestrian motion features, i.e. $x_{t,Daim} = \{x_{m,t}\}$.

\subsection{Results} \label{sec:daimler-results}

\begin{figure}[h]
\includegraphics[scale=0.4]{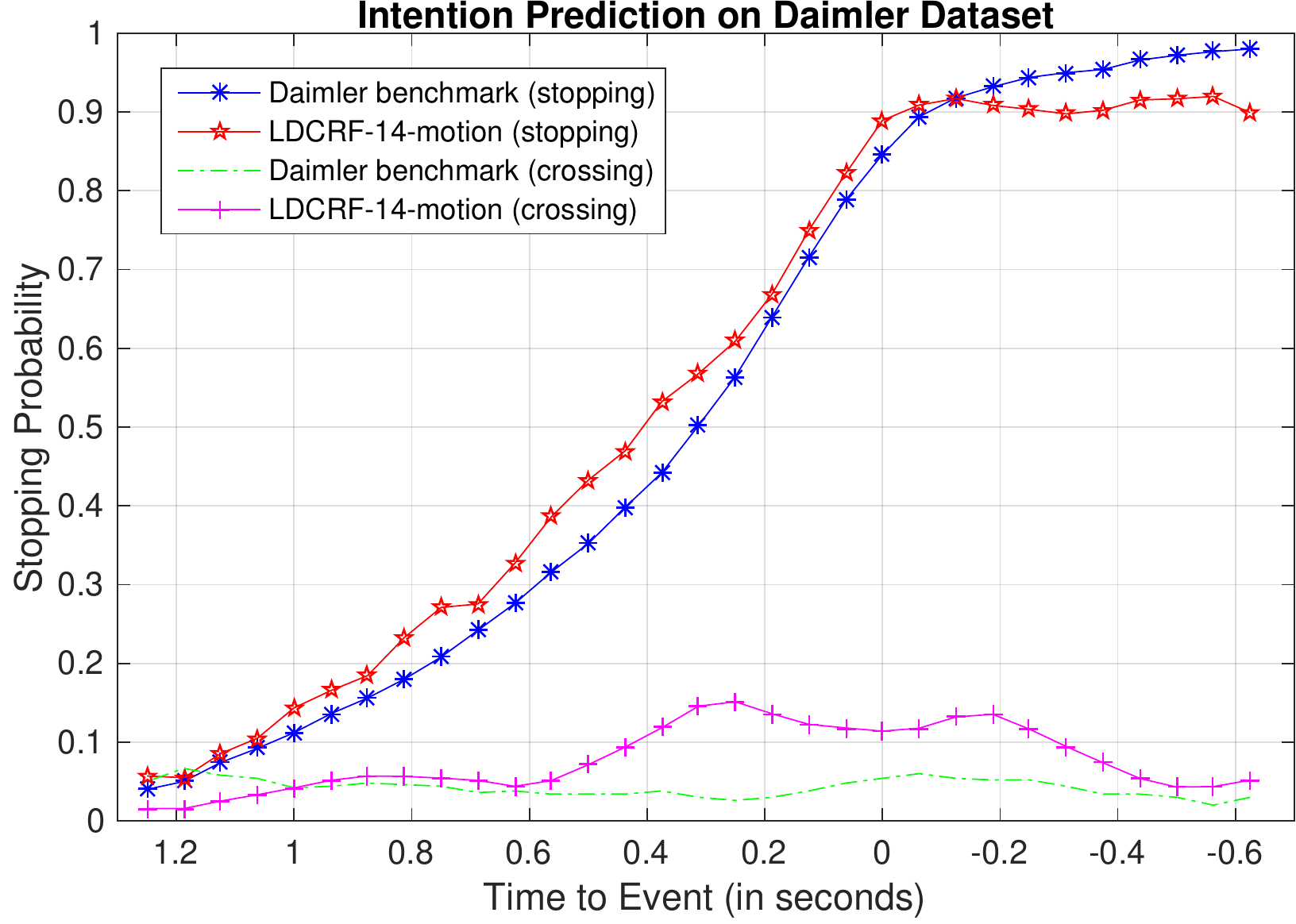}
\centering
\caption{Performance comparison between a Daimler benchmark \cite{Daimler} and our motion based approach.}
\label{fig:Daimler-results}
\end{figure}

Fig. \ref{fig:Daimler-results} presents a comparison of our approach with the best results presented in \cite{Daimler}. `0' along x-axis represent the event (`crossing' or `stopping') instants and are labeled for all test sequences in the dataset. As stated earlier, we find vehicle context irrelevant within the Daimler dataset and hence only test our motion based approach for intention prediction, i.e. here $x_{t,Daim} = \{x_{m,t}\}$. We apply a FLDCRF with 1 layer and 4 states per label over our motion features. 

The best results in \cite{Daimler} are obtained from a Gaussian Process Dynamic Model (GPDM) applied on dense optical flow derived features from stereo images. As we can see from Fig. \ref{fig:Daimler-results}, our motion features with FLDCRF perform comparably to the results in \cite{Daimler}. While our approach performs marginally better for the stopping scenarios, \cite{Daimler} is more consistent through time over the crossing sequences. Our methods although enjoys the liberty of using only single camera images.

\section{JAAD additional material} \label{Appen:JAAD}

\subsection{Sequence details} \label{sec:JAAD-sequences}

\begin{figure}[h]
\includegraphics[scale=0.65]{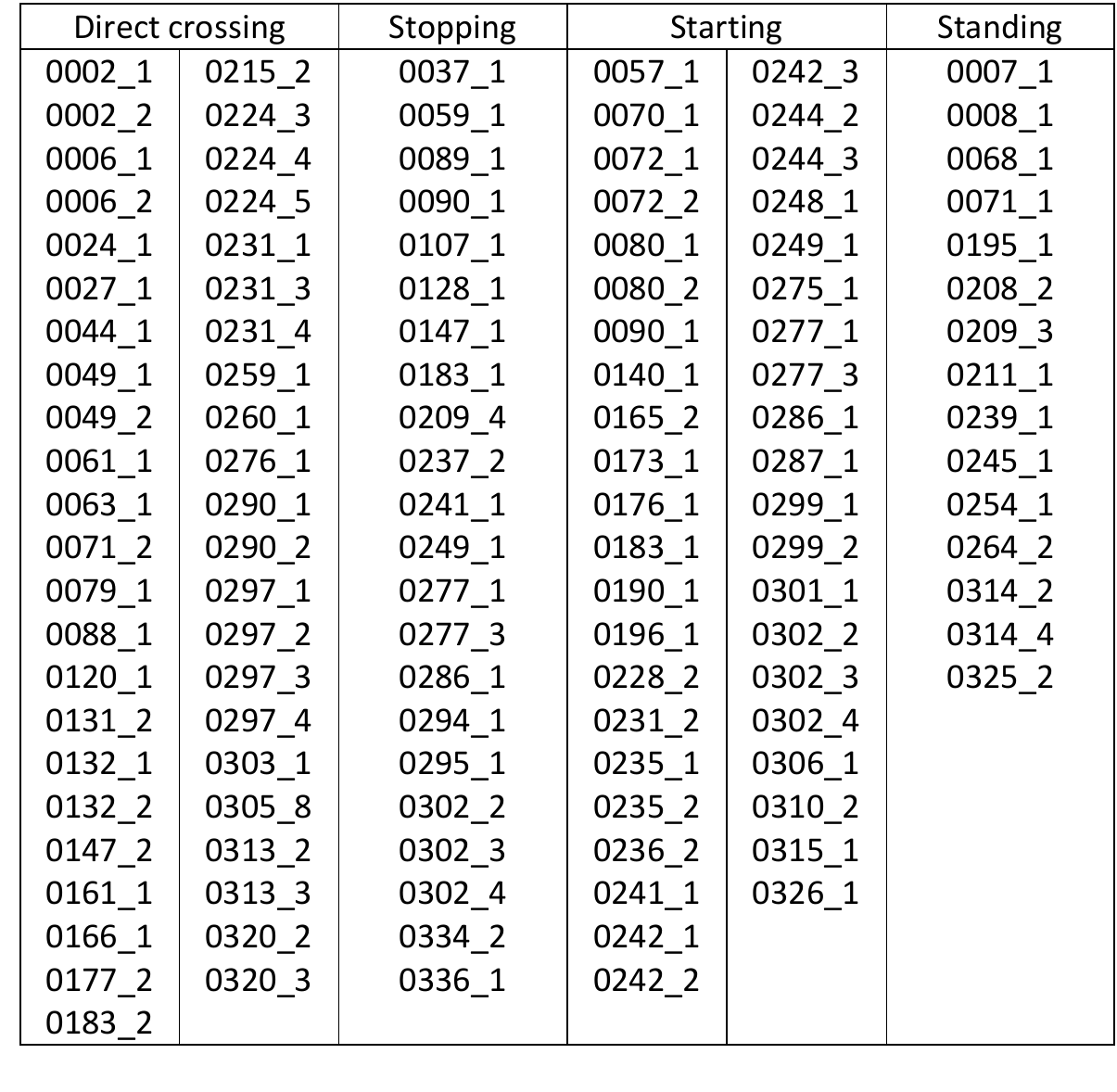}
\centering
\caption{Considered JAAD pedestrian sequences in our evaluation.}
\label{fig:JAAD-seq-details}
\end{figure}

Considered JAAD sequences are presented in Fig. \ref{fig:JAAD-seq-details}. A pedestrian sequence is denoted by its $video\_id$ and $pedestrian\_id$ in JAAD dataset, in the following format: $<$$video\_id$$>$\_$<$$pedestrian\_id$$>$.

\subsection{Stopping Probability vs Time curves} \label{sec:JAAD-results-excess}

\begin{figure}%
\centering
\subfloat[\scriptsize{Continuous crossing Sequences}]{{\includegraphics[width=4.1cm]{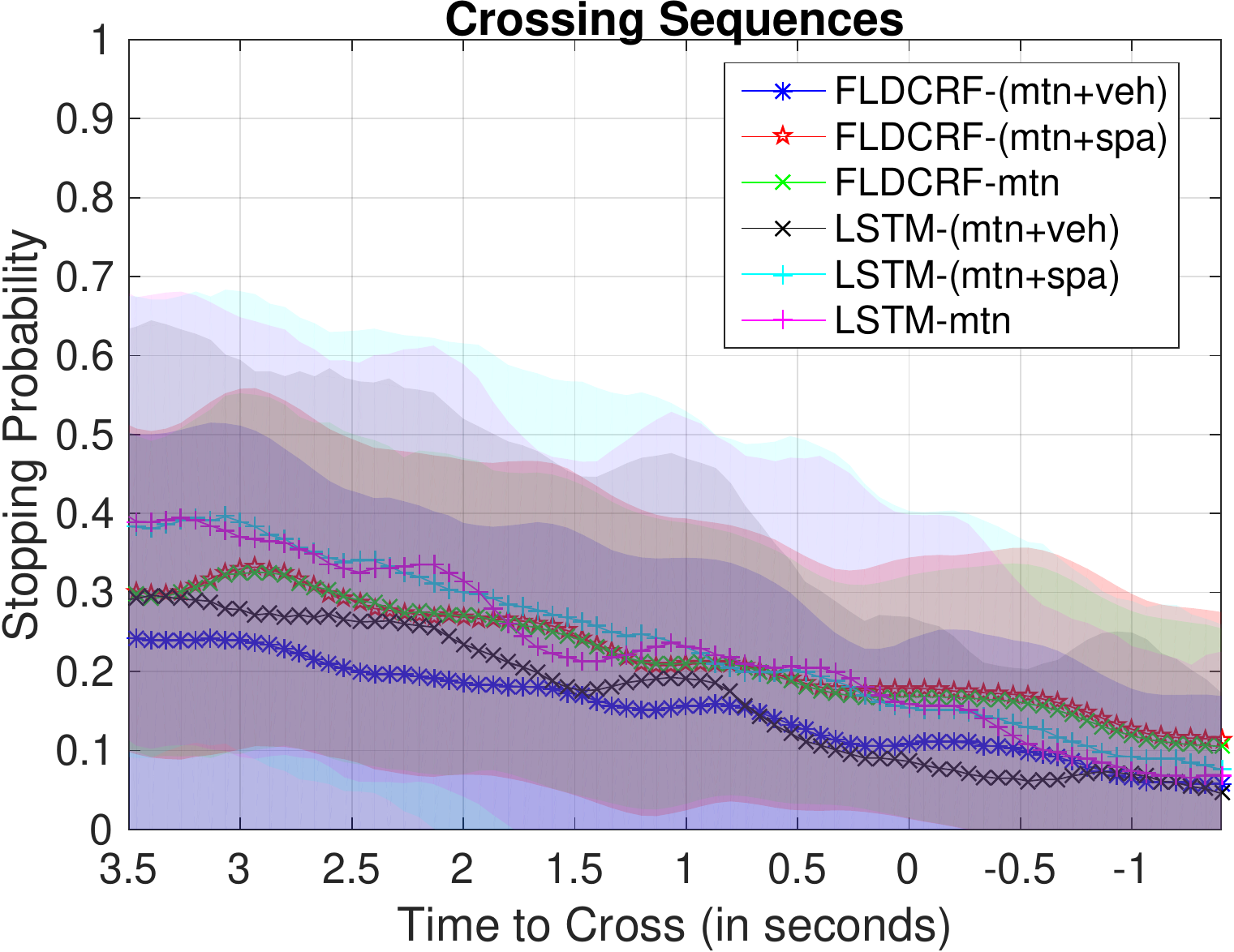} }}%
\quad
\subfloat[\scriptsize{Stopping Sequences}]{{\includegraphics[width=4.1cm]{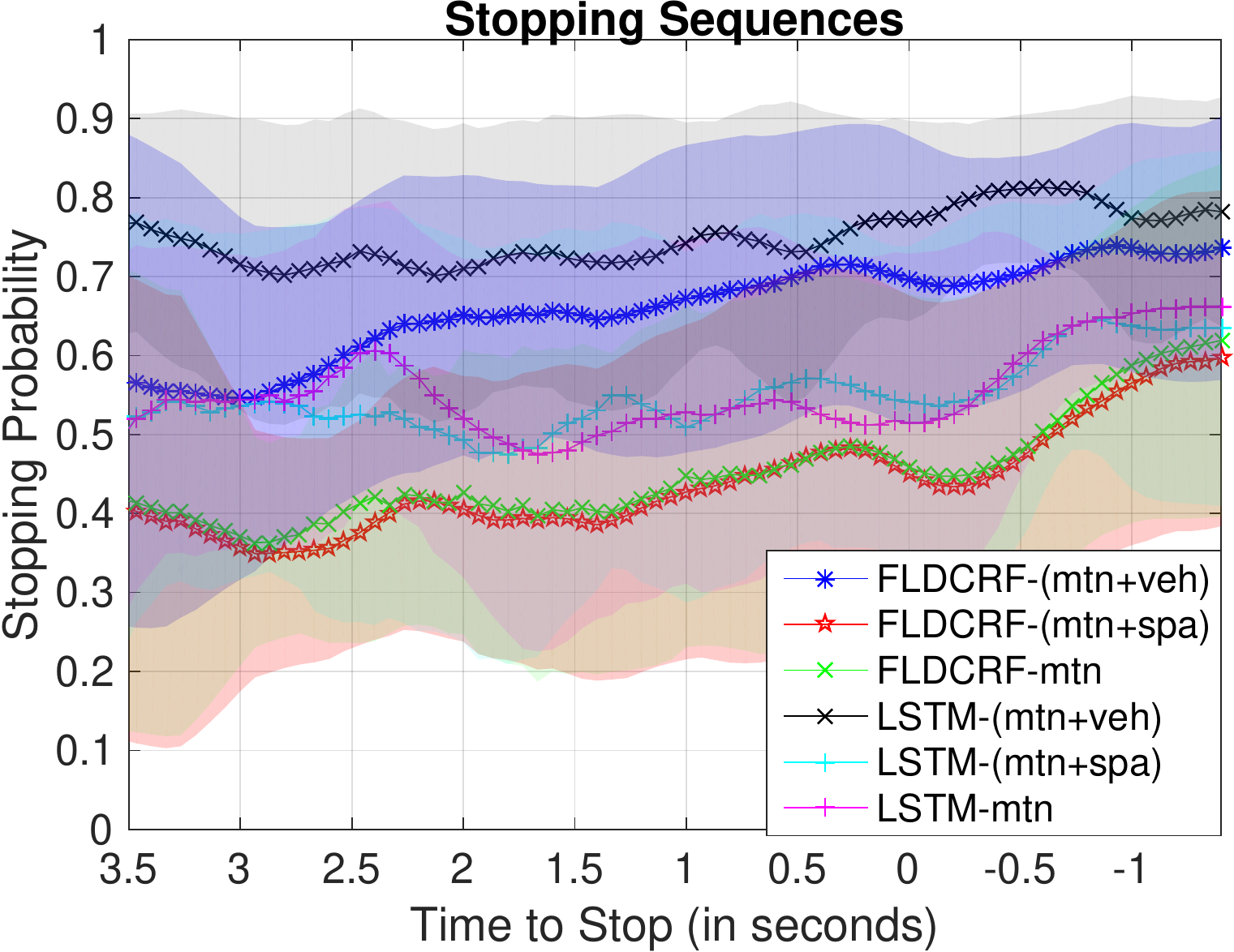} }}%
\quad
\subfloat[\scriptsize{Starting Sequences}]{{\includegraphics[width=4.1cm]{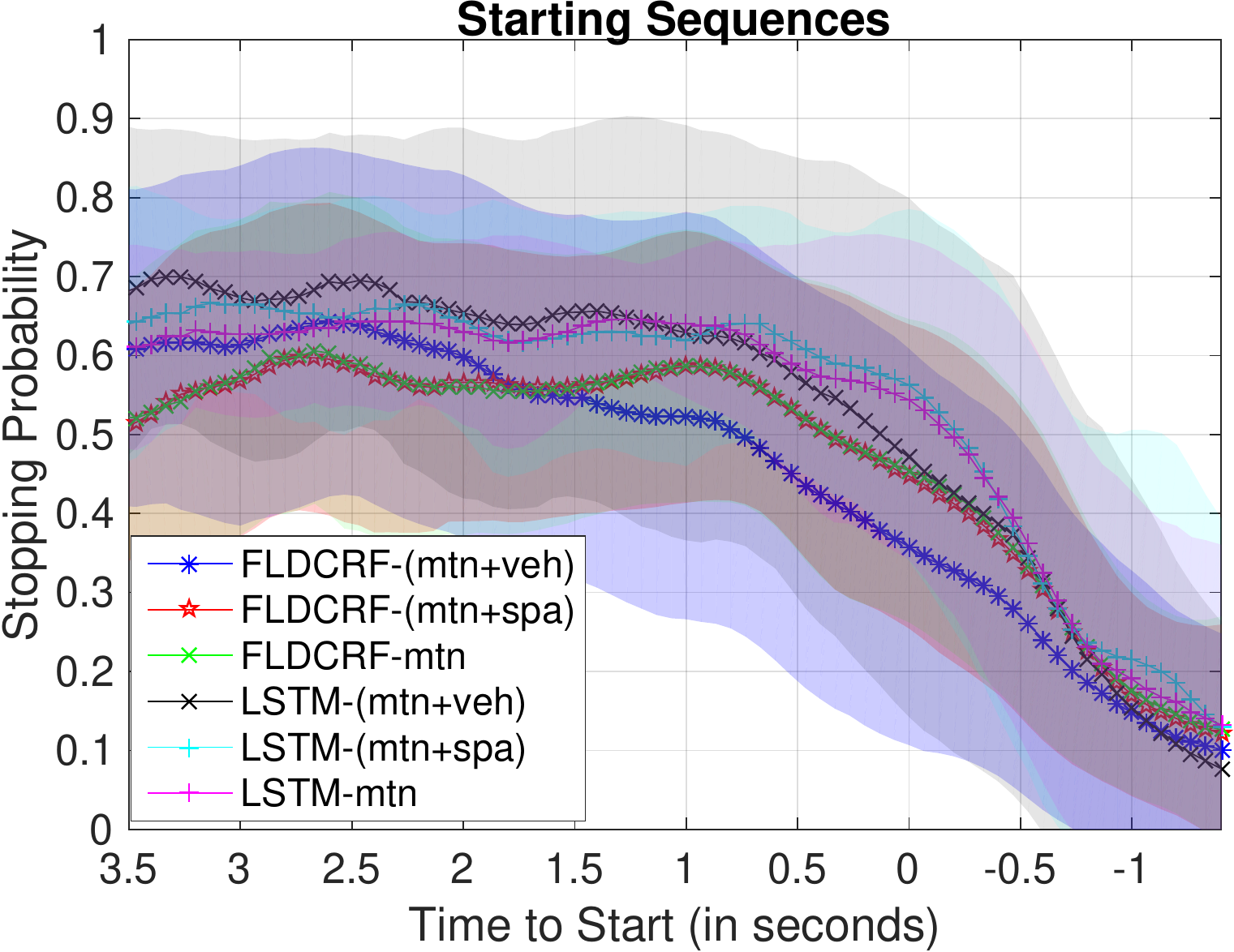} }}%
\quad
\subfloat[\scriptsize{Standing Sequences}]{{\includegraphics[width=4.1cm]{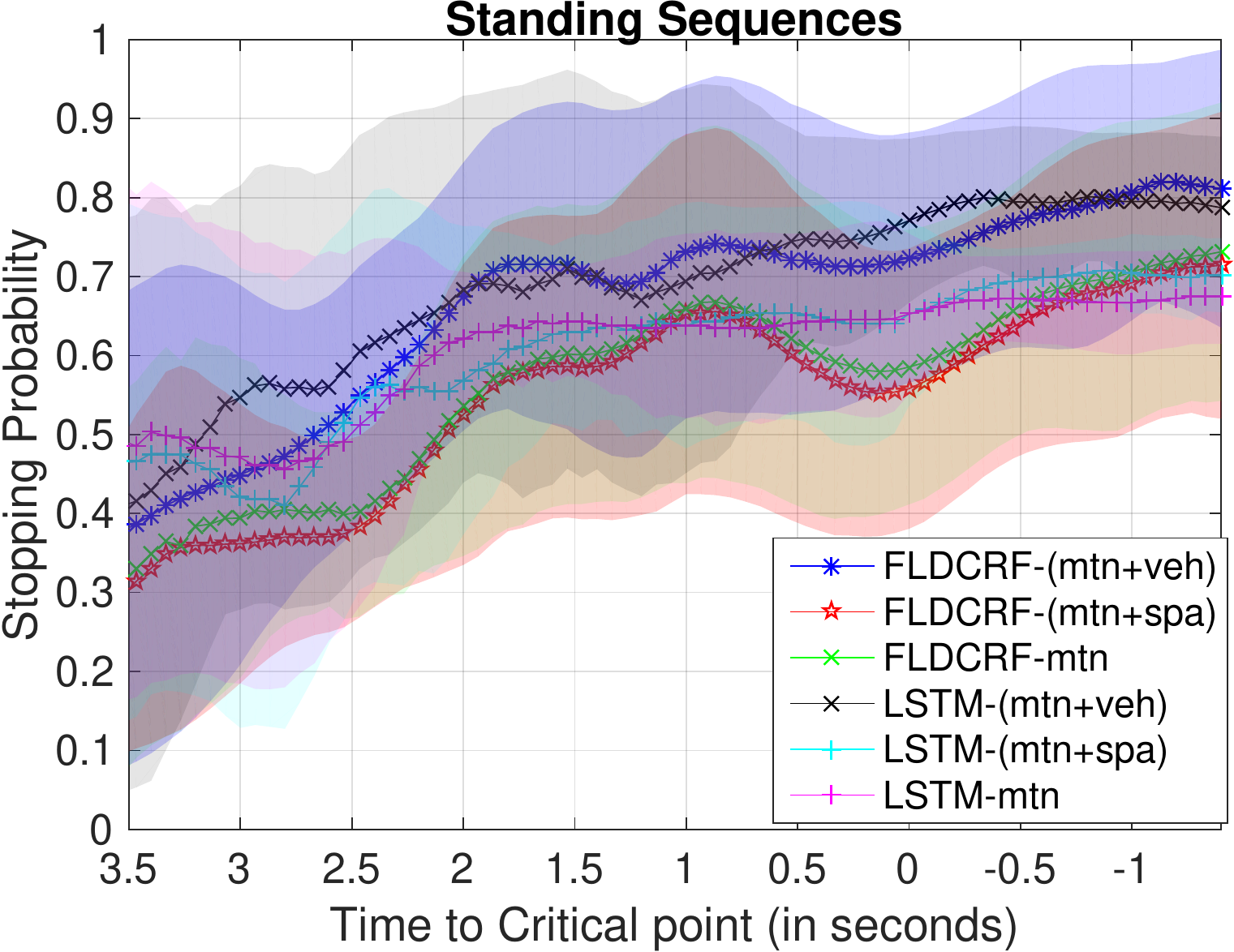} }}%
\caption{Stopping probability performances of different systems vs time on JAAD sequences.}%
\label{fig:JAAD-excess}%
\end{figure}

Figures \ref{fig:JAAD-excess}(a)-(d) compare performances of different systems on stopping, continuous crossing, starting and standing sequences of JAAD dataset by the stopping probability vs time performances.

While $FLDCRF-(mtn+veh)$ and $LSTM-(mtn+veh)$ produce comparable probability performances on continuous crossing and standing sequences, one outperforms the other in terms of stopping probability on each of the remaining two sequence types (stopping and starting).

\subsection{Early prediction tables} \label{sec:JAAD-results-tables}

The systems are compared on the task of predicting early transitions (standing-starting and crossing-stopping) in Tables \ref{table:stand-start} and \ref{table:cross-stop}. We consider accuracy of the systems over stipulated prediction windows. As expected, systems with vehicle context $FLDCRF-mv$ and $LSTM-mv$ perform significantly better than systems without vehicle context. $FLDCRF-mv$ outperforms $LSTM-mv$ in all early prediction windows. However, $LSTM-mv$ performs marginally better than $FLDCRF-mv$ after the `0' mark on crossing and stopping scenarios considered together (see Table \ref{table:cross-stop}).

\begin{table}
\vspace{2mm}
\caption{Classification accuracy of all considered JAAD standing \& starting sequences by different systems within different prediction windows.}
\centering
\renewcommand{\arraystretch}{1.2}
\setlength\tabcolsep{1.5pt}
\begin{tabular}{|p{3.35cm}|c|c|c|c|c|c|}
\hline
\multirow{2}{5cm}{\textbf{System}} & \multicolumn{6}{c|}{\textbf{Classification Accuracy(\%)}}\\
\cline{2-7}
& 2-0 s & 1.5-0 s & 1-0 s & 0.5-0 s & 0-(-0.5) s & 0-(-1) s\\
\hline \hline
FLDCRF- (mtn+veh) & \textbf{56.02} & \textbf{57.98} & \textbf{61.02} & \textbf{68.18} & \textbf{77.17} & \textbf{82.50} \\ \hline
FLDCRF- (mtn+spa) & 45.84 & 48.17 & 50.11 & 54.32 & 72.53 & 78.52 \\ \hline
FLDCRF- (mtn) & 46.15 & 47.84 & 50.45 & 55.23 & 70.91 & 78.18 \\ \hline
LSTM- (mtn+veh) & 51.82 & 53.15 & 56.48 & 60.91 & 71.92 & 77.95 \\ \hline
LSTM- (mtn+spa) & 35.24 & 36.84 & 38.07 & 40.91 & 56.16 & 66.82 \\ \hline
LSTM- (mtn) & 33.59 & 34.46 & 37.05 & 42.05 & 54.55 & 66.36 \\ \hline
\end{tabular}
\label{table:stand-start}
\vspace{-3mm}
\end{table}

\begin{table}
\vspace{3mm}
\caption{Classification accuracy of all considered JAAD continuous crossing \& stopping sequences by different systems within different prediction windows.}
\centering
\renewcommand{\arraystretch}{1.2}
\setlength\tabcolsep{1.5pt}
\begin{tabular}{|p{3.35cm}|c|c|c|c|c|c|}
\hline
\multirow{2}{5cm}{\textbf{System}} & \multicolumn{6}{c|}{\textbf{Classification Accuracy(\%)}}\\
\cline{2-7}
& 2-0 s & 1.5-0 s & 1-0 s & 0.5-0 s & 0-(-0.5) s & 0-(-1) s\\
\hline \hline
FLDCRF- (mtn+veh) & \textbf{90.47} & \textbf{91.39} & \textbf{91.83} & \textbf{93.08} & 93.68 & 95.29 \\ \hline
FLDCRF- (mtn+spa) & 74.85 & 75.57 & 77.21 & 78.85 & 78.63 & 79.62 \\ \hline
FLDCRF- (mtn) & 75.93 & 76.31 & 77.88 & 80.19 & 78.29 & 80.67 \\ \hline
LSTM- (mtn+veh) & 86.73 & 87.94 & 90.10 & \textbf{93.08} & \textbf{95.90} & \textbf{95.38} \\ \hline
LSTM- (mtn+spa) & 73.53 & 75.77 & 77.12 & 80.38 & 82.39 & 85.87 \\ \hline
LSTM- (mtn) & 75.87 & 76.79 & 77.4 & 78.46 & 83.93 & 87.5 \\ \hline
\end{tabular}
\label{table:cross-stop}
\vspace{-3mm}
\end{table}

\subsection{Failed cases} \label{sec:JAAD-failed}

Also described in Main text with the figures. See pages 12-13.

Fig. \ref{fig:JAAD-failed} depicts failures in bold red from $FLDCRF-mv$ system. We denote a pedestrian sequence by its $video\_id$ and $pedestrian\_id$ in JAAD dataset, i.e. $<$$video\_id$$>$\_$<$$pedestrian\_id$$>$.

Crossing sequences $0161\_1$ and $0177\_2$, where the system fails to establish stable and accurate (stopping probability $<$ 0.5) outputs after -0.25 s, are highlighted in Fig. \ref{fig:JAAD-failed}a. In sequence  0161\_1, we find a momentary prediction glitch within [-0.5 -1] s window caused by a temporary hesitance from the pedestrian to continue crossing. The vehicle happened to be quite far from the pedestrian during the glitch avoiding a critical failure by the system. The failure in sequence 0177\_2 is caused by inappropriate vehicle context as the ego-vehicle is moving fast quite near ($<$15 m) the pedestrian when the crossing event commenced in a different lane. However, the vehicle decelerated within a short period and stable, accurate output was achieved shortly after the -1.5 s mark. Such errors can be corrected by adding the lane information as context.


All stopping scenarios were predicted correctly by the $FLDCRF-mv$ system, latest by 0.75 s after the stopping instant. The system fails to make early prediction (before `0') of the event on the highlighted sequences (0334\_2 and 0336\_1). A few sequences fall below probability 0.5 after the -1 s mark, but all such cases correspond to a starting event followed by the stopping event.

We have found early prediction of the `starting' event quite challenging. Fig. \ref{fig:JAAD-failed}c highlights three starting scenarios that fail to be stable and accurate by -1.5 s on the curves. However, all of them become stable and accurate shortly after the -2 s mark. 

Standing sequence $0208\_2$ is wrongly predicted as `crossing' by $FLDCRF-mv$ before and well after ($\approx$1 s) the defined critical point. A possible reason behind this is incomplete and inaccurate vehicle context. In this case, the other vehicles before the ego-vehicle are primarily responsible for the pedestrian to remain stationery (see Fig. \ref{fig:stand_fail}). The ego-vehicle is moving slowly during the event in a busy scene and is quite far from the pedestrian, causing the inaccurate context. We defined the critical point randomly due to limited number of images in the sequence. In such cases, we need to consider more complete pedestrian-vehicle interactions, which include other vehicles in the scene.

\section{NTU data failed cases} \label{Appen:NTU}

We analyze the stability of sequence predictions and failed cases by the $FLDCRF-msv$ system on NTU data, as depicted in Fig. \ref{fig:NTU-failed}. Material with the figure is also available in main text, see pages 10-11.

Fig. \ref{fig:NTU-failed} shows individual sequence outputs by the $FLDCRF-msv$ system. We highlight sequences (in bold red) where the system fails to make early prediction (i.e., before `0') of the events. 

All crossing events were predicted correctly before respective \textit{crossing instants} (see Fig. \ref{fig:NTU-failed}a). An unwanted spike can be observed on sequence $crossing\_17$ near the `0' mark, caused by an approaching vehicle. However, the system was able to correct the prediction before the \textit{event instant} to avoid any critical failure. 

Individual probability outputs of 35 stopping sequences by $FLDCRF-msv$ system are displayed in Fig. \ref{fig:NTU-failed}b. The system fails to make early prediction (before `0') of the stopping event in the two highlighted sequences (by bold red), $stopping\_9$ and $stopping\_32$.


\end{document}